\documentclass[runningheads]{llncs}
\usepackage{graphicx}

\usepackage{tikz}
\usepackage{comment}
\usepackage{amsmath,amssymb}
\usepackage{color}

\usepackage[accsupp]{axessibility}

\usepackage{tabularx}
\usepackage{bbding}
\usepackage{algorithm}
\usepackage{soul}
\usepackage{multirow}
\usepackage{tipa}
\usepackage{bbold}
\usepackage{mathrsfs}
\usepackage{colortbl}
\usepackage{booktabs}
\usepackage{hyperref}
\usepackage{wrapfig, lipsum}
\newcounter{savefootnote}

\begin{document}
\pagestyle{headings}
\mainmatter
\def\ECCVSubNumber{5270}

\newcommand\red[1]{\textcolor{red}{#1}}
\newcommand\blue[1]{\textcolor{blue}{#1}}
\newcommand\gray[1]{\textcolor{gray}{#1}}
\newcommand\green[1]{\textcolor{green}{#1}}
\newcommand\magenta[1]{\textcolor{magenta}{#1}}

\definecolor{coolblack}{rgb}{0.0, 0.18, 0.39}
\definecolor{darkblue}{rgb}{0.0, 0.0, 0.55}
\definecolor{brilliantrose}{rgb}{1.0, 0.33, 0.64}
\definecolor{tablegray}{gray}{.9} 
\newcommand\tablecheck[1]{\textcolor{black}{#1}}
\newcommand\tablecross[1]{\textcolor{green}{#1}}
\definecolor{aureolin}{rgb}{0.99, 0.93, 0.0}
\definecolor{ballblue}{rgb}{0.13, 0.67, 0.8}

\newcommand{\etal}{\textit{et al.}}

\newcommand{\norm}[1]{\left\lVert#1\right\rVert}

\newcommand{\ours}{\texttt{COMPOSER}\ }
\newcommand{\ourseos}{\texttt{COMPOSER}} 

\newcommand{\mtx}{Multiscale  Transformer\ }
\newcommand{\mtxeos}{Multiscale  Transformer}

\newcommand{\APP}{Appendix\ }
\newcommand{\APPeos}{Appendix}

\newcommand{\bestGroupAccVolleyball}{$94.6$}

\title{\ourseos: Compositional Reasoning of Group Activity in Videos with Keypoint-Only Modality}

\titlerunning{Compositional Reasoning of Group Activity in Videos}
\author{Honglu Zhou\inst{1}\thanks{Work done as a NEC Labs intern.} \and
Asim Kadav\inst{2} \and
Aviv Shamsian\inst{3} \and
Shijie Geng\inst{1} \and
Farley Lai\inst{2} \and
Long Zhao\inst{4} \and
Ting Liu\inst{4} \and
Mubbasir Kapadia\inst{1} \and
Hans Peter Graf\inst{2}\index{Graf, Hans Peter}}
\authorrunning{H. Zhou et al.}
\institute{
Department of Computer Science, Rutgers University, Piscataway, NJ, USA\\
\email{\{hz289,sg1309,mk1353\}@cs.rutgers.edu} \and
NEC Laboratories America, Inc., San Jose, CA, USA\\ 
\email{\{asim,farleylai,hpg\}@nec-labs.com} \and
Bar-Ilan University, Israel\\ 
\email{aviv.shamsian@biu.ac.il} \and
Google Research, Los Angeles, CA, USA\\
\email{\{longzh,liuti\}@google.com}
}

\maketitle

\begin{abstract}

Group Activity Recognition  
detects the activity collectively performed by a group of actors,
which requires compositional reasoning of 
actors and objects.
We approach the task
by modeling the video 
as tokens that represent the multi-scale semantic concepts in the video.
We propose \ourseos, a Multiscale Transformer based architecture that
performs attention-based \textit{reasoning} over tokens at each scale and learns group activity \textit{compositionally}.
In addition, prior works suffer from scene biases with privacy and ethical concerns.
We only use the keypoint modality 
which reduces scene biases
and prevents 
acquiring detailed visual data that may contain private or biased information of users.
We improve the multiscale representations 
in \ourseos\
by clustering the intermediate scale representations, while maintaining consistent cluster assignments between scales. 
Finally, we use techniques such as auxiliary prediction and 
data augmentations 
tailored to the keypoint signals
to aid model training.
We demonstrate the model's strength and interpretability on two widely-used datasets (Volleyball and Collective Activity).
\ours achieves up to $+5.4\%$ improvement with just the keypoint modality\setcounter{savefootnote}{\value{footnote}}
\setcounter{footnote}{0}\footnote{Code is available at \url{https://github.com/hongluzhou/composer}}.

\keywords{Keypoint-only group activity recognition $\mathord{\cdot}$ Compositionality $\mathord{\cdot}$ 
Multiscale representations 
$\mathord{\cdot}$ Transformer 
$\mathord{\cdot}$ Video understanding}  

\end{abstract}

\section{Introduction}
\label{sec:intro}

\begin{figure}[t]
\centering
	\includegraphics[scale=0.15]{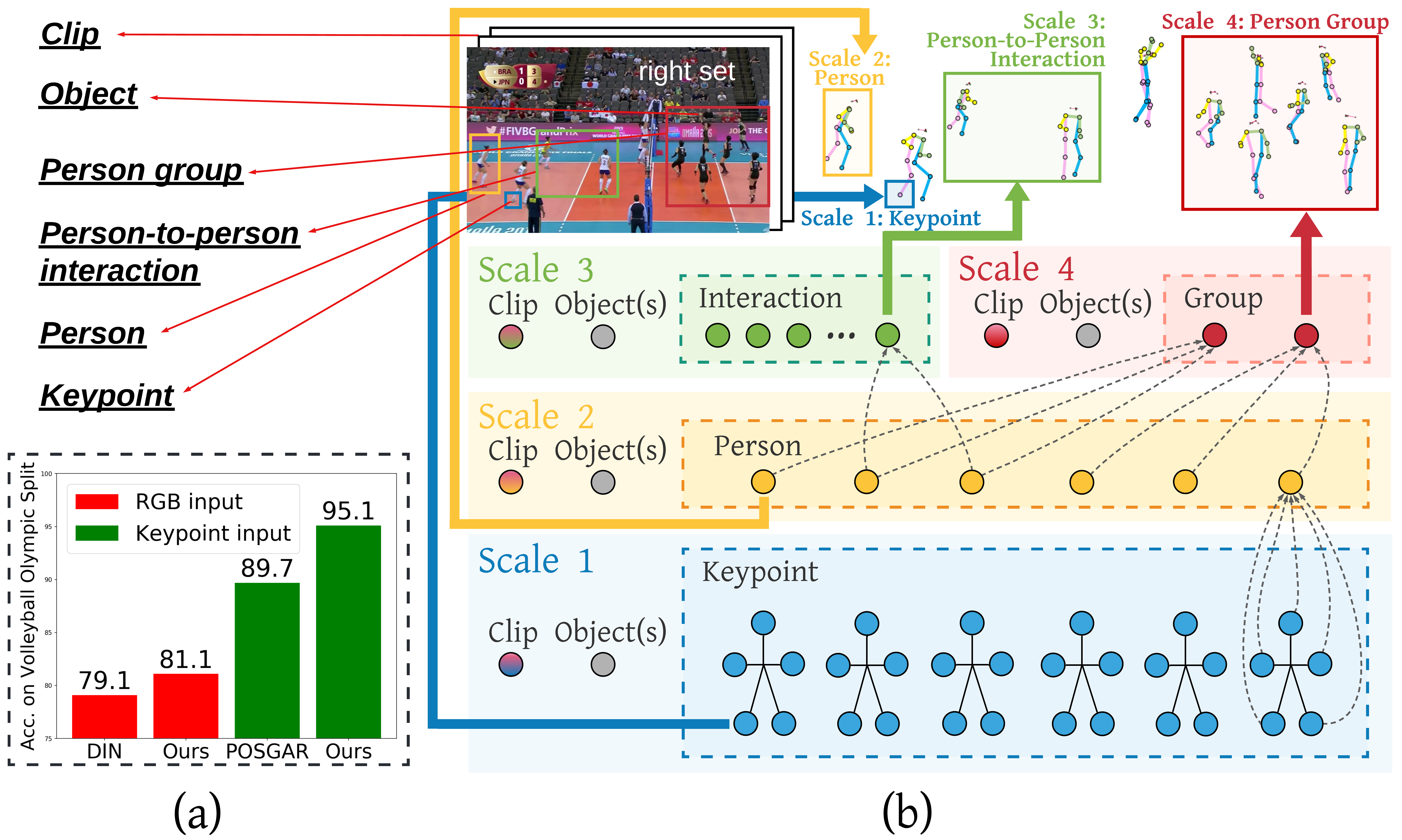}
    \caption{\textbf{(a) The keypoint-only setup generalizes better for GAR.} The Volleyball Olympic split~\cite{POGARS} ensures videos having vastly different scene background between training and testing, which can examine GAR model's scene generalization ability. RGB-based methods severely suffer from scene biases and have poor model generalizability. 
\textbf{(b) Main idea.} We propose \ours 
that uses keypoint only modality
for GAR 
by modeling a 
video
as 
\textit{tokens} that represent the multiscale semantic concepts in the video, which include \textit{keypoint}, \textit{person}, person-to-person \textit{interaction}, person \textit{group}, \textit{object} if present, and the 
\textit{clip}.
Four scales are formed by grouping actor-related tokens according to their semantic hierarchy.   
Representations of tokens in coarser scales are learned and aggregated from tokens of the finer scales.
\ours (Fig.~\ref{fig:composer}) facilitates compositional reasoning of group activity in videos. 
}
\label{fig:four_scales}
\end{figure}

Group Activity Recognition (GAR) detects
the activity collectively performed by a group of actors 
in a short video clip~\cite{choi2013understanding,wu2021comprehensive}.  
GAR has widespread societal implications in a variety of domains including security, surveillance, kinesiology, sports analysis, robot-human interaction, and rehabilitation~\cite{actor-transformer,pramono2020empowering,yuan2021spatio,ehsanpour2020joint}.

The task requires addressing two challenges.
First, GAR requires a \textit{compositional understanding}
of the scene~\cite{abkenar2019groupsense}.
Because of the crowded scene, it is challenging to learn meaningful representations for GAR over the entire scene~\cite{wu2021comprehensive}. 
Since group activity often consists of
sub-group(s) of actors and scene objects, the final action label depends on a 
compositional understanding of these entities~\cite{wu2021comprehensive,yuan2021learning}.
Second, GAR benefits from \textit{relational reasoning} over scene elements to understand the relative importance of entities and their interactions~\cite{prl,sam}. For example, in a volleyball game,
persons around the ball performing the jumping action are more important than others standing in the scene.

Existing work has proposed to jointly learn the group activity with individual actions~\cite{ibrahim2016hierarchical,cern,stagnet,rcrg,ssu,crm} or person sub-groups~\cite{GroupFormer,nakatani2021group,ehsanpour2020joint} for
a compositional understanding of the group activity. Meanwhile, graph~\cite{yuan2021spatio,ibrahim2018hierarchical,arg,prl} and transformer~\cite{actor-transformer,GroupFormer} based models have been proposed for relational reasoning over scene entities. However,
these methods 
do not sufficiently make use of the 
multiscale scene elements in the GAR task
by modeling over entities at either one semantic scale (e.g., person~\cite{actor-transformer,yuan2021spatio,arg,prl}) or two scales (person and person group~\cite{GroupFormer,nakatani2021group,ehsanpour2020joint}, or keypoint and person~\cite{GIRN}).
More importantly, explicit multiscale modeling is neglected, lacking consistent 
compositional representations for
the 
group action tasks. 
Furthermore, majority of the prior GAR methods rely on the RGB modality (see Table.~\ref{table:vd_cad}), which causes the model more likely to have privacy and ethical issues when deployed in real-world applications~\cite{hinojosa2021learning}.
Last but not least, the RGB input hinders the model's robustness to changes in background, lighting conditions or textures, and often results in poor model generalizability due to scene biases (see Fig.~\ref{fig:four_scales} (a))~\cite{scenebias,singh2020don}.

In this paper, we present \ours 
that
addresses \textit{compositional learning} of entities in the video and \textit{relational reasoning} about these entities. 
Inspired by how humans are particularly adept at
representing
objects in different granularities meanwhile reasoning  their interactions
to turn 
sensory signals into a high-level knowledge~\cite{hudson2019learning,lake2017building}, 
we approach GAR
by modeling a video as
tokens that represent the multi-scale semantic concepts in the video (Fig.~\ref{fig:four_scales} (b)).
Compared to the aforementioned prior works, we consider more fine-grained scene entities that are grouped into \textit{four} scales.
By combining the scales together with \mtx (Fig.~\ref{fig:mtx}), 
\ours provides attention-based reasoning over tokens at each scale,
which makes the higher-level understanding of the group activity possible. 
Moreover, \ours uses only the keypoint modality.
Using only the $2$D (or $3$D) keypoints as input, our method can prevent the sensor camera from acquiring detailed visual data that may contain private or biased information of users~\footnote{Even for the keypoint extraction backbone which our method is agnostic to, there are existing works~\cite{hinojosa2021learning} that perform privacy-preserving keypoint estimation.}. 
Keypoints also allow the model to focus on the action-specific cues, and help the model be more invariant to the scene biases.
\ours 
generalizes much better 
to testing data with different scene backgrounds 
(see the Volleyball Olympic split results in Table.~\ref{table:vd_olympic}).

\begin{figure}[t]
\centering
    \includegraphics[scale=0.20]{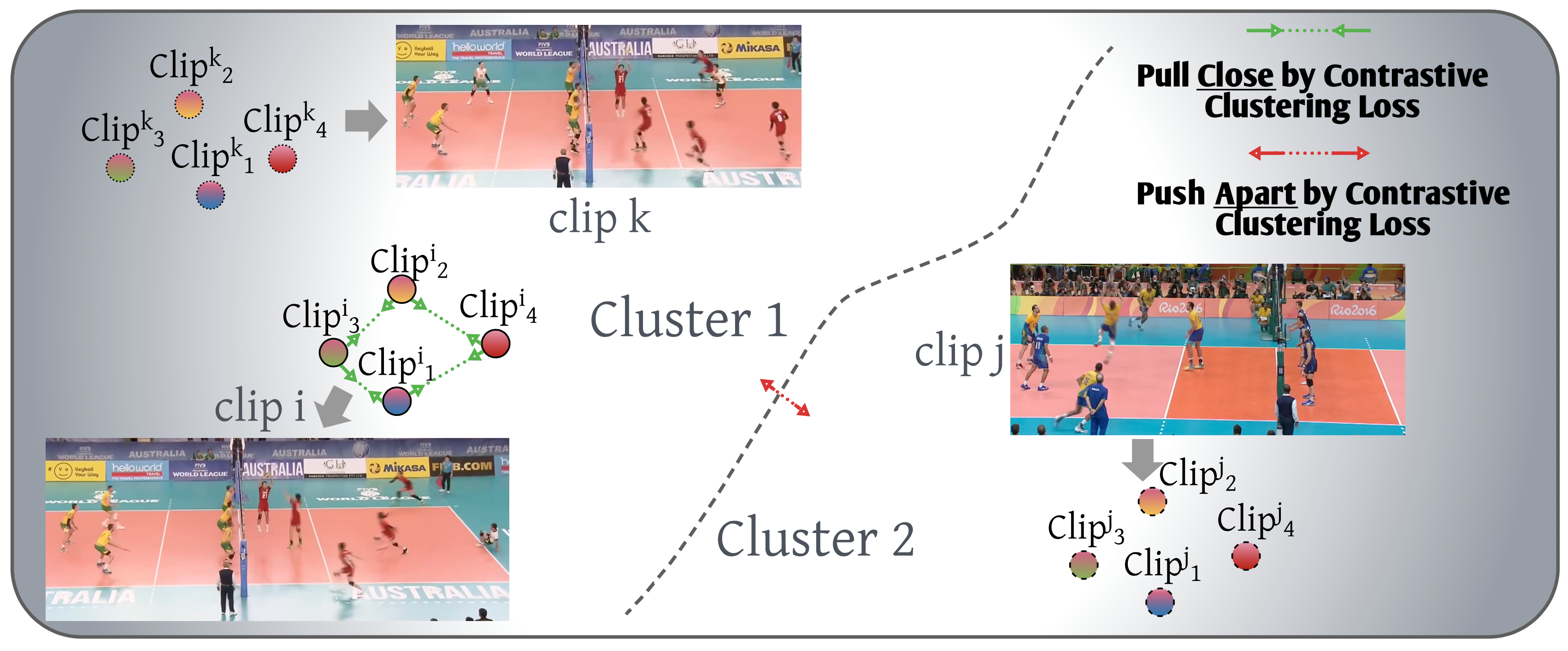}
    \caption{\textbf{Embedding space learned by \ourseos.}
	 \ours exploits a contrastive clustering objective (Sec.~\ref{subsec:contrastive}) to learn \textit{consistent} multiscale representations for GAR. This is achieved
	 by clustering clip representations learned at all scales.
	 The clustering objective encourages an ``agreement'' between scales on the high-level knowledge learned (`Pull Close' representations of the same clip).  Contrastive learning is performed on the clusters, which also helps the model to discriminate between clips 
	 with different semantic characteristics (`Pull Close' representations of the semantically-similar clips and `Push Apart' those that are semantically-different). 
	 In the illustration, we use subscript to denote the scale and use superscript to indicate different clips.}
\label{fig:embedding_space}
\end{figure}

\ours learns 
\textit{consistent} multiscale representations
which boost the performance for GAR (Fig.~\ref{fig:embedding_space}). This is achieved by
contrastive clustering assignments of clips.
Intuitively,  a model can recognize the group activity using 
representations of entities at just one particular scale. Hence, we consider representations of the clip token learned across scales as representations of different \textit{views} of the clip.  
Such perspective allows us to cluster clip representations learned at all scales while enforcing consistency between cluster assignments produced from different scales of the \textit{same} clip. 
In order to enforce this consistency, we follow~\cite{swav} and use a \emph{swapped} prediction mechanism where we predict the cluster assignment of a scale from the representation of another scale.
However, distinct from related works~\cite{swav,asano2020labelling,chen2021multimodal},
which use information from
multiple augmentations or modalities for self-supervised learning from unlabelled images or videos,
we use information from multiple scales for the task of group activity recognition.
Contrasting clustering assignments enhance
our intermediate representations and the overall performance. 
Finally, we use techniques such as auxiliary prediction at each scale
and data augmentation methods
such as \textit{Actor Dropout} 
to aid training.

Our contributions are three-fold: 
\begin{enumerate}
    \item We present \ours 
for compositional reasoning of group activity in videos. \ours can distill and convey high-level semantic knowledge from the elementary elements of the human-centered videos.
We learn contrastive clustering assignment to improve the multiscale representations. By maintaining a consistent cluster assignment across the multiple scales of the \textit{same} clip, an agreement between scales on the high-level knowledge
    learned 
    can be promoted to
    optimize the representations across scales.
    \item We use only
    the keypoint modality that allows \ours
    to address the privacy and ethical concerns and 
    to be robust to changes in background, 
    with 
    auxiliary prediction and data augmentation methods tailored to learning group activity from the keypoint modality.
    \item  We demonstrate the model’s strength and interpretability on two commonly-used datasets (Volleyball and Collective Activity) and \ours achieves up to $+5.4\%$ improvement using just the keypoint modality.
 
\end{enumerate}

\section{Related Work}
\label{sec:related}

Much of the recent research on GAR explores how to capture the actor relations~\cite{ibrahim2018hierarchical,azar2019convolutional,arg,prl,pramono2020empowering}.
Several works tackle this problem from a graph-based perspective~\cite{ibrahim2018hierarchical,lu2019gaim,sam,higcin}.
Some utilize attention modeling~\cite{stagnet,xu2020group,lu2019gaim,yuan2021learning} including using
Transformers~\cite{actor-transformer,GroupFormer}.
Existing works 
have primarily used RGB- and/or optical-flow-based features with RoIAlign~\cite{he2017mask} to represent actors~\cite{higcin,stagnet,arg,ssu}.
A few recent works replace or augment these
features with 
keypoints/poses of the actors~\cite{POGARS,GIRN,actor-transformer,yuan2021learning}. 
 In this paper, we use only 
 the light-weight coordinate-based
 keypoint representation. 
We propose a \mtx block to hierarchically reason about entities at different semantic scales and we aid learning group activities by improving the musicale representations. Please see an in-depth discussion on related works in Appendix~\ref{appendix_sec:related}.

\section{Methodology}
\label{sec:method}

\begin{figure}[t]
\centering
	\includegraphics[scale=0.28]{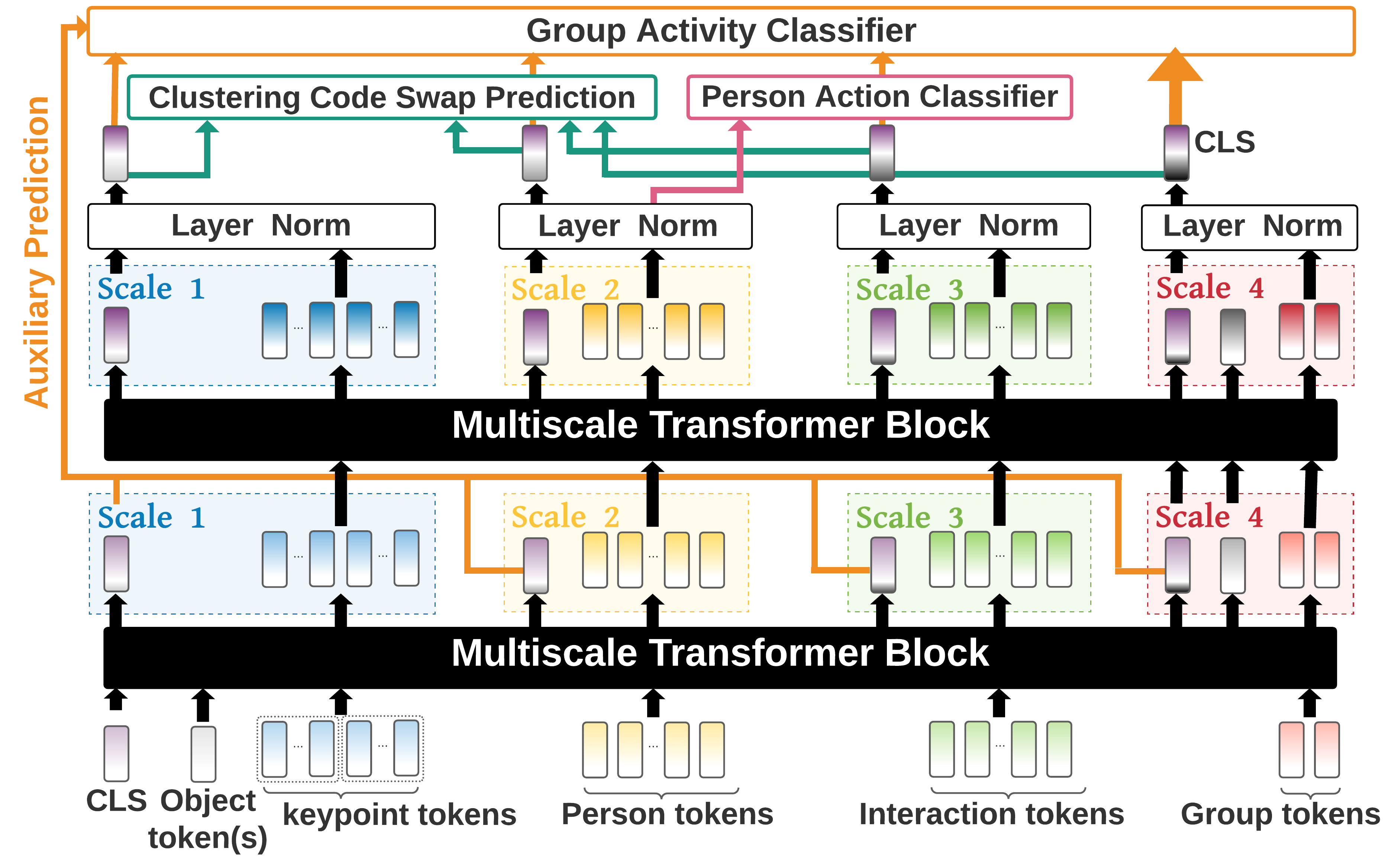}
    \caption{\textbf{\ourseos.}
	Given tokens that represent the
	multiscale
	semantic concepts (Fig.~\ref{fig:four_scales})
	in the
	human-centered 
	video,
	\ours jointly learns group activity, individual actions and contrastive clustering assignments of clips. Auxiliary predictions are enforced to aid training (Sec.~\ref{subsubsection:loss}).
}
\label{fig:composer}
\end{figure}

We present  
\ours (Fig.~\ref{fig:composer}), a novel \mtx based architecture for GAR.
In Sec.~\ref{subsec:entity_repre},
we describe the multi-scale semantic tokens representing a video with group activities.
We introduce \ours
and especially its reasoning module \mtx in Sec.~\ref{subsec:multiscale_tx}.
We describe data augmentations in Sec.~\ref{subsec:data_augmentation} and the exact formulation of auxiliary prediction in Sec.~\ref{subsubsection:loss}.

\subsection{Tokenizing a Video as Hierarchical Semantic Entities}
\label{subsec:entity_repre}

We model a video as semantic tokens 
that allow our method easily adaptable to understanding any videos with multi-actor multi-object interactions~\cite{luo2021moma}.

\noindent \textbullet\ \textbf{Person Keypoint}. 
We define a person keypoint token, $\mathbf{k}_p^j\in \mathbb{R}^d$ that represents a keypoint joint  $j$ ($j=1,\dots,j^{\prime}$) of person $p$ ($p=1,\dots,p^{\prime}$)  in all timestamps,
where $j^{\prime}$ is the number of joint
types and $p^{\prime}$ is the number of actors.
The initial $d$-dimensional person keypoint token is learned by encoding the numerical coordinates (in the image space) of a certain keypoint track\footnote{We use track-based representations~\cite{POGARS,zappardino2021learning,actor-transformer,GroupFormer} to represent each token.}. 
The procedure of encoding includes coordinate embedding, time positional embedding, keypoint type embedding, and OKS-based feature embedding~\cite{snower202015} to mitigate the issue of noisy estimated keypoints. 
Details
are available in Appendix~\ref{subsec:entity_repre}.

\noindent \textbullet\
\textbf{Person}. 
A person token is defined as  $\mathbf{p}_p\in \mathbb{R}^d$, initially obtained by aggregating the
standardized 
keypoint coordinates of person $p$ over time
through concatenation and FFN-based transformation.

\noindent \textbullet\ \textbf{Person-to-Person Interaction}. 
Modeling the person-to-person interactions is critical for GAR~\cite{wu2021comprehensive}. Unlike existing works that typically consider an interaction as an \textit{edge} connecting two person nodes and learn a scalar to depict its importance~\cite{sam}, we model interaction as \textit{nodes} (tokens) to allow for the modeling of complex higher-order interactions~\cite{luo2021moma}.
The person-to-person interaction token is defined as $\mathbf{i}_i\in \mathbb{R}^d$ where $i=1,\dots,p^{\prime}\times(p^{\prime}-1)$ (bi-directed interactions). Initial representation of the interaction between person $p$ and $q$ is learned from
concatenation of
$\mathbf{p}_p$ and $\mathbf{p}_q$, followed by FFN-based transformation.

\noindent \textbullet\ \textbf{Person Group}. 
We define the group token $\mathbf{g}_g \in \mathbb{R}^d$ where $g=1,\dots,g^{\prime}$ for videos where sub-groups are often separable.
$g^{\prime}$ denotes the num. of sub-groups in the video.
Given the person-to-group mapping which can be obtained through various mechanisms (e.g., heuristics~\cite{GIRN}, k-means~\cite{GroupFormer}, etc~\cite{ehsanpour2020joint,koshkina2021contrastive}.), representation of a group is an aggregate over representations of persons in the group similarly through concatenation and FFN.

\noindent \textbullet\ \textbf{Clip}. The special \text {[\texttt{CLS}]} token ($\in \mathbb{R}^d$) is a learnable embedding vector and is considered as the clip representation. 
\texttt{CLS} stands for classification and is often used in Transformers to ``summarize'' the task-related representative information from all tokens in the input sequence~\cite{devlin2018bert}.

\begin{figure}[t]
\centering
	\includegraphics[scale=0.28]{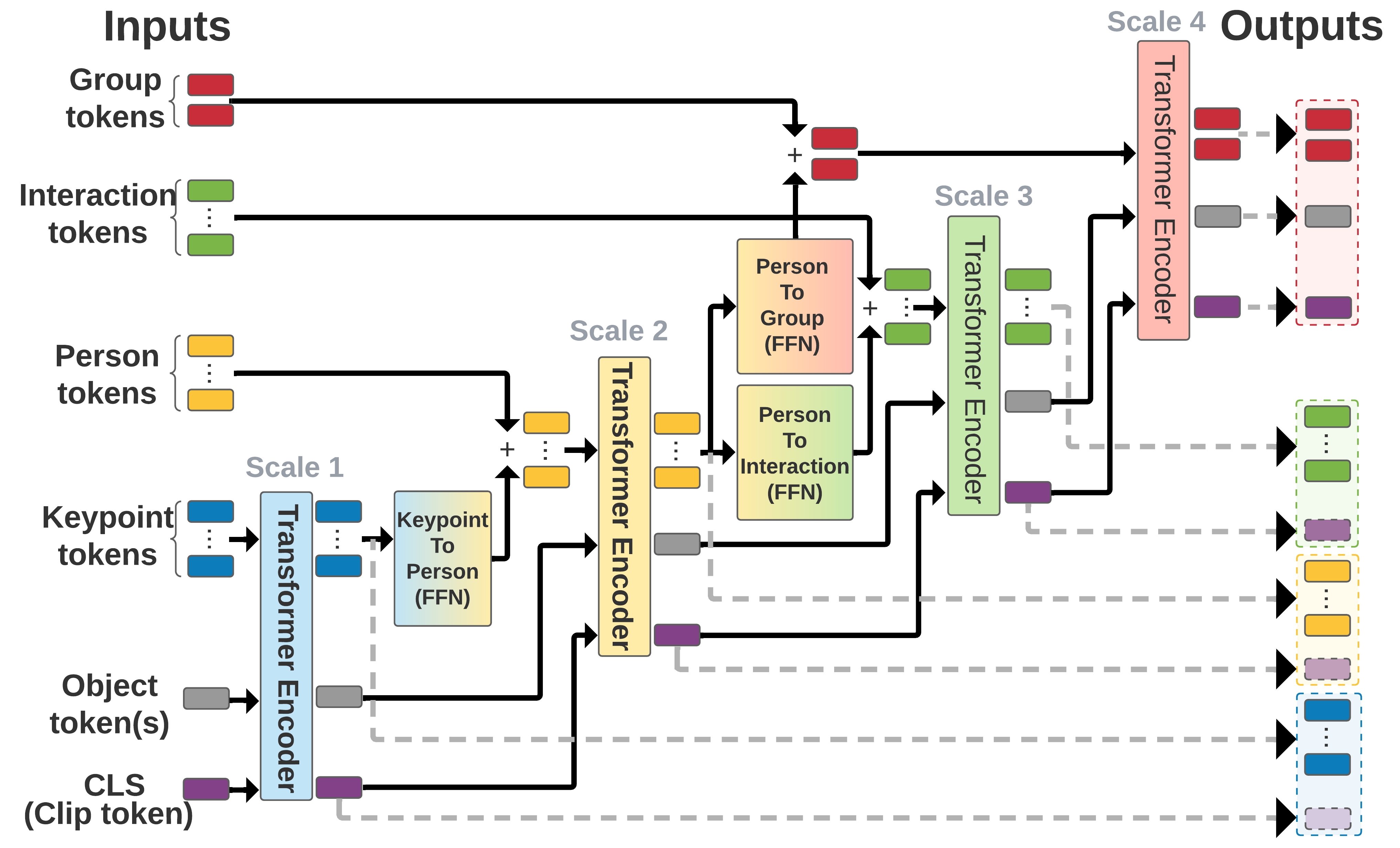}
    \caption{\textbf{\mtx}performs relational reasoning with four
	Transformer Encoders to operate self-attention on tokens of each scale, while stringing tokens of the four
	scales together
	with FFNs and Skip Connections
	to learn hierarchical representations that make a high-level understanding
	of group activity
	possible. 
}
\label{fig:mtx}
\end{figure}

\noindent \textbullet\ \textbf{Object}. 
Scene objects can
play a crucial role in videos where human(s) interact with object(s).
E.g., in
a volleyball game where one
person 
is spiking and multiple nearby actors are all jumping with arms up, it can be difficult to tell which person is the key person with information of just the person keypoints due to their similar poses.
The ball keypoints can help to distinguish the key person. 
Object keypoints can be used to represent an object in the scene with similar benefits of person keypoints (e.g., to boost model robustness~\cite{jaiswal2021keypoints}). 
Object keypoint detection~\cite{blomqvist2022semi,lu2021few} benefits downstream tasks such as human action recognition~\cite{huangyi}, object detection~\cite{jaiswal2021keypoints,yang2019reppoints}, tracking~\cite{nebehay2014consensus}, etc~\cite{kulkarni2019unsupervised}.
Thus, we use object keypoints to represent each object for GAR.
We denote object token $\mathbf{e}_e\in \mathbb{R}^d$ where $e=1,\dots,e^{\prime}$ and $e^{\prime}$ is the maximal number of objects a video might have.
Similar to person tokens, the initial object tokens are learned from aggregating the coordinate-represented object keypoints.

\subsection{\mtx}
\label{subsec:multiscale_tx}

\mtx takes a sequence of multiple-scale tokens
as input, and refines
representations of these tokens.
Specifically,
tokens of the four scales are:
\begin{equation}
\label{eq:tokens_scales}
\begin{aligned}
& \text{\textit{Scale 1:}}  \quad \left\{\left[\text{\texttt{CLS}}\right], \mathbf{e}_1, \cdots, \mathbf{e}_{e^{\prime}}, \mathbf{k}_1^1, \cdots, \mathbf{k}_{p^{\prime}}^{j^{\prime}} \right\},\\
& \text{\textit{Scale 2:}}  \quad  \left\{\left[\text{\texttt{CLS}}\right], \mathbf{e}_1, \cdots, \mathbf{e}_{e^{\prime}}, \mathbf{p}_1, \cdots, \mathbf{p}_{p^{\prime}} \right\},\\
& \text{\textit{Scale 3:}}  \quad  \left\{\left[\text{\texttt{CLS}}\right], \mathbf{e}_1, \cdots, \mathbf{e}_{e^{\prime}}, \mathbf{i}_1, \cdots, \mathbf{i}_{{p^{\prime}}\times({p^{\prime}}-1)} \right\},\\
& \text{\textit{Scale 4:}}  \quad  \left\{\left[\text{\texttt{CLS}}\right], \mathbf{e}_1, \cdots, \mathbf{e}_{e^{\prime}}, \mathbf{g}_1, \cdots, \mathbf{g}_{g^{\prime}} \right\}.
\end{aligned}
\end{equation}
We utilize a Transformer encoder~\cite{vaswani2017attention} at each scale to perform relational reasoning of tokens in that scale.
We review 
details of
Transformer in Appendix~\ref{appendix_subsec:transformer_review}.

Hierarchical representations of tokens are maintained in an elaborately designed \mtx block (Fig.~\ref{fig:mtx}). In the \mtx block, operations in the four scales are the same (but with different parameters) to maintain simplicity.
Specifically, given a sequence of tokens of scale $s$ (Eq.~\ref{eq:tokens_scales}), Transformer encoder outputs refined representations of these tokens.
Then, concatenation and FFN are used to aggregate refined representations of \textit{actor-related} tokens,
in order to form representations of actor-related tokens in the subsequent coarser scale $s$+$1$. 
Such learned representations
are summed with their initial representations (input to the \mtxeos) (i.e. Skip Connection). 
The resulting 
actor-related tokens, as well as scale $s$ updated {[\texttt{CLS}]} token and object token(s) form the input sequence of the Transformer encoder in the scale $s$+$1$ (see wiring in Fig.~\ref{fig:mtx}).

\ours uses
the initial representations of the multi-scale semantic tokens (Sec.~\ref{subsec:entity_repre}) as input, and utilizes multiple blocks of \mtx to perform relational reasoning over these tokens.
With refined token representations, \ours \textit{jointly} learns group activity, individual actions and contrastive clustering of clips (the multitask-learning details are in Sec.~\ref{subsubsection:loss}).

\subsection{Contrastive Clustering for Scale Agreement}
\label{subsec:contrastive}

We consider the
clip tokens learned at different scales as representations of different \textit{views} of the clip
instance. 
Then, we cluster clip representations learned in all scales while enforcing consistency between cluster assignments produced from different scales of the clip.
This can act as regularization of the embedding space during training (Fig.~\ref{fig:embedding_space}).
To enforce consistency, we use a swapped prediction mechanism~\cite{swav} where we predict the cluster assignment 
of a scale from the representation of another scale. \ours jointly
learns GAR and the swapped prediction task 
to capture an agreement of the common semantic information hidden across the scales.

\noindent \textbf{Preliminaries.} 
Suppose $\mathbf{v}_{n,s}\in\mathbb{R}^{d}$ represents the learned representation of clip $n$
in scale $s$,
where $s\in\{1,2,3,4\}$. Following prior works~\cite{swav,khosla2020supervised}, we first project the representation to the unit sphere.
We then compute a code (i.e., cluster assignment) $\mathbf{q}_{n,s}\in\mathbb{R}^K$ 
by mapping $\mathbf{v}_{n,s}$ to a set of $K$ \textit{trainable} prototype vectors, $\left\{\mathbf{c}_{1}, \ldots, \mathbf{c}_{K}\right\}$. We denote by $C\in\mathbb{R}^{K\times d}$ the matrix whose rows are the $\mathbf{c}_{1}, \ldots, \mathbf{c}_{K}$.

\noindent \textbf{Swapped Prediction.} Suppose $s$ and $w$ denote $2$ different scales from the four representation scales.
The swapped prediction problem aims to predict the code $\mathbf{q}_{n,s}$ from 
$\mathbf{v}_{n,w}$, and $\mathbf{q}_{n,w}$ from $\mathbf{v}_{n,s}$, with the following loss function:
\begin{equation}
\label{eq:swap_loss_one_pair}
    \mathcal{L}_{\text {swap}}\left(\mathbf{v}_{n,w}, \mathbf{v}_{n,s}\right)=\ell\left(\mathbf{v}_{n,w}, \mathbf{q}_{n,s}\right)+\ell\left(\mathbf{v}_{n,s}, \mathbf{q}_{n,w}\right)
\end{equation}
where $\ell\left(\mathbf{v}_{n,w}, \mathbf{q}_{n,s}\right)$ measures the fit between 
$\mathbf{v}_{n,w}$ and 
$\mathbf{q}_{n,s}$.
$\ell\left(\mathbf{v}_{n,w}, \mathbf{q}_{n,s}\right)$ 
is the cross entropy loss between 
$\mathbf{q}_{n,s}$ and the probability obtained by taking a softmax of the dot products of $\mathbf{v}_{n,w}$ and 
prototypes in $C$:
\begin{equation}
\label{eq:swap_subloss}
\ell\left(\mathbf{v}_{n,w}, \mathbf{q}_{n,s}\right)=-\sum_{k=1}^{K} \mathbf{q}_{n,s}^{(k)} \log \frac{\exp \left(\frac{1}{\tau} \mathbf{v}_{n,w} \mathbf{c}_{k}^{\top}\right)}{\sum_{k^{\prime}=1}^{K} \exp \left(\frac{1}{\tau} \mathbf{v}_{n,w} \mathbf{c}_{k^{\prime}}^{\top}\right)}
\end{equation}
where $\tau$ is a temperature parameter.
The total loss of the swapped prediction problem is taking Eq.~(\ref{eq:swap_loss_one_pair}) computed over all pairs of scales
and all $N$ clips, 
\begin{equation}
\label{eq:swap_totalloss}
    \mathcal{L}_{\text {cluster}} = \frac{1}{N}\sum_{n=1}^{N}\left( \sum_{w,s\in\{1,2,3,4\}\\ \& w\neq s}\mathcal{L}_{\text {swap}}\left(\mathbf{v}_{n,w}, \mathbf{v}_{n,s}\right)\right)
\end{equation}

\noindent \textbf{Online Clustering.} 
This step produces the cluster assignments using the learned prototypes $C$ and the learned clip representations \textit{only within a batch}, $V\in\mathbb{R}^{B\times d}$ where $B$ denotes the batch size.
We perform the clustering in an online fashion for faster training and use the method proposed in~\cite{swav}.
Specifically, online clustering yields the codes $Q\in\mathbb{R}^{B\times K}$. 
We compute codes $Q$ 
such that all examples in a batch are equally partitioned by the prototypes (which prevents the trivial solution where every clip has the same code).
$Q$ is optimized to maximize the similarity between the learned clip representations and the prototypes,
\begin{equation}
\label{eq:optimal_transport_problem}
    \max _{Q \in \mathcal{Q}} \operatorname{Tr}\left(Q C V^{\top}\right)+\varepsilon H(Q),\\
\end{equation}
\begin{equation*}
  \mathcal{Q}=\left\{Q \in \mathbb{R}_{+}^{B \times K} \mid \mathbf{1}_{B} Q  =\frac{1}{K} \mathbf{1}_{K}, Q \mathbf{1}_{K}^{\top}=\frac{1}{B} \mathbf{1}_{B}^{\top}\right\}
\end{equation*}
where the trace $\operatorname{Tr}$ is 
the sum of the elements on the main diagonal, 
$H$
is the entropy function, and $\varepsilon$ is a parameter that controls the smoothness of the mapping. 
$\mathbf{1}_{K}\in\mathbb{R}^{K}$ and $\mathbf{1}_{B}\in\mathbb{R}^{B}$ are a vector of ones to enforce the equipartition constraint.
The continuous solution
$Q^{*}$ 
of Eq.~(\ref{eq:optimal_transport_problem})
is computed with the iterative Sinkhorn-Knopp algorithm~\cite{cuturi2013sinkhorn,swav}.

\subsection{Data Augmentation for Keypoint Modality}
\label{subsec:data_augmentation}

We use the following data augmentations to aid training and improve generalization ability of the model learned from the \textit{keypoint} modality.

\noindent \textbf{Actor Dropout} is performed by removing a random actor in a random frame, inspired by ~\cite{ngiam2021scene} that masks
agents with probabilities to predict agent behaviors for autonomous driving.
We remove actors 
by replacing the 
representation of the actor with a zero vector.

\noindent \textbf{Horizontal Flip} is often used by existing GAR methods~\cite{zappardino2021learning,POGARS,GIRN}, which is performed on the video frame level. This augmentation causes the pose of each person and positions of (left and right) sub-groups flipped horizontally.
We add a small random perturbation
on each flipped keypoint.

\noindent \textbf{Horizontal Move} means we horizontally move all keypoints in the clip by a certain number of pixel locations, which is randomly determined per video and bounded by a pre-defined number (i.e., 10). Similarly, afterwards a small random perturbation is applied
on each keypoint. 

\noindent \textbf{Vertical Move} is done similar to the Horizontal Move, except we move the keypoints in the vertical direction. 

Novel practices like Actor Dropout, Horizontal/Vertical Move and random perturbations help the model to perform GAR from noisy estimated keypoints.

\subsection{Auxiliary Prediction and Multitask Learning}
\label{subsubsection:loss}
We take the learned representation of the clip
at \textit{each} scale of \textit{each} \mtx block, and perform \textit{auxiliary group activity predictions} (Fig.~\ref{fig:composer}).
Specifically, each of the clip representations learned
at each scale of each block is sent as input to the group activity classifier to produce one GAR result.
In addition,
person representation from the last \mtx block is
the input to a person action classifier. Meanwhile, the loss of the swapped prediction problem
is computed given the learned representations of the clip 
of all $4$ scales from the last \mtx block. The total loss is:
\begin{equation}
    \mathcal{L}_{\text {total}} = \sum_{m=1}^{M-1}\mathcal{L}_{\text {groupAux}} + \lambda \left(\mathcal{L}_{\text {groupLast}} + \mathcal{L}_{\text {person}} + \mathcal{L}_{\text {cluster}}\right)
\end{equation}
where $\mathcal{L}_{\text {groupAux}}$ represents the loss from Auxiliary Prediction incurred by clip representations at different scales and early blocks of the \mtxeos, $\mathcal{L}_{\text {groupLast}}$
is from the last \mtx block, $\mathcal{L}_{\text {person}}$ is the person action classification loss, and $\mathcal{L}_{\text {cluster}}$ is the contrastive clustering loss (Eq.~\ref{eq:swap_totalloss}). 
$m$ denotes the index of the \mtx block, $M$ is the total number of the \mtx blocks, and $\lambda$ is a hyper-parameter that weights the importance of predictions from the last
block.
For metric evaluation, we use the clip
token from the last scale in the last \mtx as input to the group activity classifier.

\section{Experimental Evaluation}
\subsection{Dataset}
\label{sec:dataset}

\textbf{The Volleyball dataset}~\cite{ibrahim2016hierarchical} (VD) comprises
$4,830$ clips 
from $55$ 
videos.
The group activity labels include $8$ activities: 
$4$ main activities (\textit{set}, \textit{spike}, \textit{pass}, \textit{winpoint}) which are divided into two subgroups, \textit{left} and \textit{right}.
Each player can perform one of the $9$ actions: \textit{blocking}, \textit{digging}, \textit{falling}, \textit{jumping}, \textit{moving}, \textit{setting}, \textit{spiking}, \textit{standing} and \textit{waiting}. The dataset has a default `\textbf{Original}' split in which train/test videos were randomly splitted ($39$ train and $16$ test videos). 
A skewed `\textbf{Olympic}' split~\cite{POGARS} 
was later released in which 
train/test videos are splitted according to the match venues: 
$29$ train videos are from the same
2012 London Olympics venue, while the rest $26$ test videos are from numerous venues, and thus largely differs from the train videos w.r.t. the scene background.

\noindent \textbf{The Collective Activity dataset}~\cite{choi2009they} (CAD) is a dataset with 44 real-life videos~\cite{wu2021comprehensive}. The group activity labels are \textit{crossing}, \textit{waiting}, \textit{queueing}, \textit{walking} and \textit{talking} (person action labels have an additional `N/A' class). 
We follow prior works to merge the class \textit{crossing} and \textit{walking} into \textit{moving}~\cite{yuan2021spatio,wang2017recurrent,PCTDM,higcin}, and use the same train-test split~\cite{yuan2021spatio,arg,stagnet} and actor tracklets~\cite{yuan2021spatio,ssu}.
Please refer to Appendix~\ref{appendix_sec:imple_details} for implementation details on both datasets.

\subsection{Comparison with State-of-the-Arts}
\label{sec:sota_compare}

\begin{table}[t]
\begin{center}
\begin{minipage}{.45\textwidth}
\caption{Test accuracy on VD \textbf{under different train/test splits}.
Yellow shaded rows highlight the methods use RGB input, and blue for keypoint
  }
\label{table:vd_olympic}
\begin{tabular}{l|c|c}
\hline
\multirow{2}{*}{Model} & \multicolumn{2}{c}{\textbf{VD Acc. ($\%$) $\uparrow$}}       \\  \cline{2-3} 
   &   Olympic  &   Original  \\ \hline\hline \rowcolor{aureolin!10}
 I3D~\cite{carreira2017quo}  &   73.9    &   84.6 \\  \hline \rowcolor{aureolin!10}
 VGG-16~\cite{simonyan2014very}  &   76.4    &  91.6 \\ \hline \rowcolor{aureolin!10}
 PCTDM~\cite{PCTDM}   &    75.2   &   91.7 \\  \hline \rowcolor{aureolin!10}
 SACRF~\cite{pramono2020empowering}  &  71.1     &  91.8  \\  \hline \rowcolor{aureolin!10}
 AT~\cite{actor-transformer}  &  76.9   &   93.0  \\  \hline \rowcolor{aureolin!10}
 ARG~\cite{arg}  &     77.8 &    93.3 \\  \hline \rowcolor{aureolin!10}
 TCE-STBiP~\cite{yuan2021learning}   &   78.5     &  93.5\\  \hline \rowcolor{aureolin!10}
 DIN~\cite{yuan2021spatio}  &  79.1   &   93.6   \\ \hline
 \rowcolor{ballblue!10}
 POGARS~\cite{POGARS}   &    89.7   &  93.2  \\  \hline   \rowcolor{ballblue!10}
 \ours (ours) &    \textbf{95.1}   &  \textbf{93.7} \\ \hline 
  Improvement &    $+5.4\%$ &   $+0.1\%$   \\ 
\hline
\end{tabular}
\scriptsize{*Note: Keypoint-based methods do NOT use ball keypoint in this table in order to have a rigorous
comparison because
RGB-based methods are unaware of such info.
}
\end{minipage}
\begin{minipage}{.1 \textwidth}
\end{minipage}
\begin{minipage}{.1 \textwidth}
\end{minipage}
\begin{minipage}{.1 \textwidth}
\end{minipage}
\begin{minipage}{.5 \textwidth}
\caption{Comparisons with state-of-the-art (SOTA) methods that leverage \textbf{only keypoint information} on the \textbf{VD Original split}.  
  \ours 
  outperforms existing methods and achieves a new highest record ($+0.7\%$ improvement)
  }
\label{table:vd_keypoint}
\centering
\begin{tabular}{l|c|c|c}
\hline
\multirow{2}{*}{Model} & \multicolumn{2}{c|}{\textbf{Keypoint}}  &  \multirow{2}{*}{Acc.}     \\  \cline{2-3} 
     & Actor & Object &    \\ \hline\hline
Zappardino~\etal~\cite{zappardino2021learning}  & \tablecheck{\CheckmarkBold}    &   & $91.0$  \\  \hline
\multirow{2}{*}{GIRN~\cite{GIRN}} &  \tablecheck{\CheckmarkBold}   &   &    $88.4$   \\     \cline{2-4}\rule{0pt}{1.2EM}
  &   \tablecheck{\CheckmarkBold}   &   \tablecheck{\CheckmarkBold}      &  $92.2$    \\     \hline \rule{0pt}{1.2EM}
\multirow{2}{*}{AT~\cite{actor-transformer}}  &  \tablecheck{\CheckmarkBold}  &    &   $92.3$ \\  \cline{2-4}\rule{0pt}{1.2EM}
 &  \tablecheck{\CheckmarkBold}  &  \tablecheck{\CheckmarkBold}    &   $92.8$ \\\hline \rule{0pt}{1.2EM}
\multirow{2}{*}{POGARS~\cite{POGARS}}  &     \tablecheck{\CheckmarkBold}    &      & $93.2$   \\  \cline{2-4}\rule{0pt}{1.2EM}  
  &     \tablecheck{\CheckmarkBold}    &   \tablecheck{\CheckmarkBold}     & $93.9$  \\  \hline \rule{0pt}{1.2EM}
\multirow{2}{*}{\ours (ours)}   &  \tablecheck{\CheckmarkBold} &    & $93.7$ \\   \cline{2-4}\rule{0pt}{1.2EM}  
  &  \tablecheck{\CheckmarkBold}  &  \tablecheck{\CheckmarkBold}   &   $\mathbf{94.6}$ \\
\hline
\end{tabular}
\end{minipage}
\end{center}
\end{table}

\setlength{\tabcolsep}{4pt}
\begin{table}[t]
\fontsize{7.5pt}{7.5pt}\selectfont
\begin{center}
\caption{Comparisons with SOTA methods that use \textbf{a single or multiple modalities} on the original split of VD and CAD.
 ``Flow'' denotes optical flow input, and ``Scene'' denotes
 features of the entire frames. Fewer modalities 
indicates a stronger capability of the model itself (\textit{\ul{fewer checks are better}}). 
The top 3 performance scores are highlighted as:
\textbf{\red{First}}, \blue{\textit{Second$^{\ast}$}}, \magenta{\textit{Third}}. 
\ours outperforms the latest GAR methods that use a single modality ($+0.7\%$ improvement on VD and $+2.8\%$ improvement on CAD), and performs favorably compared with methods that exploit multiple expensive modalities
  }
\label{table:vd_cad}
\begin{tabular}{l|c|c|c|c|c|c}
\hline
\multirow{2}{*}{Model} &  \multicolumn{4}{c|}{Modality}  &  \multicolumn{2}{c}{Dataset}     \\  \cline{2-7} 
      &   Keypoint  &  RGB   &   Flow   &    Scene    & VD   & CAD     \\ \hline\hline  \rowcolor{aureolin!10}
HDTM~\cite{ibrahim2016hierarchical}     &    &   \tablecheck{\CheckmarkBold}   &      &        &  $81.9$     &  $81.5$      \\ \hline \rowcolor{aureolin!10}
CERN~\cite{cern} &     &    \tablecheck{\CheckmarkBold}    &      &        &  $83.3$   & $87.2$  \\ \hline \rowcolor{aureolin!10}
stagNet~\cite{stagnet}  &     &     \tablecheck{\CheckmarkBold}   &      &        &  $89.3$   & $89.1$   \\ \hline \rowcolor{aureolin!10}
RCRG~\cite{rcrg}   &     &     \tablecheck{\CheckmarkBold}   &      &        &  $89.5$   &  N/A    \\ \hline \rowcolor{aureolin!10}
SSU~\cite{ssu}    &     &     \tablecheck{\CheckmarkBold}   &      &        &   $90.6$   &  N/A    \\ \hline \rowcolor{aureolin!10}
PRL~\cite{prl}     &     &     \tablecheck{\CheckmarkBold}   &      &        &    $91.4$   &  N/A    \\ \hline \rowcolor{aureolin!10}
ARG~\cite{arg}  &     &    \tablecheck{\CheckmarkBold}  &      &        &  $92.5$   &  $91.0$  \\ \hline \rowcolor{aureolin!10}
HiGCIN~\cite{higcin} &     &   \tablecheck{\CheckmarkBold}  &      &        &  $91.5$   &  $93.4$   \\ \hline \rowcolor{aureolin!10}
DIN~\cite{yuan2021spatio} &     &  \tablecheck{\CheckmarkBold}    &      &        &   $93.6$  & N/A   \\ \hline\hline \rowcolor{ballblue!10}
Zappardino~\etal~\cite{zappardino2021learning}  &   \tablecheck{\CheckmarkBold}    &     &      &        &  $91.0$   &  N/A   \\ \hline \rowcolor{ballblue!10} 
GIRN~\cite{GIRN}  &    \tablecheck{\CheckmarkBold}   &     &      &        &  $92.2$   &   N/A  \\ \hline \rowcolor{ballblue!10}
AT~\cite{actor-transformer}  &    \tablecheck{\CheckmarkBold}   &     &      &        &  $92.3$   &   N/A    \\ \hline \rowcolor{ballblue!10}
POGARS~\cite{POGARS}  &   \tablecheck{\CheckmarkBold}    &     &      &        &   $93.9$    &   N/A     \\ \hline\hline 
CRM~\cite{crm}  &     &   \tablecheck{\CheckmarkBold}     &     \tablecheck{\CheckmarkBold}    &        &   $93.0$  & $85.8$  \\ \hline  
AT~\cite{actor-transformer} &            &   \tablecheck{\CheckmarkBold}   &    \tablecheck{\CheckmarkBold}     &    &  $93.0$      &  $92.8$   \\ \hline   
Ehsanpour~\etal~\cite{ehsanpour2020joint}   &     &   \tablecheck{\CheckmarkBold}  &      &     \tablecheck{\CheckmarkBold}     & $93.1$    &  $89.4$  \\ \hline  
GIRN~\cite{GIRN}     &  \tablecheck{\CheckmarkBold}  &    \tablecheck{\CheckmarkBold}   &   \tablecheck{\CheckmarkBold}     &        &  $94.0$   &  N/A   \\ \hline   
TCE+STBiP~\cite{yuan2021learning}    &      \tablecheck{\CheckmarkBold}       &   \tablecheck{\CheckmarkBold}   &      &    \tablecheck{\CheckmarkBold}     &   \magenta{\textit{94.7}}  &   N/A    \\ \hline  
SACRF~\cite{pramono2020empowering}   &  \tablecheck{\CheckmarkBold} &  \tablecheck{\CheckmarkBold}    &   \tablecheck{\CheckmarkBold}    &     \tablecheck{\CheckmarkBold}    &   \blue{\textit{95.0$^{\ast}$}}  &  \magenta{\textit{95.2}}    \\ \hline  
GroupFormer~\cite{GroupFormer}   & \tablecheck{\CheckmarkBold}   &   \tablecheck{\CheckmarkBold}    &   \tablecheck{\CheckmarkBold}     &  \tablecheck{\CheckmarkBold}        &  \red{$\mathbf{95.7}$}   & \red{$\mathbf{96.3}$}   \\ \hline \rowcolor{ballblue!10}
\ours (ours)    &     \tablecheck{\CheckmarkBold}        &     &      &        &  $94.6$   & \blue{\textit{96.2$^{\ast}$}}    \\ \hline 
\end{tabular}
\end{center}
\scriptsize{*Note: The best results of each method that were reported by the method authors are listed in the table in order to be compared with ours most rigidly. `N/A' stands for `not available'. Yellow shaded rows highlight that the methods use just the RGB-based input, whereas blue for just keypoint.  }
\end{table}
\setlength{\tabcolsep}{1.4pt}

\begin{figure*}[t]
	\centering
	\includegraphics[scale=0.165]{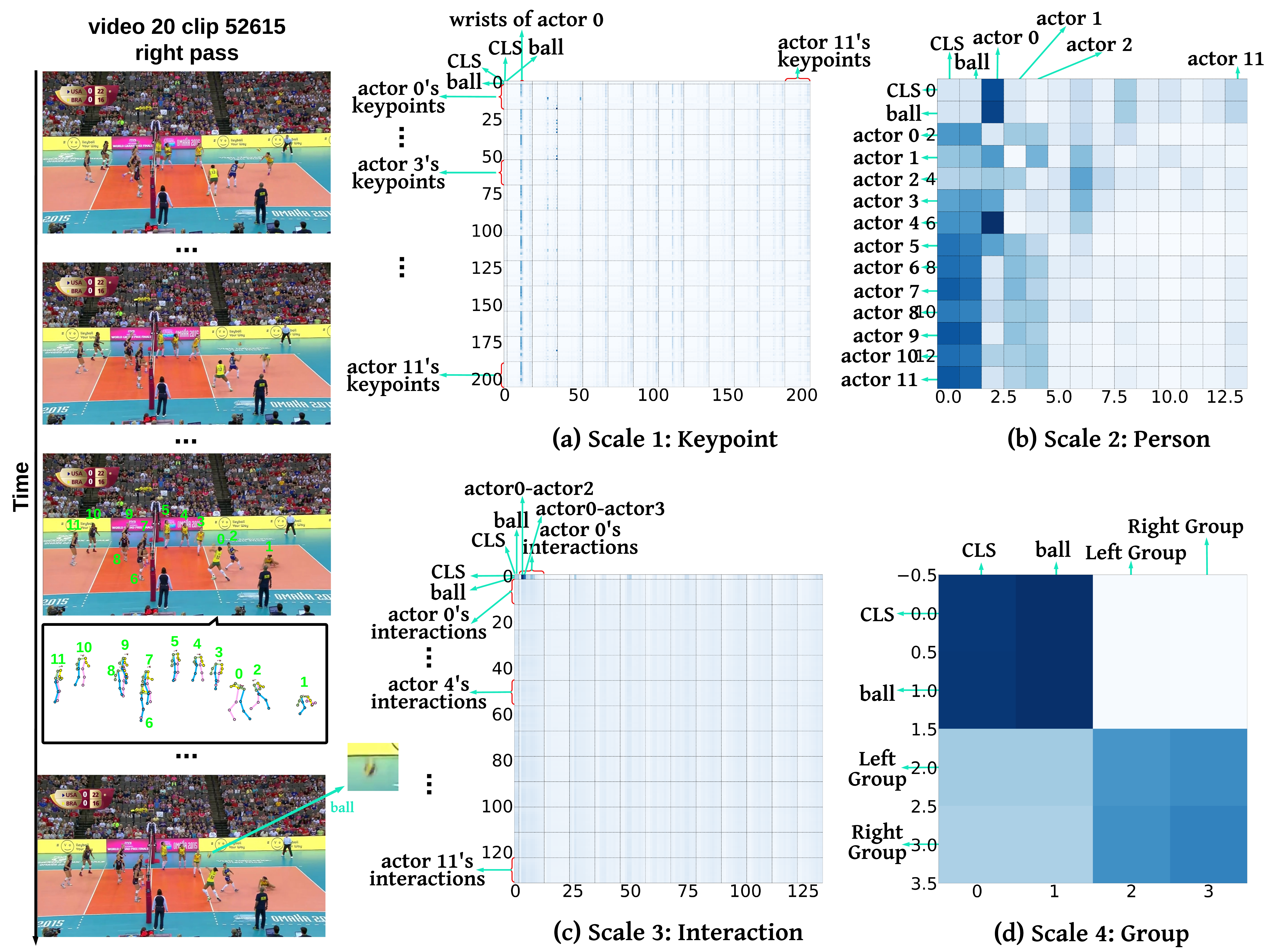}
	\caption{\textbf{Qualitative results of \ours on VD} -- showcasing attention matrices
	of an instance in the ``\textbf{right pass}'' class (key actor is actor $0$).}
	\label{fig:VD_quali}
\end{figure*}

\subsubsection{Scene Generalization for Keypoint-only Setup}
To support the keypoint-only setup for GAR,
we first compare the generalization capability of models using either RGB or the keypoint modality.
In Table~\ref{table:vd_olympic}, 
I3D 
and VGG-16 are two commonly-used image backbone by prior RGB-based GAR methods; the rest are all GAR models (all use VGG-16 as the backbone).

On VD Olympic split, the best prior RGB-based method is DIN~\cite{yuan2021spatio} in Table~\ref{table:vd_olympic}. 
We substitute DIN with a \ours variant~\footnote{This \ours variant consumes RGB-based ROI-aligned person features as input, and thus only models $3$ scales: person, interaction, and the group scale.} (Sec.~\ref{sec:intro}) that also consumes RGB input instead of keypoint, and the result is $81.1\%$ 
which is $2\%$ higher than DIN, suggesting the stronger reasoning strength of \ourseos,
but the accuracy is still low due to the RGB signals. 
POGARS~\cite{POGARS} uses the keypoint modality and has an accuracy of $89.7\%$, higher than all RGB-based methods. 
\ours with the  keypoint-only modality obtains $95.1\%$ accuracy and \textit{significantly outperforms} prior methods, yielding $\mathbf{+5.4\%}$ improvement.
These results imply that the keypoint-only setup 
can reduce scene biases, and generalize better than approaches relying on the RGB modality to testing data with different visual characteristics from training.

We also report the results of these methods that we obtained on VD Original split in Table~\ref{table:vd_olympic}. From this side-by-side comparison, the difference between the Olympic and Original split is vivid. Current GAR methods have quite saturated performances on the Original split of VD and the results are all very high (more evidence later).
Therefore, we recommend readers using the more challenging VD Olympic split
for future research on GAR. 
Note that the \ours that outperforms prior methods in Table~\ref{table:vd_olympic} is only an ablated version of ours in that not using the object token(s). In addition, GroupFormer~\cite{GroupFormer}  is  currently the best-performing method (Table~\ref{table:VD_compare_sota_detailed} in Appendix) and its RGB-only variant has the result of $94.1\%$ accuracy on VD Original split. However, GroupFormer uses additional scene features with the Inception-v3 backbone.

\subsubsection{Comparisons of Methods Using Keypoint-only Modality}
In Table~\ref{table:vd_keypoint}, we compare \ours with more GAR methods that use
only the keypoint modality
on VD Original split following conventions. 
\ours achieves a new SOTA
$94.6\%$ accuracy with $\mathbf{+0.7\%}$ improvement.

Among these methods, Zappardino~\etal~\cite{zappardino2021learning} use CNNs to learn group activity in Volleyball games,
given sequence of person keypoint coordinates, their temporal differences, and keypoint differences from each actor to the pivot-actor that is selected by the model. The model does not model human-object interactions.
AT~\cite{actor-transformer} does not consider human-object interactions either, but because AT is also a Transformer-based  model like ours, we can easily improve it by feeding our object tokens as additional 
inputs to AT.
Moreoever, GIRN~\cite{GIRN} and POGARS~\cite{POGARS} are designed to leverage ball trajectory for learning group activity in videos of Volleyball games.
As shown in Table~\ref{table:vd_keypoint},  the object keypoint information can greatly boost the performance 
by providing additional context.
GIRN models interactions between joints within an actor and across actors, as well as joint-object interactions.
POGARS uses $1$D CNNs to learn spatiotemporal dynamics of actors. 
AT, GIRN, and POGARS all use dot-product-based attention mechanisms similar to ours, however, they fail to fully model the hierarchical entities in the video (e.g., they all only use attention to learn person-wise importance, and at most consider two scales: keypoint and person), 
and more importantly, they lack explicit strategy 
to improve the multiscale representations in order to aid the compositional reasoning of group activity recognition.

\subsubsection{Comparisons of Methods Using Other Modalities}
We compare results of \ours with the best reported results of SOTA methods that use a single or multiple modalities in Table~\ref{table:vd_cad} on both VD and CAD.
\ours still achieves competitive performance --
outperforming methods that use only RGB signals, obtaining $+0.7\%$ improvement on VD and $+2.8\%$ improvement on CAD if compared with methods that use a single modality (RGB or keypoint),
and performing favorably compared with methods that exploit multiple expensive input modalities.

GroupFormer~\cite{GroupFormer} has the highest accuracy on VD and CAD due to learning the representations of the multiscale scene entities (person and person group) with a Clustered Spatial-Temporal Transformer, and leveraging scene context and multiple \textit{expensive} modalities (FLOPs: GroupFormer $\mathbf{595}$\textbf{M} v.s.  \ours $\mathbf{297}$\textbf{M}; details are in Appendix~\ref{appendix_sec:efficiency_compare}).

\begin{figure*}
\centering
\includegraphics[scale=0.073]{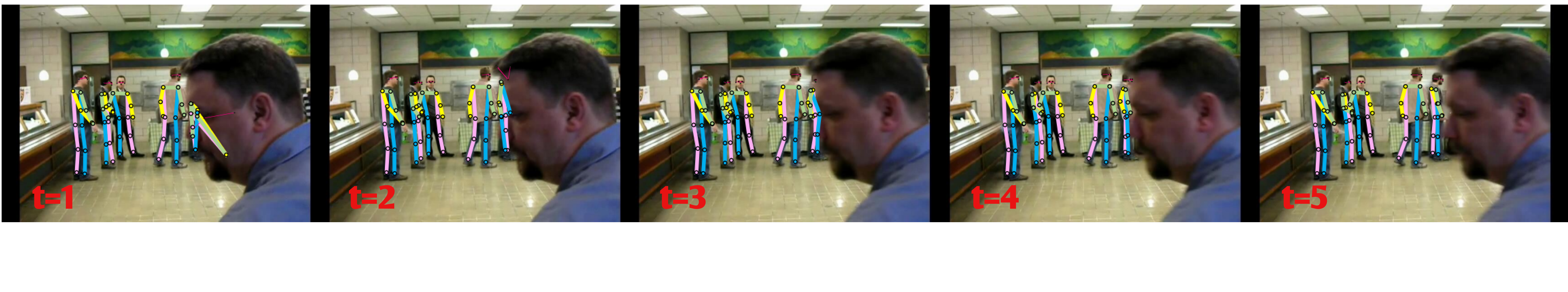}
\caption{\textbf{Qualitative results on CAD} (video ID `10'). \ours successfully predicts `Queueing' even when the input keypoints are partially noisy due to occlusion.}
\label{fig:CAD_quali}
\end{figure*}

\subsection{Qualitative Results}
\label{section:quali}  

We visualize the attention weights in Fig.~\ref{fig:VD_quali}. We highlight the tokens that the model has mostly attended to at each scale (e.g., wrists of actor $0$ at the person keypoint scale). \ours is able to attend to relevant information across different scales, and it can produce interpretable results. 
In Fig.~\ref{fig:CAD_quali}, we visualize the keypoint input to \ours on a CAD instance. 
\ours implicitly learns the human motion patterns from the keypoint features to handle partial occlusions.

Please  check \APP for more analyses including ablation studies, confusion matrices, parameter sensitivity analyses w.r.t. the number of scales and the number of prototypes, more qualitative results including failure cases, etc.

\section{Conclusion}
\label{sec:conclusion}

We propose \ours that uses a \mtx to learn compositional reasoning at different scales for group activity recognition. 
We also improve the intermediate representations using contrastive clustering, auxiliary prediction, and data augmentation techniques.
We demonstrate the model's strength and interpretability on two widely-used datasets (Volleyball and Collective Activity).  
\ours achieves up to $+5.4\%$ improvement with just the keypoint modality. 

One limitation is that videos with severe occlusions remain
challenging for \ours like other existing methods, due to errors from detecting keypoints. Adopting $3$D keypoints or stronger backbones that estimate keypoints directly from the video~\cite{pavllo20193d,mediapipeBlazePose} can help to address
the issue. 
Possible future directions include 1)
expanding our methods to more complex scenarios, such as crowd understanding that may require modeling additional hierarchical scales;
and 2) exploring
effective 
multimodal fusion methods in order to use additional modalities like RGB but without suffering from scene biases, 
since RGB can be beneficial for activities that involve significant interaction with the background scene.

\textbf{Acknowledgments}
The research was supported in part by NSF awards: IIS-1703883, IIS-1955404, IIS-1955365, RETTL-2119265, and EAGER-2122119.
This material is based upon work supported by the U.S. Department of Homeland Security under Grant Award Number 22STESE00001 01 01.
Disclaimer: The views and conclusions contained in this document are those of the authors and should not be interpreted as necessarily representing the official policies, either expressed or implied, of the U.S. Department of Homeland Security.

\bibliographystyle{splncs04}
\bibliography{egbib}

\begin{thebibliography}{100}
\providecommand{\url}[1]{\texttt{#1}}
\providecommand{\urlprefix}{URL }
\providecommand{\doi}[1]{https://doi.org/#1}

\bibitem{balltracktool}
Ball recognition and tracking in live volleyball game.
  \url{https://github.com/tprlab/vball}

\bibitem{mediapipeBlazePose}
Mediapipe pose: Ml solution for high-fidelity body pose tracking from rgb video
  frames. \url{https://google.github.io/mediapipe/solutions/pose.html}

\bibitem{abkenar2019groupsense}
Abkenar, A.B., Loke, S.W., Zaslavsky, A., Rahayu, W.: Groupsense: recognizing
  and understanding group physical activities using multi-device embedded
  sensing. ACM Transactions on Embedded Computing Systems (TECS)
  \textbf{17}(6),  1--26 (2019)

\bibitem{asano2020labelling}
Asano, Y.M., Patrick, M., Rupprecht, C., Vedaldi, A.: Labelling unlabelled
  videos from scratch with multi-modal self-supervision. arXiv preprint
  arXiv:2006.13662  (2020)

\bibitem{crm}
Azar, S.M., Atigh, M.G., Nickabadi, A., Alahi, A.: Convolutional relational
  machine for group activity recognition. In: Proceedings of the IEEE/CVF
  Conference on Computer Vision and Pattern Recognition. pp. 7892--7901 (2019)

\bibitem{azar2019convolutional}
Azar, S.M., Atigh, M.G., Nickabadi, A., Alahi, A.: Convolutional relational
  machine for group activity recognition. In: Proceedings of the IEEE/CVF
  Conference on Computer Vision and Pattern Recognition. pp. 7892--7901 (2019)

\bibitem{ba2016layer}
Ba, J.L., Kiros, J.R., Hinton, G.E.: Layer normalization. arXiv preprint
  arXiv:1607.06450  (2016)

\bibitem{ssu}
Bagautdinov, T., Alahi, A., Fleuret, F., Fua, P., Savarese, S.: Social scene
  understanding: End-to-end multi-person action localization and collective
  activity recognition. In: Proceedings of the IEEE conference on computer
  vision and pattern recognition. pp. 4315--4324 (2017)

\bibitem{blomqvist2022semi}
Blomqvist, K., Chung, J.J., Ott, L., Siegwart, R.: Semi-automatic 3d object
  keypoint annotation and detection for the masses. arXiv preprint
  arXiv:2201.07665  (2022)

\bibitem{bottou2014machine}
Bottou, L.: From machine learning to machine reasoning. Machine learning
  \textbf{94}(2),  133--149 (2014)

\bibitem{swav}
Caron, M., Misra, I., Mairal, J., Goyal, P., Bojanowski, P., Joulin, A.:
  Unsupervised learning of visual features by contrasting cluster assignments.
  In: Thirty-fourth Conference on Neural Information Processing Systems
  (NeurIPS) (2020)

\bibitem{carreira2017quo}
Carreira, J., Zisserman, A.: Quo vadis, action recognition? a new model and the
  kinetics dataset. In: proceedings of the IEEE Conference on Computer Vision
  and Pattern Recognition. pp. 6299--6308 (2017)

\bibitem{chen2021multimodal}
Chen, B., Rouditchenko, A., Duarte, K., Kuehne, H., Thomas, S., Boggust, A.,
  Panda, R., Kingsbury, B., Feris, R., Harwath, D., et~al.: Multimodal
  clustering networks for self-supervised learning from unlabeled videos. arXiv
  preprint arXiv:2104.12671  (2021)

\bibitem{chen2019group}
Chen, H.Y., Lai, S.H.: Group activity recognition via computing human pose
  motion history and collective map from video. In: Asian Conference on Pattern
  Recognition. pp. 705--718. Springer (2019)

\bibitem{chen2020rit}
Chen, J., Hao, H., Hong, H., Kong, Y.: Rit-18: A novel dataset for
  compositional group activity understanding. In: Proceedings of the IEEE/CVF
  Conference on Computer Vision and Pattern Recognition Workshops. pp. 362--363
  (2020)

\bibitem{cheng2010group}
Cheng, Z., Qin, L., Huang, Q., Jiang, S., Tian, Q.: Group activity recognition
  by gaussian processes estimation. In: 2010 20th International Conference on
  Pattern Recognition. pp. 3228--3231. IEEE (2010)

\bibitem{scenebias}
Choi, J., Gao, C., Messou, J.C., Huang, J.B.: Why can't i dance in a mall?
  learning to mitigate scene bias in action recognition. In: Proceedings of the
  33rd International Conference on Neural Information Processing Systems. pp.
  853--865 (2019)

\bibitem{choi2012unified}
Choi, W., Savarese, S.: A unified framework for multi-target tracking and
  collective activity recognition. In: European Conference on Computer Vision.
  pp. 215--230. Springer (2012)

\bibitem{choi2013understanding}
Choi, W., Savarese, S.: Understanding collective activitiesof people from
  videos. IEEE transactions on pattern analysis and machine intelligence
  \textbf{36}(6),  1242--1257 (2013)

\bibitem{choi2009they}
Choi, W., Shahid, K., Savarese, S.: What are they doing?: Collective activity
  classification using spatio-temporal relationship among people. In: 2009 IEEE
  12th international conference on computer vision workshops, ICCV Workshops.
  pp. 1282--1289. IEEE (2009)

\bibitem{choi2011learning}
Choi, W., Shahid, K., Savarese, S.: Learning context for collective activity
  recognition. In: CVPR 2011. pp. 3273--3280. IEEE (2011)

\bibitem{cuturi2013sinkhorn}
Cuturi, M.: Sinkhorn distances: Lightspeed computation of optimal transport.
  Advances in neural information processing systems  \textbf{26},  2292--2300
  (2013)

\bibitem{dang2020sensor}
Dang, L.M., Min, K., Wang, H., Piran, M.J., Lee, C.H., Moon, H.: Sensor-based
  and vision-based human activity recognition: A comprehensive survey. Pattern
  Recognition  \textbf{108},  107561 (2020)

\bibitem{dankers2021paradox}
Dankers, V., Bruni, E., Hupkes, D.: The paradox of the compositionality of
  natural language: a neural machine translation case study. arXiv preprint
  arXiv:2108.05885  (2021)

\bibitem{deng2016structure}
Deng, Z., Vahdat, A., Hu, H., Mori, G.: Structure inference machines: Recurrent
  neural networks for analyzing relations in group activity recognition. In:
  Proceedings of the IEEE Conference on Computer Vision and Pattern
  Recognition. pp. 4772--4781 (2016)

\bibitem{devlin2018bert}
Devlin, J., Chang, M.W., Lee, K., Toutanova, K.: Bert: Pre-training of deep
  bidirectional transformers for language understanding. arXiv preprint
  arXiv:1810.04805  (2018)

\bibitem{ehsanpour2020joint}
Ehsanpour, M., Abedin, A., Saleh, F., Shi, J., Reid, I., Rezatofighi, H.: Joint
  learning of social groups, individuals action and sub-group activities in
  videos. In: Computer Vision--ECCV 2020: 16th European Conference, Glasgow,
  UK, August 23--28, 2020, Proceedings, Part IX 16. pp. 177--195. Springer
  (2020)

\bibitem{fan2021multiscale}
Fan, H., Xiong, B., Mangalam, K., Li, Y., Yan, Z., Malik, J., Feichtenhofer,
  C.: Multiscale vision transformers. arXiv preprint arXiv:2104.11227  (2021)

\bibitem{actor-transformer}
Gavrilyuk, K., Sanford, R., Javan, M., Snoek, C.G.: Actor-transformers for
  group activity recognition. In: Proceedings of the IEEE/CVF Conference on
  Computer Vision and Pattern Recognition. pp. 839--848 (2020)

\bibitem{glorot2011deep}
Glorot, X., Bordes, A., Bengio, Y.: Deep sparse rectifier neural networks. In:
  Proceedings of the fourteenth international conference on artificial
  intelligence and statistics. pp. 315--323. JMLR Workshop and Conference
  Proceedings (2011)

\bibitem{gong2019cnn}
Gong, Z., Zhong, P., Yu, Y., Hu, W., Li, S.: A cnn with multiscale convolution
  and diversified metric for hyperspectral image classification. IEEE
  Transactions on Geoscience and Remote Sensing  \textbf{57}(6),  3599--3618
  (2019)

\bibitem{grunde2021agqa}
Grunde-McLaughlin, M., Krishna, R., Agrawala, M.: Agqa: A benchmark for
  compositional spatio-temporal reasoning. In: Proceedings of the IEEE/CVF
  Conference on Computer Vision and Pattern Recognition. pp. 11287--11297
  (2021)

\bibitem{haber2018learning}
Haber, E., Ruthotto, L., Holtham, E., Jun, S.H.: Learning across
  scales---multiscale methods for convolution neural networks. In:
  Thirty-Second AAAI Conference on Artificial Intelligence (2018)

\bibitem{hajimirsadeghi2015visual}
Hajimirsadeghi, H., Yan, W., Vahdat, A., Mori, G.: Visual recognition by
  counting instances: A multi-instance cardinality potential kernel. In:
  Proceedings of the IEEE conference on computer vision and pattern
  recognition. pp. 2596--2605 (2015)

\bibitem{han2021transformer}
Han, K., Xiao, A., Wu, E., Guo, J., Xu, C., Wang, Y.: Transformer in
  transformer. arXiv preprint arXiv:2103.00112  (2021)

\bibitem{he2017mask}
He, K., Gkioxari, G., Doll{\'a}r, P., Girshick, R.: Mask r-cnn. In: Proceedings
  of the IEEE international conference on computer vision. pp. 2961--2969
  (2017)

\bibitem{he2016deep}
He, K., Zhang, X., Ren, S., Sun, J.: Deep residual learning for image
  recognition. In: Proceedings of the IEEE conference on computer vision and
  pattern recognition. pp. 770--778 (2016)

\bibitem{hendrycks2016gaussian}
Hendrycks, D., Gimpel, K.: Gaussian error linear units (gelus). arXiv preprint
  arXiv:1606.08415  (2016)

\bibitem{hinojosa2021learning}
Hinojosa, C., Niebles, J.C., Arguello, H.: Learning privacy-preserving optics
  for human pose estimation. In: Proceedings of the IEEE/CVF International
  Conference on Computer Vision. pp. 2573--2582 (2021)

\bibitem{hochreiter1997long}
Hochreiter, S., Schmidhuber, J.: Long short-term memory. Neural computation
  \textbf{9}(8),  1735--1780 (1997)

\bibitem{prl}
Hu, G., Cui, B., He, Y., Yu, S.: Progressive relation learning for group
  activity recognition. In: Proceedings of the IEEE/CVF Conference on Computer
  Vision and Pattern Recognition. pp. 980--989 (2020)

\bibitem{huangyi}
Huang, Y., Kadav, A., Lai, F., Patel, D., Graf, H.P.: Learning higher-order
  object interactions for keypoint-based video understanding  (2021)

\bibitem{hudson2019learning}
Hudson, D., Manning, C.D.: Learning by abstraction: The neural state machine.
  Advances in Neural Information Processing Systems  \textbf{32},  5903--5916
  (2019)

\bibitem{hudson2018compositional}
Hudson, D.A., Manning, C.D.: Compositional attention networks for machine
  reasoning. In: International Conference on Learning Representations (2018)

\bibitem{rcrg}
Ibrahim, M.S., Mori, G.: Hierarchical relational networks for group activity
  recognition and retrieval. In: Proceedings of the European conference on
  computer vision (ECCV). pp. 721--736 (2018)

\bibitem{ibrahim2018hierarchical}
Ibrahim, M.S., Mori, G.: Hierarchical relational networks for group activity
  recognition and retrieval. In: Proceedings of the European conference on
  computer vision (ECCV). pp. 721--736 (2018)

\bibitem{ibrahim2016hierarchical}
Ibrahim, M.S., Muralidharan, S., Deng, Z., Vahdat, A., Mori, G.: A hierarchical
  deep temporal model for group activity recognition. In: Proceedings of the
  IEEE Conference on Computer Vision and Pattern Recognition. pp. 1971--1980
  (2016)

\bibitem{irsoy2014deep}
Irsoy, O., Cardie, C.: Deep recursive neural networks for compositionality in
  language. In: Advances in neural information processing systems. pp.
  2096--2104 (2014)

\bibitem{jaiswal2021keypoints}
Jaiswal, A., Singh, S., Wu, Y., Natarajan, P., Natarajan, P.: Keypoints-aware
  object detection. In: NeurIPS 2020 Workshop on Pre-registration in Machine
  Learning. pp. 62--72. PMLR (2021)

\bibitem{ji2020action}
Ji, J., Krishna, R., Fei-Fei, L., Niebles, J.C.: Action genome: Actions as
  compositions of spatio-temporal scene graphs. In: Proceedings of the IEEE/CVF
  Conference on Computer Vision and Pattern Recognition. pp. 10236--10247
  (2020)

\bibitem{jia2020lemma}
Jia, B., Chen, Y., Huang, S., Zhu, Y., Zhu, S.c.: Lemma: A multi-view dataset
  for learning multi-agent multi-task activities. In: European Conference on
  Computer Vision. pp. 767--786. Springer (2020)

\bibitem{khosla2020supervised}
Khosla, P., Teterwak, P., Wang, C., Sarna, A., Tian, Y., Isola, P., Maschinot,
  A., Liu, C., Krishnan, D.: Supervised contrastive learning. arXiv preprint
  arXiv:2004.11362  (2020)

\bibitem{kim2018discriminative}
Kim, P.S., Lee, D.G., Lee, S.W.: Discriminative context learning with gated
  recurrent unit for group activity recognition. Pattern Recognition
  \textbf{76},  149--161 (2018)

\bibitem{kim2020safcar}
Kim, T.S., Hager, G.D.: Safcar: Structured attention fusion for compositional
  action recognition. arXiv preprint arXiv:2012.02109  (2020)

\bibitem{kingma2014adam}
Kingma, D.P., Ba, J.: Adam: A method for stochastic optimization. arXiv
  preprint arXiv:1412.6980  (2014)

\bibitem{kipf2016semi}
Kipf, T.N., Welling, M.: Semi-supervised classification with graph
  convolutional networks. arXiv preprint arXiv:1609.02907  (2016)

\bibitem{kong2012learning}
Kong, Y., Jia, Y., Fu, Y.: Learning human interaction by interactive phrases.
  In: European conference on computer vision. pp. 300--313. Springer (2012)

\bibitem{koshkina2021contrastive}
Koshkina, M., Pidaparthy, H., Elder, J.H.: Contrastive learning for sports
  video: Unsupervised player classification. In: Proceedings of the IEEE/CVF
  Conference on Computer Vision and Pattern Recognition. pp. 4528--4536 (2021)

\bibitem{kulkarni2019unsupervised}
Kulkarni, T.D., Gupta, A., Ionescu, C., Borgeaud, S., Reynolds, M., Zisserman,
  A., Mnih, V.: Unsupervised learning of object keypoints for perception and
  control. Advances in neural information processing systems  \textbf{32}
  (2019)

\bibitem{lake2017building}
Lake, B.M., Ullman, T.D., Tenenbaum, J.B., Gershman, S.J.: Building machines
  that learn and think like people. Behavioral and brain sciences  \textbf{40}
  (2017)

\bibitem{lan2011discriminative}
Lan, T., Wang, Y., Yang, W., Robinovitch, S.N., Mori, G.: Discriminative latent
  models for recognizing contextual group activities. IEEE transactions on
  pattern analysis and machine intelligence  \textbf{34}(8),  1549--1562 (2011)

\bibitem{le2021c}
Le, H., Chen, N.F., Hoi, S.C.: C$^3$: Compositional counterfactual constrastive
  learning for video-grounded dialogues. arXiv preprint arXiv:2106.08914
  (2021)

\bibitem{li2016visual}
Li, G., Yu, Y.: Visual saliency detection based on multiscale deep cnn
  features. IEEE transactions on image processing  \textbf{25}(11),  5012--5024
  (2016)

\bibitem{li2020dynamic}
Li, M., Chen, S., Zhao, Y., Zhang, Y., Wang, Y., Tian, Q.: Dynamic multiscale
  graph neural networks for 3d skeleton based human motion prediction. In:
  Proceedings of the IEEE/CVF Conference on Computer Vision and Pattern
  Recognition. pp. 214--223 (2020)

\bibitem{GroupFormer}
Li, S., Cao, Q., Liu, L., Yang, K., Liu, S., Hou, J., Yi, S.: Groupformer:
  Group activity recognition with clustered spatial-temporal transformer. In:
  Proceedings of the IEEE/CVF International Conference on Computer Vision. pp.
  13668--13677 (2021)

\bibitem{li2017sbgar}
Li, X., Choo~Chuah, M.: Sbgar: Semantics based group activity recognition. In:
  Proceedings of the IEEE international conference on computer vision. pp.
  2876--2885 (2017)

\bibitem{lin2020ms2l}
Lin, L., Song, S., Yang, W., Liu, J.: Ms2l: Multi-task self-supervised learning
  for skeleton based action recognition. In: Proceedings of the 28th ACM
  International Conference on Multimedia. pp. 2490--2498 (2020)

\bibitem{liu2021swin}
Liu, Z., Lin, Y., Cao, Y., Hu, H., Wei, Y., Zhang, Z., Lin, S., Guo, B.: Swin
  transformer: Hierarchical vision transformer using shifted windows. arXiv
  preprint arXiv:2103.14030  (2021)

\bibitem{lu2021few}
Lu, C., Koniusz, P.: Few-shot keypoint detection with uncertainty learning for
  unseen species. arXiv preprint arXiv:2112.06183  (2021)

\bibitem{lu2019spatio}
Lu, L., Di, H., Lu, Y., Zhang, L., Wang, S.: Spatio-temporal attention
  mechanisms based model for collective activity recognition. Signal
  Processing: Image Communication  \textbf{74},  162--174 (2019)

\bibitem{lu2019gaim}
Lu, L., Lu, Y., Yu, R., Di, H., Zhang, L., Wang, S.: Gaim: Graph attention
  interaction model for collective activity recognition. IEEE Transactions on
  Multimedia  \textbf{22}(2),  524--539 (2019)

\bibitem{luo2021moma}
Luo, Z., Xie, W., Kapoor, S., Liang, Y., Cooper, M., Niebles, J.C., Adeli, E.,
  Li, F.F.: Moma: Multi-object multi-actor activity parsing. Advances in Neural
  Information Processing Systems  \textbf{34} (2021)

\bibitem{materzynska2020something}
Materzynska, J., Xiao, T., Herzig, R., Xu, H., Wang, X., Darrell, T.:
  Something-else: Compositional action recognition with spatial-temporal
  interaction networks. In: Proceedings of the IEEE/CVF Conference on Computer
  Vision and Pattern Recognition. pp. 1049--1059 (2020)

\bibitem{nabi2013temporal}
Nabi, M., Bue, A., Murino, V.: Temporal poselets for collective activity
  detection and recognition. In: Proceedings of the IEEE International
  Conference on Computer Vision Workshops. pp. 500--507 (2013)

\bibitem{nakatani2021group}
Nakatani, C., Sendo, K., Ukita, N.: Group activity recognition using joint
  learning of individual action recognition and people grouping. In: 2021 17th
  International Conference on Machine Vision and Applications (MVA). pp.~1--5.
  IEEE (2021)

\bibitem{nebehay2014consensus}
Nebehay, G., Pflugfelder, R.: Consensus-based matching and tracking of
  keypoints for object tracking. In: IEEE Winter Conference on Applications of
  Computer Vision. pp. 862--869. IEEE (2014)

\bibitem{ngiam2021scene}
Ngiam, J., Caine, B., Vasudevan, V., Zhang, Z., Chiang, H.T.L., Ling, J.,
  Roelofs, R., Bewley, A., Liu, C., Venugopal, A., et~al.: Scene transformer: A
  unified architecture for predicting multiple agent trajectories. arXiv
  preprint arXiv:2106.08417  (2021)

\bibitem{nguyen2021geomnet}
Nguyen, X.S.: Geomnet: A neural network based on riemannian geometries of spd
  matrix space and cholesky space for 3d skeleton-based interaction
  recognition. In: Proceedings of the IEEE/CVF International Conference on
  Computer Vision. pp. 13379--13389 (2021)

\bibitem{patrick2021keeping}
Patrick, M., Campbell, D., Asano, Y., Misra, I., Metze, F., Feichtenhofer, C.,
  Vedaldi, A., Henriques, J.F.: Keeping your eye on the ball: Trajectory
  attention in video transformers. Advances in Neural Information Processing
  Systems  \textbf{34} (2021)

\bibitem{pavllo20193d}
Pavllo, D., Feichtenhofer, C., Grangier, D., Auli, M.: 3d human pose estimation
  in video with temporal convolutions and semi-supervised training. In:
  Proceedings of the IEEE/CVF Conference on Computer Vision and Pattern
  Recognition. pp. 7753--7762 (2019)

\bibitem{perez2021interaction}
Perez, M., Liu, J., Kot, A.C.: Interaction relational network for mutual action
  recognition. IEEE Transactions on Multimedia  (2021)

\bibitem{GIRN}
Perez, M., Liu, J., Kot, A.C.: Skeleton-based relational reasoning for group
  activity analysis. Pattern Recognition p. 108360 (2021)

\bibitem{pramono2020empowering}
Pramono, R.R.A., Chen, Y.T., Fang, W.H.: Empowering relational network by
  self-attention augmented conditional random fields for group activity
  recognition. In: European Conference on Computer Vision. pp. 71--90. Springer
  (2020)

\bibitem{stagnet}
Qi, M., Qin, J., Li, A., Wang, Y., Luo, J., Van~Gool, L.: stagnet: An attentive
  semantic rnn for group activity recognition. In: Proceedings of the European
  Conference on Computer Vision (ECCV). pp. 101--117 (2018)

\bibitem{rai2021home}
Rai, N., Chen, H., Ji, J., Desai, R., Kozuka, K., Ishizaka, S., Adeli, E.,
  Niebles, J.C.: Home action genome: Cooperative compositional action
  understanding. In: Proceedings of the IEEE/CVF Conference on Computer Vision
  and Pattern Recognition. pp. 11184--11193 (2021)

\bibitem{ramanathan2016detecting}
Ramanathan, V., Huang, J., Abu-El-Haija, S., Gorban, A., Murphy, K., Fei-Fei,
  L.: Detecting events and key actors in multi-person videos. In: Proceedings
  of the IEEE conference on computer vision and pattern recognition. pp.
  3043--3053 (2016)

\bibitem{sendo2019heatmapping}
Sendo, K., Ukita, N.: Heatmapping of people involved in group activities. In:
  2019 16th International Conference on Machine Vision Applications (MVA).
  pp.~1--6. IEEE (2019)

\bibitem{shao2020finegym}
Shao, D., Zhao, Y., Dai, B., Lin, D.: Finegym: A hierarchical video dataset for
  fine-grained action understanding. In: Proceedings of the IEEE/CVF Conference
  on Computer Vision and Pattern Recognition. pp. 2616--2625 (2020)

\bibitem{shi2020online}
Shi, C., Holtz, C., Mishne, G.: Online adversarial purification based on
  self-supervised learning. In: International Conference on Learning
  Representations (2020)

\bibitem{cern}
Shu, T., Todorovic, S., Zhu, S.C.: Cern: confidence-energy recurrent network
  for group activity recognition. In: Proceedings of the IEEE conference on
  computer vision and pattern recognition. pp. 5523--5531 (2017)

\bibitem{shu2017cern}
Shu, T., Todorovic, S., Zhu, S.C.: Cern: confidence-energy recurrent network
  for group activity recognition. In: Proceedings of the IEEE conference on
  computer vision and pattern recognition. pp. 5523--5531 (2017)

\bibitem{shu2019hierarchical}
Shu, X., Tang, J., Qi, G., Liu, W., Yang, J.: Hierarchical long short-term
  concurrent memory for human interaction recognition. IEEE transactions on
  pattern analysis and machine intelligence  (2019)

\bibitem{simonyan2014very}
Simonyan, K., Zisserman, A.: Very deep convolutional networks for large-scale
  image recognition. arXiv preprint arXiv:1409.1556  (2014)

\bibitem{singh2020don}
Singh, K.K., Mahajan, D., Grauman, K., Lee, Y.J., Feiszli, M., Ghadiyaram, D.:
  Don't judge an object by its context: Learning to overcome contextual bias.
  In: Proceedings of the IEEE/CVF Conference on Computer Vision and Pattern
  Recognition. pp. 11070--11078 (2020)

\bibitem{snower202015}
Snower, M., Kadav, A., Lai, F., Graf, H.P.: 15 keypoints is all you need. In:
  Proceedings of the IEEE/CVF Conference on Computer Vision and Pattern
  Recognition. pp. 6738--6748 (2020)

\bibitem{socher2014grounded}
Socher, R., Karpathy, A., Le, Q.V., Manning, C.D., Ng, A.Y.: Grounded
  compositional semantics for finding and describing images with sentences.
  Transactions of the Association for Computational Linguistics  \textbf{2},
  207--218 (2014)

\bibitem{srivastava2014dropout}
Srivastava, N., Hinton, G., Krizhevsky, A., Sutskever, I., Salakhutdinov, R.:
  Dropout: a simple way to prevent neural networks from overfitting. The
  journal of machine learning research  \textbf{15}(1),  1929--1958 (2014)

\bibitem{sun2020view}
Sun, J.J., Zhao, J., Chen, L.C., Schroff, F., Adam, H., Liu, T.: View-invariant
  probabilistic embedding for human pose. In: European Conference on Computer
  Vision. pp. 53--70. Springer (2020)

\bibitem{sun2021counterfactual}
Sun, P., Wu, B., Li, X., Li, W., Duan, L., Gan, C.: Counterfactual debiasing
  inference for compositional action recognition. In: Proceedings of the 29th
  ACM International Conference on Multimedia. pp. 3220--3228 (2021)

\bibitem{suris2021learning}
Sur{\'\i}s, D., Liu, R., Vondrick, C.: Learning the predictability of the
  future. In: Proceedings of the IEEE/CVF Conference on Computer Vision and
  Pattern Recognition. pp. 12607--12617 (2021)

\bibitem{tancik2020fourier}
Tancik, M., Srinivasan, P.P., Mildenhall, B., Fridovich-Keil, S., Raghavan, N.,
  Singhal, U., Ramamoorthi, R., Barron, J.T., Ng, R.: Fourier features let
  networks learn high frequency functions in low dimensional domains. arXiv
  preprint arXiv:2006.10739  (2020)

\bibitem{POGARS}
Thilakarathne, H., Nibali, A., He, Z., Morgan, S.: Pose is all you need: The
  pose only group activity recognition system (pogars). arXiv preprint
  arXiv:2108.04186  (2021)

\bibitem{vaswani2017attention}
Vaswani, A., Shazeer, N., Parmar, N., Uszkoreit, J., Jones, L., Gomez, A.N.,
  Kaiser, {\L}., Polosukhin, I.: Attention is all you need. In: Advances in
  neural information processing systems. pp. 5998--6008 (2017)

\bibitem{vendrov2015order}
Vendrov, I., Kiros, R., Fidler, S., Urtasun, R.: Order-embeddings of images and
  language. arXiv preprint arXiv:1511.06361  (2015)

\bibitem{hrnet}
Wang, J., Sun, K., Cheng, T., Jiang, B., Deng, C., Zhao, Y., Liu, D., Mu, Y.,
  Tan, M., Wang, X., et~al.: Deep high-resolution representation learning for
  visual recognition. IEEE transactions on pattern analysis and machine
  intelligence  (2020)

\bibitem{wang2017recurrent}
Wang, M., Ni, B., Yang, X.: Recurrent modeling of interaction context for
  collective activity recognition. In: Proceedings of the IEEE Conference on
  Computer Vision and Pattern Recognition. pp. 3048--3056 (2017)

\bibitem{arg}
Wu, J., Wang, L., Wang, L., Guo, J., Wu, G.: Learning actor relation graphs for
  group activity recognition. In: Proceedings of the IEEE/CVF Conference on
  Computer Vision and Pattern Recognition. pp. 9964--9974 (2019)

\bibitem{wu2021comprehensive}
Wu, L.F., Wang, Q., Jian, M., Qiao, Y., Zhao, B.X.: A comprehensive review of
  group activity recognition in videos. International Journal of Automation and
  Computing pp. 1--17 (2021)

\bibitem{xu2020group}
Xu, D., Fu, H., Wu, L., Jian, M., Wang, D., Liu, X.: Group activity recognition
  by using effective multiple modality relation representation with
  temporal-spatial attention. IEEE Access  \textbf{8},  65689--65698 (2020)

\bibitem{PCTDM}
Yan, R., Tang, J., Shu, X., Li, Z., Tian, Q.: Participation-contributed
  temporal dynamic model for group activity recognition. In: Proceedings of the
  26th ACM international conference on Multimedia. pp. 1292--1300 (2018)

\bibitem{yan2020interactive}
Yan, R., Xie, L., Shu, X., Tang, J.: Interactive fusion of multi-level features
  for compositional activity recognition. arXiv preprint arXiv:2012.05689
  (2020)

\bibitem{higcin}
Yan, R., Xie, L., Tang, J., Shu, X., Tian, Q.: Higcin: hierarchical graph-based
  cross inference network for group activity recognition. IEEE Transactions on
  Pattern Analysis and Machine Intelligence  (2020)

\bibitem{sam}
Yan, R., Xie, L., Tang, J., Shu, X., Tian, Q.: Social adaptive module for
  weakly-supervised group activity recognition. In: European Conference on
  Computer Vision. pp. 208--224. Springer (2020)

\bibitem{yan2018spatial}
Yan, S., Xiong, Y., Lin, D.: Spatial temporal graph convolutional networks for
  skeleton-based action recognition. In: Thirty-second AAAI conference on
  artificial intelligence (2018)

\bibitem{yang2019reppoints}
Yang, Z., Liu, S., Hu, H., Wang, L., Lin, S.: Reppoints: Point set
  representation for object detection. In: Proceedings of the IEEE/CVF
  International Conference on Computer Vision. pp. 9657--9666 (2019)

\bibitem{yuan2021learning}
Yuan, H., Ni, D.: Learning visual context for group activity recognition. In:
  Proceedings of the AAAI Conference on Artificial Intelligence. vol.~35, pp.
  3261--3269 (2021)

\bibitem{yuan2021spatio}
Yuan, H., Ni, D., Wang, M.: Spatio-temporal dynamic inference network for group
  activity recognition. In: Proceedings of the IEEE/CVF International
  Conference on Computer Vision. pp. 7476--7485 (2021)

\bibitem{yun2012two}
Yun, K., Honorio, J., Chattopadhyay, D., Berg, T.L., Samaras, D.: Two-person
  interaction detection using body-pose features and multiple instance
  learning. In: 2012 IEEE Computer Society Conference on Computer Vision and
  Pattern Recognition Workshops. pp. 28--35. IEEE (2012)

\bibitem{zappardino2021learning}
Zappardino, F., Uricchio, T., Seidenari, L., Del~Bimbo, A.: Learning group
  activities from skeletons without individual action labels. In: 2020 25th
  International Conference on Pattern Recognition (ICPR). pp. 10412--10417.
  IEEE (2021)

\bibitem{zhan2020multi}
Zhan, Y., Yu, J., Yu, T., Tao, D.: Multi-task compositional network for visual
  relationship detection. International Journal of Computer Vision
  \textbf{128}(8),  2146--2165 (2020)

\bibitem{zhao2021learning}
Zhao, L., Wang, Y., Zhao, J., Yuan, L., Sun, J.J., Schroff, F., Adam, H., Peng,
  X., Metaxas, D., Liu, T.: Learning view-disentangled human pose
  representation by contrastive cross-view mutual information maximization. In:
  Proceedings of the IEEE/CVF Conference on Computer Vision and Pattern
  Recognition. pp. 12793--12802 (2021)

\bibitem{zhu2016co}
Zhu, W., Lan, C., Xing, J., Zeng, W., Li, Y., Shen, L., Xie, X.: Co-occurrence
  feature learning for skeleton based action recognition using regularized deep
  lstm networks. In: Proceedings of the AAAI conference on artificial
  intelligence. vol.~30 (2016)

\end{thebibliography}

\clearpage
\title{\ourseos: Compositional Reasoning of Group Activity in Videos with Keypoint-Only Modality Supplementary Material}

\titlerunning{Compositional Reasoning of Group Activity in Videos}
\author{Honglu Zhou\inst{1}
\and
Asim Kadav\inst{2} \and
Aviv Shamsian\inst{3} \and
Shijie Geng\inst{1} \and
Farley Lai\inst{2} \and
Long Zhao\inst{4} \and
Ting Liu\inst{4} \and
Mubbasir Kapadia\inst{1} \and
Hans Peter Graf\inst{2}\index{Graf, Hans Peter}}
\authorrunning{H. Zhou et al.}
\institute{
Department of Computer Science, Rutgers University, Piscataway, NJ, USA\\
\email{\{hz289,sg1309,mk1353\}@cs.rutgers.edu} \and
NEC Laboratories America, Inc., San Jose, CA, USA\\ 
\email{\{asim,farleylai,hpg\}@nec-labs.com} \and
Bar-Ilan University, Israel\\ 
\email{aviv.shamsian@biu.ac.il} \and
Google Research, Los Angeles, CA, USA\\
\email{\{longzh,liuti\}@google.com}
}
\maketitle

\bigskip
\bigskip
\bigskip

\noindent This appendix is organized as follows:
\bigskip

\noindent \ref{appendix_sec:diff_num_protos}\quad Results Using Different Num. of Prototypes\\
\noindent \ref{appendix_sec:ablation_study}\quad Ablation Study\\
\noindent \ref{appendix_sec:efficiency_compare}\quad Efficiency Comparison\\
\noindent \ref{appendix_sec:more_qualitative}\quad Additional Qualitative Results\\
\noindent \ref{appendix_sec:failure_cases}\quad Confusion Matrices and Failure Cases\\
\noindent \ref{appendix_sec:imple_details}\quad Method and Implementation Details\\ 
\noindent \ref{appendix_sec:related}\quad Extended Discussion on Related Work\\ 
\noindent \ref{appendix_sec:discussion}\quad Discussion of \ours (e.g., Societal Impact)\\ 

\bigskip

\renewcommand{\thesection}{A}
\section{Results Using Different Num. of Prototypes}
\label{appendix_sec:diff_num_protos} 

In Table~\ref{table:num_clusters}, we evaluate the impact of the number of prototypes $K$ (i.e., the number of clip clusters) that is used for contrastive clustering learning on the GAR accuracy of \ourseos.  We use the Original split of the Volleyball dataset for this evaluation.  We observe  
that varying the number of prototypes 
does not affect much the performance. The performance  
first improves as the number of prototypes increases, then decreases as the number of prototypes keeps increasing. 
The number of prototypes has little influence as long as it is reasonably ``enough''. 
The practice is to set the number of prototypes at least one order of magnitude larger than the true number of classes in the dataset~\cite{swav}.
Hence, for simplicity, we do not spend extensive efforts in fine-tuning \ours w.r.t. this hyper-parameter; results reported in the main paper are from \ours trained using $1,000$ prototypes for all datasets and splits.

\begin{table}[ht]\centering
\setlength{\tabcolsep}{4.8pt}
\caption{\textbf{Impact of the number of prototypes.} GAR accuracy of \ours on the Original split of the Volleyball dataset using different number of prototypes. 
   }
   \label{table:num_clusters}
   \begin{tabular}[t]{lcccccc}
   \toprule
   \textbf{Number of Prototypes} &  $10$ &  $50$ & $100$ & $1,000$ & $5,000$ & $10,000$ \\
   \midrule
 GAR Accuracy ($\%$)   &  $94.02$   &  $94.54$  &  $94.69$  & $94.62$  & $94.54$   &   $94.32$   \\
   \bottomrule
   \end{tabular}
\end{table}

\begin{figure}[ht]
	\centering
	\includegraphics[scale=0.32]{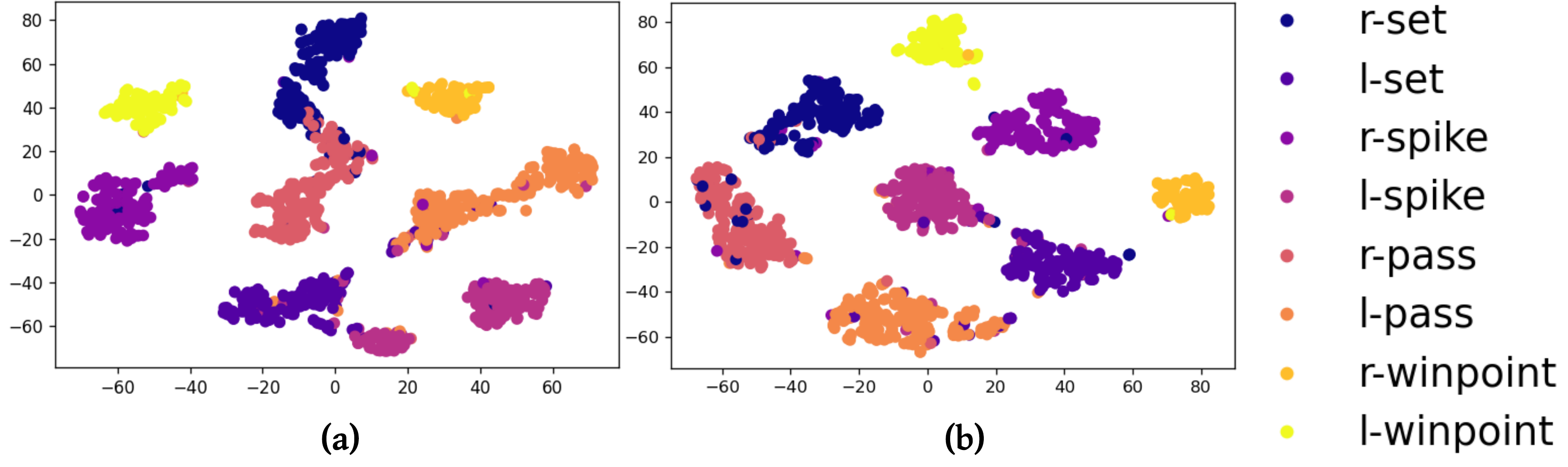}
	\caption{t-SNE visualizations on the Volleyball dataset  show that the clip embedding space learned by \ours using different number of prototypes: (a) $10$ prototypes, and (b) $1000$ prototypes. Best viewed in color. More number of prototypes can lead to a  
	better separation of the clips in distinct group activity classes. 
}
	\label{fig:embedding_diff_clusters} 
\end{figure}

We visualize the clip embedding space (after t-SNE $2$D projection) learned by \ours using $10$ prototypes and $1000$ prototypes in Fig.~\ref{fig:embedding_diff_clusters} (a) and (b), respectively. We take the representation of the \texttt{CLS} token from the last scale and the last \mtx block as the representation of the clip to produce the embedding space visualization. In Fig.~\ref{fig:embedding_diff_clusters}, each dot represents a test clip and the color of the dot indicates the group activity label of the clip. A higher number of prototypes can lead to a better grouping of clips with the same group activity class, as well as a better separation of clips in different group activity classes; this accords with the quantitative results shown in Table~\ref{table:num_clusters}.

\renewcommand{\thesection}{B}
\section{Ablation Study}
\label{appendix_sec:ablation_study}

We conduct ablation experiments to verify the effectiveness of proposed techniques; results are in Table~\ref{table:ablations} (methodology details 
are in Appendix~\ref{appendix_subsec:ablation_imple_details}).

\noindent \textbf{Contrastive clustering and scale agreement regularize the multiscale representations}. 
As demonstrated in Table.~\ref{table:ablations}, performance drops without the contrastive clustering learning. 
The swapped prediction setup 
helps the model to maintain consistency across representations of the multiple scales of the same clip, which regularizes the intermediate representations.

We also experiment with the `Label Consistency' method~\cite{shi2020online} that minimizes the $L2$ distance between $2$ views of an instance in the logit space. Replacing contrastive clustering 
with Label Consistency for scale agreement,  result is better than the previous ablation, but worse than \ourseos. 
Better performance of \ours can be attributed to the additional benefits of the clustering loss, which draws clips that are semantically related close together by comparing with the prototypes. Both experiments indicate that 
encouraging scale agreement can bring benefits for the multi-scale  
learning models, which is unfortunately neglected by prior works.

\noindent \textbf{Greater number of scales 
yields more information and an effective scale agreement}.  
We find that increasing the number of  
scales leads to a higher accuracy.  
More scales indicates more information about the entities in the scene. 
Besides, 
given more scales, hierarchical representations are able to be better maintained, and techniques such as contrastive clustering and auxiliary predictions are 
more 
effective.

\setlength{\tabcolsep}{4pt}
\begin{table}[t] 
\begin{center}
\caption{\textbf{Ablation study of \ours on Volleyball original split.} The ablation study verifies the effectiveness of each proposed technique.
  } 
\label{table:ablations}
\begin{tabular}{l|c}
\hline 
\textbf{Ablation} & \textbf{Test Acc. ($\%$) $\uparrow$ } \\\hline  
 No Clustering  & $93.4$  \\
 Label Consistency for Scale Agreement  &   $93.9$   \\ \hline
 $1$-Scale: Keypoint  &  $91.2$ \\
 $2$-Scale: Keypoint + Person  & $93.2$  \\ 
 $3$-Scale: Keypoint + Person + Interaction &  $93.9$  \\ \hline
 No Actor Dropout     &  $94.0$ \\
 No Horizontal Flip    &  $94.0$  \\
 No Horizontal Move  & $94.2$   \\
 No Vertical Move   & $94.2$   \\ \hline
 No Auxiliary Prediction    & $93.3$ 
 \\
 No Multiscale Transformer   & $88.1$  \\ \hline
 Transformer Encoder Reverse Order   & $86.8$  \\  
 Transformer Encoder Parameter Sharing   & $93.4$  \\ 
 All Tokens to One Transformer Encoder   & $92.4$  \\  
 Time-Varying Person Grouping  & $92.8$  \\ \hline\hline 
 \ours (our full model)  &  $\mathbf{94.6}$   \\
\hline
\end{tabular}
\end{center}
\end{table}
\setlength{\tabcolsep}{1.4pt}

\noindent \textbf{Data augmentations increase the training data size and inject benign noises, leading to generalization}.
Results in Table~\ref{table:ablations} show the gains brought by each  
data augmentation technique described in Sec. 3.4 of the main paper. 
Among the four types of data augmentation, Horizontal Flip (which is commonly used by existing works~\cite{zappardino2021learning,POGARS}) and Actor Dropout 
are the most critical ones.
Even though data augmentation is less effective than other techniques proposed, it increases the training data size and injects noises that help model to generalize.

\noindent \textbf{Auxiliary prediction aids learning the intermediate representations}. For the `No Auxiliary Prediction' ablation, the loss of group activity  
is only
computed from 
the clip 
token from the last scale of the last \mtx (note that the person action loss and clustering loss are still in use).  
Performance of this ablation largely drops, indicating auxiliary prediction is a simple but yet effective technique.

\noindent \textbf{\mtx learns higher-level knowledge of the video by compositionally reasoning over concepts from finest-grained to coarsest-grained}. 
We design an ablation to remove \mtxeos. Specifically, the group activity classifier directly takes features of the object token and person tokens (learned from features of keypoint tokens) as an input, and the person action classifier takes the features of person tokens as an input. 
The outcome of 
this 
ablation is worse performance than the $1$-scale ablation that does not even use person features and person action labels -- which indicates the importance of relational reasoning.

\noindent \textbf{Misc}. Unlike previous works~\cite{ehsanpour2020joint,koshkina2021contrastive}, person-to-group association  mechanism
is not our focus. 
Hence, we use heuristics~\cite{GIRN} for the Volleyball dataset, K-means~\cite{GroupFormer} for the Collective Activity dataset,
and set $g\prime$=$2$
without tuning.
With K-means, actors
are mapped to groups
adaptively in each \mtx
block 
as person representations are refined. 
We have experimented with
\textit{time}-varying
person 
grouping 
with K-means, and the result is 
$\mathbf{92.8\%}$ 
on the original split of the Volleyball dataset.  
For this
experiment, 
since we need person and group representations at each timestamp, the number of tokens is increased (multiplied by $T$) for the 2 transformer encoders, which potentially leads to issues such as over-smoothing~\cite{yuan2021spatio}
and raises the challenge for attention.  
We hypothesize that a carefully-crafted mechanism for learning spatio-temporal relations
is required for time-varying person grouping.

We have experimented with \mtx variations.
In \mtxeos, 4 different transformer
encoders separately model the contextual information of each scale, which eases learning (because the information granularity and features across scales can vary significantly). 
Parameter-sharing across the 4 encoders yielded a result of 
$\mathbf{93.4\%}$ 
on the original split of the Volleyball dataset, and feeding all tokens to one encoder obtained $\mathbf{92.4\%}$.
The order of encoders in \mtx is in accordance with the hierarchy of person-related tokens, from fine to coarse.
This allows \ours to \textit{compose} the high-level representations from low-level ones and distill the knowledge.
One can perform grid search 
to find the optimal order
of the 4 encoders, 
but that will lead to $24$ 
permutations which require lots of 
resources and runtime.
We have performed an experiment with the reverse order (representations of coarser tokens are broadcast to finer tokens), and the result is $\mathbf{86.8\%}$, 
which is much worse.

 \renewcommand{\thesection}{C}
\section{Efficiency Comparison}
\label{appendix_sec:efficiency_compare}

\noindent \textbf{Backbone efficiency comparison to support the keypoint-only setup. }
Backbones used by prior works 
vary \textit{a lot} (see Table~\ref{table:VD_compare_sota_detailed} and Table~\ref{table:CAD_compare_sota_detailed}). 
Because AT~\cite{actor-transformer} is  a Transformer-based model like ours and it  
has reported result of using the keypoint-only modality, 
we follow AT to use the HRNet~\cite{hrnet} as the person keypoint estimation backbone.

We also report the results of \ours using POGARS's person keypoint estimation backbone -- Hourglass~\cite{newell2016stacked}  
since POGARS~\cite{POGARS} is the keypoint-only method that has the closest result to ours on the Volleyball dataset. Comparing \ours with POGARS (both use Hourglass as the backbone): on VD Olympic split \ours $92.9\%$ v.s. POGARS $89.7\%$; and on VD Original split \ours $94.3\%$ v.s. POGARS $93.9\%$. 
The  superiority of \ours is not affected  
because \ours is able to address noisy estimated keypoints and 
disregard inaccurate keypoints by modeling attention over the keypoints (unlike POGARS or AT that just model attention at the person scale).

\setlength{\tabcolsep}{4pt}
\begin{table}[t]
\fontsize{7.5pt}{7.5pt}\selectfont
\begin{center}
\caption{Efficiency Comparison (FLOPs). Please see details in Appendix~\ref{appendix_sec:efficiency_compare}.
  } 
\label{table:efficiency}
\begin{tabular}{|c|c|c|c|c|c|c|c|}
\hline 
 \multicolumn{6}{|c|}{Backbone}  &  \multicolumn{2}{c|}{GAR Model}     \\  \cline{1-8} 
  HRNet  &  VGG-19   &   VGG-16   &    Inception-v3     & ResNet-18   & AlexNet     &   \ours &  GroupFormer~\cite{GroupFormer}    \\ \hline 
  \textbf{0.9 T} & 3.6 T   & 2.8 T     & 0.7 T       &   0.3 T   & 0.1 T    &\textbf{ 297 M }&    595 M \\   \hline  
\end{tabular}
\end{center}
\vspace{-7pt}
\scriptsize{*Note: `T' stands for trillion and `M' for million. Ours are marked in bold.  }
\end{table}
\setlength{\tabcolsep}{1.4pt}

To support the keypoint-only setup, we compute the FLOPs for HRNet (our keypoint backbone) and RGB backbones used by prior works 
when obtaining features of all persons in a clip on the Volleyball dataset. 
FLOPs are:
{HRNet} $\mathbf{0.9}${T}, 
{VGG-19}~\cite{simonyan2014very} $\mathbf{3.6}${T}, 
{VGG-16}~\cite{simonyan2014very} $\mathbf{2.8}${T}, {Inception-v3}~\cite{szegedy2016rethinking} $\mathbf{0.7}${T}, {ResNet-18}~\cite{he2016deep} $\mathbf{0.3}${T}, and {AlexNet}~\cite{krizhevsky2012imagenet} $\mathbf{0.1}${T}  
(Table~\ref{table:efficiency}). 
VGG-19 and VGG-16 are more computational expensive than HRNet. 
Our method is agnostic to the type of the keypoint estimation backbone and robust w.r.t. noisy estimated keypoints. 
Therefore, real-time applications can use an efficient  
backbone.

\noindent\textbf{Efficiency comparison with SOTA GAR methods. } 
We also compare FLOPs of prior Transformer-based GAR methods with ours. 
For a fair comparison,  
the methods all have $2$ blocks of their respective GAR reasoning module (e.g., \mtx in \ours and  the CSTT module in GroupFormer) and share  hyper-parameters of the Transformers inside (e.g., dimension of the FFN layer inside Transformer).
In addition, computation  spent on person feature extraction  
is excluded since different backbones can be used. 

FLOPs are (given a Collective Activity dataset's input):
Ours \textbf{ 297M}, 
GroupFormer~\cite{GroupFormer} \textbf{595M}, and
AT~\cite{actor-transformer}]  \textbf{17M} where `M' stands for million. 
While AT is the most efficient one, its efficacy (only $92.8\%$ on Collective Activity) and generalization are unsatisfactory. 
GroupFormer is the most
computational 
expensive one 
due to $5$ Transformers  
inside its CSTT module and leveraging image scenes -- 
even we only report FLOPs for its most basic version that leverages the least signals possible (RGB + Scene). 

Note that the  \ours variant that uses RGB modality (mentioned in the main paper) has \textbf{127M} FLOPs and is more efficient than \ours that uses keypoints because the former only models $3$ scales (person, interaction and group).
To further reduce latency for real-life applications (e.g., on-device scenarios),
we can have a light-weight \ours variant that only models one scale during inference  or uses smaller hidden dimensions  and yet retains the most efficacy and generalization by using techniques like knowledge distillation~\cite{hinton2015distilling}.

\begin{table*}[t] 
\fontsize{7.5pt}{7.5pt}\selectfont
   \aboverulesep=0ex
   \belowrulesep=0ex 
  \begin{center}
  \caption{Detailed comparisons between our results and the reported SOTA methods' results on the \textbf{Original} split of the \textbf{Volleyball} dataset.  
  We have made an effort to collect and list results of prior works using various backbones and modalities. Note that the backbones used by prior works vary a lot. 
The top 3 performance scores are highlighted as:
\textbf{\red{First}}, \blue{\textit{Second$^{\ast}$}}, \magenta{\textit{Third}}. 
\ours outperforms the latest GAR methods that use a single modality ($+0.7\%$ improvement), and performs favorably compared against methods that exploit multiple 
expensive
modalities (Appendix~\ref{appendix_sec:efficiency_compare})}
\label{table:VD_compare_sota_detailed}
\vspace{-15pt}
   \begin{tabular}[t]{c|c|c|c|c|c|c|c}
   \toprule
   \rule{0pt}{1.0EM} 
   \multirow{2}{*}{\textbf{Method}} & \multicolumn{4}{c|}{\textbf{Modality}}    & \multicolumn{2}{c|}{\textbf{Backbone}}  & \multirow{2}{*}{\textbf{Acc. $\uparrow$ ($\%$)}}  \\ \cline{2-7}  
  & Keypoint & RGB & Flow & Scene & Keypoint & RGB/Flow/Scene   &   \\
   \bottomrule\rowcolor{aureolin!10}
   \rule{0pt}{1.2EM} 
 HDTM~\cite{ibrahim2016hierarchical}   &       &  \tablecheck{\CheckmarkBold}    &      &       &   --   &   AlexNet &      $81.9$   \\ \midrule\rowcolor{aureolin!10} \rule{0pt}{1.2EM}   
 CERN~\cite{cern}   &          &  \tablecheck{\CheckmarkBold}    &      &       &   --   &   VGG-16 &      $83.3$    \\ \midrule\rowcolor{aureolin!10} \rule{0pt}{1.2EM}
 stagNet~\cite{stagnet}   &          &  \tablecheck{\CheckmarkBold}    &      &       &   --   &   VGG-16 &    $89.3$  \\ \midrule\rowcolor{aureolin!10} \rule{0pt}{1.2EM}
 RCRG~\cite{rcrg}   &          &  \tablecheck{\CheckmarkBold}    &      &       &    --  &   VGG-19 &     $89.5$  \\ \midrule\rowcolor{aureolin!10} \rule{0pt}{1.2EM}
 SSU~\cite{ssu}   &         &  \tablecheck{\CheckmarkBold}    &      &       &   --   &    Inception-v3 &      $90.6$ \\ \midrule\rowcolor{aureolin!10} \rule{0pt}{1.2EM}
 PRL~\cite{prl}   &      &  \tablecheck{\CheckmarkBold}    &      &       &   --   &   VGG-16   &   $91.4$    \\ \midrule\rowcolor{aureolin!10} \rule{0pt}{1.2EM}
AT~\cite{actor-transformer}     &      &   \tablecheck{\CheckmarkBold}    &   &     & --  &  I3D    & $91.4$   \\ \midrule\rowcolor{aureolin!10} \rule{0pt}{1.2EM}
   \cellcolor{aureolin!10}   &     \cellcolor{aureolin!10}   &  \cellcolor{aureolin!10}  \tablecheck{\CheckmarkBold}    &   \cellcolor{aureolin!10}    &   \cellcolor{aureolin!10}     &  \cellcolor{aureolin!10}  --   &  \cellcolor{aureolin!10} VGG-16   &   \cellcolor{aureolin!10}  $91.9$    \\
   \cellcolor{aureolin!10}   &     \cellcolor{aureolin!10}   &  \cellcolor{aureolin!10}  \tablecheck{\CheckmarkBold}    &   \cellcolor{aureolin!10}    &   \cellcolor{aureolin!10}     &  \cellcolor{aureolin!10}  --   &  \cellcolor{aureolin!10} Inception-v3   &   \cellcolor{aureolin!10}  $92.5$    \\
    \cellcolor{aureolin!10}  \multirow{-3}{*}{  ARG~\cite{arg}}     &     \cellcolor{aureolin!10}     &  \cellcolor{aureolin!10} \tablecheck{\CheckmarkBold}    &    \cellcolor{aureolin!10}   &   \cellcolor{aureolin!10}     &  \cellcolor{aureolin!10}  --   &  \cellcolor{aureolin!10} VGG-19     &  \cellcolor{aureolin!10} $92.6$   \\ \midrule\rowcolor{aureolin!10} \rule{0pt}{1.2EM}
\cellcolor{aureolin!10}     &  \cellcolor{aureolin!10}          &  \cellcolor{aureolin!10}  \tablecheck{\CheckmarkBold}    & \cellcolor{aureolin!10}       &     \cellcolor{aureolin!10}    &  \cellcolor{aureolin!10}   --   & \cellcolor{aureolin!10}   AlexNet  &   \cellcolor{aureolin!10}   $88.6$   \\  
\cellcolor{aureolin!10}  \multirow{-2}{*}{ HiGCIN~\cite{higcin}}      &  \cellcolor{aureolin!10}          & \cellcolor{aureolin!10}   \tablecheck{\CheckmarkBold}    &  \cellcolor{aureolin!10}      &  \cellcolor{aureolin!10}       & \cellcolor{aureolin!10}    --   &  \cellcolor{aureolin!10}  ResNet-18  &  \cellcolor{aureolin!10}    $91.5$   \\ \midrule\rowcolor{aureolin!10} \rule{0pt}{1.2EM}   
\cellcolor{aureolin!10}      &   \cellcolor{aureolin!10}       &   \cellcolor{aureolin!10}  \tablecheck{\CheckmarkBold}   & \cellcolor{aureolin!10}    &   \cellcolor{aureolin!10}    &  \cellcolor{aureolin!10}  -- &  \cellcolor{aureolin!10} ResNet-18  &  \cellcolor{aureolin!10}   $93.1$      \\ \rowcolor{aureolin!10}  
   \cellcolor{aureolin!10}  \multirow{-2}{*}{DIN~\cite{yuan2021spatio}}    &    \cellcolor{aureolin!10}  &   \cellcolor{aureolin!10}  \tablecheck{\CheckmarkBold}   & \cellcolor{aureolin!10}    &  \cellcolor{aureolin!10}     &  \cellcolor{aureolin!10} --  &  \cellcolor{aureolin!10}  VGG-16   &   \cellcolor{aureolin!10}  $93.6$     \\   \bottomrule \toprule \rowcolor{ballblue!10} \rule{0pt}{1.2EM}
Zappardino~\etal~\cite{zappardino2021learning}   & \tablecheck{\CheckmarkBold}       &       &      &       &  OpenPose  &   --  &    $91.0$ \\  \midrule\rowcolor{ballblue!10} \rule{0pt}{1.2EM}
GIRN~\cite{GIRN}  & \tablecheck{\CheckmarkBold}       &       &      &       &  OpenPose  &  --   &   $92.2$ \\  \midrule\rowcolor{ballblue!10}\rule{0pt}{1.2EM}
AT~\cite{actor-transformer}  & \tablecheck{\CheckmarkBold}      &       &      &       &  HRNet  &    --  &   $92.3$ \\  \midrule\rowcolor{ballblue!10} \rule{0pt}{1.2EM}
POGARS~\cite{POGARS}   & \tablecheck{\CheckmarkBold}   &            &      &       &  Hourglass  &  --   &      $93.9$ \\   \bottomrule \toprule \rule{0pt}{1.2EM} 
CRM~\cite{crm}  &       &  \tablecheck{\CheckmarkBold}    &   \tablecheck{\CheckmarkBold}   &       &   --   &   I3D  & $93.0$ \\ \midrule \rule{0pt}{1.2EM}
 Ehsanpour~\etal~\cite{ehsanpour2020joint}     &       &  \tablecheck{\CheckmarkBold}    &      &   \tablecheck{\CheckmarkBold}    &   --   &   I3D   &    $93.1$ \\ \midrule \rule{0pt}{1.2EM}
\multirow{3}{*}{AT~\cite{actor-transformer}}     
  &          &  \tablecheck{\CheckmarkBold}  & \tablecheck{\CheckmarkBold}   &     &  -- &   I3D    &      $93.0$    \\
  &    \tablecheck{\CheckmarkBold}     &  \tablecheck{\CheckmarkBold}  &   &     & HRNet &   I3D    &      $93.5$    \\
  &   \tablecheck{\CheckmarkBold}     &     &  \tablecheck{\CheckmarkBold}  &     & HRNet &   I3D     &     $94.4$   \\  \midrule \rule{0pt}{1.2EM}
\multirow{3}{*}{GIRN~\cite{GIRN}}   & \tablecheck{\CheckmarkBold} &   \tablecheck{\CheckmarkBold}    &    &      &     OpenPose   &  Inception-v3      &     $93.5$   \\
  &   \tablecheck{\CheckmarkBold}      &      &   \tablecheck{\CheckmarkBold}    &        &   OpenPose   &    Inception-v3 &       $93.0$    \\
  &  \tablecheck{\CheckmarkBold}   &  \tablecheck{\CheckmarkBold}    &   \tablecheck{\CheckmarkBold}     &        &  OpenPose    &    Inception-v3    &   $94.0$   \\   \midrule \rule{0pt}{1.2EM}
\multirow{7}{*}{TCE+STBiP~\cite{yuan2021learning}} &   \tablecheck{\CheckmarkBold}       &    &   & \tablecheck{\CheckmarkBold}  & HRNet &  Inception-v3    &   $92.9$    \\
          &      &   \tablecheck{\CheckmarkBold}    &   & \tablecheck{\CheckmarkBold}  &  -- &  Inception-v3   &     $93.3$     \\
     &   \tablecheck{\CheckmarkBold}      &    \tablecheck{\CheckmarkBold}   &   & \tablecheck{\CheckmarkBold}  & HRNet &  Inception-v3   &    $94.1$   \\
    &   \tablecheck{\CheckmarkBold}        &    &   & \tablecheck{\CheckmarkBold}  & HRNet &  VGG-16  &    $92.9$     \\
     &         &   \tablecheck{\CheckmarkBold}    &   & \tablecheck{\CheckmarkBold}  & --  &   VGG-16   &     $94.1$     \\
     &   \tablecheck{\CheckmarkBold}    &   \tablecheck{\CheckmarkBold}   &   & \tablecheck{\CheckmarkBold}  & HRNet &  VGG-16  &   $94.7$   \\  
     \midrule \rule{0pt}{1.2EM}
\multirow{2}{*}{SACRF~\cite{pramono2020empowering}}  &        &  \tablecheck{\CheckmarkBold}   &    \tablecheck{\CheckmarkBold}  &    \tablecheck{\CheckmarkBold}   &    --   &  I3D       &  $94.5$   \\
  &   \tablecheck{\CheckmarkBold}        &   \tablecheck{\CheckmarkBold}  &  \tablecheck{\CheckmarkBold}    &    \tablecheck{\CheckmarkBold}   &   AlphaPose   &   I3D   &    \blue{\textit{\text{95.0}}$^{\ast}$} \\ \midrule \rule{0pt}{1.2EM}
\multirow{3}{*}{GroupFormer~\cite{GroupFormer}} &        &   \tablecheck{\CheckmarkBold}   &      &   \tablecheck{\CheckmarkBold}     &     --  &  Inception-v3   &   $94.1$   \\
  &          &   \tablecheck{\CheckmarkBold}   &     \tablecheck{\CheckmarkBold}  &     \tablecheck{\CheckmarkBold}   &   --    &  I3D &    \magenta{\textit{\text{94.9}}}    \\
  &   \tablecheck{\CheckmarkBold}       &   \tablecheck{\CheckmarkBold}   &     \tablecheck{\CheckmarkBold}  &    \tablecheck{\CheckmarkBold}    & AlphaPose    &  I3D &    \red{$\mathbf{95.7}$}    \\   \bottomrule \toprule \rule{0pt}{1.2EM}
 \cellcolor{ballblue!10}     &   \cellcolor{ballblue!10}    \tablecheck{\CheckmarkBold}       &     \cellcolor{ballblue!10}    &    \cellcolor{ballblue!10}      &      \cellcolor{ballblue!10}     &   \cellcolor{ballblue!10}   Hourglass    &   \cellcolor{ballblue!10}   --    &      \cellcolor{ballblue!10}  94.3 \\ 
  \cellcolor{ballblue!10}  \multirow{-3}{*}{\ours (ours)}  &   \cellcolor{ballblue!10}    \tablecheck{\CheckmarkBold}       &   \cellcolor{ballblue!10}      &    \cellcolor{ballblue!10}      &     \cellcolor{ballblue!10}      &  \cellcolor{ballblue!10}    HRNet    &    \cellcolor{ballblue!10}  --    &    \cellcolor{ballblue!10}      $94.6$ \\ 
  \bottomrule
   \end{tabular}
   \end{center}
   \vspace{-8pt}
   \scriptsize{*Note: ``Flow'' denotes optical flow input, and ``Scene'' denotes additional image context features of the entire frames (\textit{\ul{fewer checks are better}}). Yellow shaded rows highlight that the methods use just the RGB-based input, whereas blue for just keypoint.  }
  
\end{table*}

\clearpage

\begin{table*}[t]  
\fontsize{8.5pt}{8.5pt}\selectfont
   \aboverulesep=0ex
   \belowrulesep=0ex 
  \begin{center} 
  \caption{Detailed comparisons between our results and the reported SOTA methods' results on the \textbf{Collective Activity} dataset.   
The top 3 performance scores are highlighted as:
\textbf{\red{First}}, \blue{\textit{Second$^{\ast}$}}, \magenta{\textit{Third}}. 
\ours outperforms the latest GAR methods that use a single modality ($+2.8\%$ improvement), and performs favorably compared against methods that exploit multiple expensive modalities (ours is the second best)}
   \label{table:CAD_compare_sota_detailed}
\vspace{-10pt}
   \begin{tabular}[t]{c|c|c|c|c|c|c|c}
   \toprule
   \rule{0pt}{1.0EM} 
   \multirow{2}{*}{\textbf{Method}} & \multicolumn{4}{c|}{\textbf{Modality}}    & \multicolumn{2}{c|}{\textbf{Backbone}} &  \multirow{2}{*}{\textbf{Acc. $\uparrow$ ($\%$)}}  \\ \cline{2-7}  
  & Keypoint & RGB & Flow & Scene & Keypoint & RGB/Flow/Scene  &    \\
   \bottomrule\rowcolor{aureolin!10}
   \rule{0pt}{1.2EM}
 HDTM~\cite{ibrahim2016hierarchical}   &            &  \tablecheck{\CheckmarkBold}    &       &        &    --   &   AlexNet &     $81.5$ \\ \midrule\rowcolor{aureolin!10} \rule{0pt}{1.2EM}
 CERN~\cite{cern}   &      &  \tablecheck{\CheckmarkBold}    &       &        &      --  &   VGG-16 &   $87.2$ \\ \midrule\rowcolor{aureolin!10} \rule{0pt}{1.2EM}
 stagNet~\cite{stagnet}      &        &  \tablecheck{\CheckmarkBold}    &       &        &    --    &   VGG-16    & $89.1$ \\ \midrule\rowcolor{aureolin!10} \rule{0pt}{1.2EM}
 \cellcolor{aureolin!10}       &    \cellcolor{aureolin!10}      &  \cellcolor{aureolin!10}  \tablecheck{\CheckmarkBold}    &   \cellcolor{aureolin!10}      &   \cellcolor{aureolin!10}       &      \cellcolor{aureolin!10}  --  &   \cellcolor{aureolin!10} VGG-16 &   \cellcolor{aureolin!10}   $90.1$  \\
 \cellcolor{aureolin!10}  \multirow{-2}{*}{ARG~\cite{arg}}       &     \cellcolor{aureolin!10}     &  \cellcolor{aureolin!10}  \tablecheck{\CheckmarkBold}    &    \cellcolor{aureolin!10}     &     \cellcolor{aureolin!10}     &     \cellcolor{aureolin!10}   --  &  \cellcolor{aureolin!10}  Inception-v3   &    \cellcolor{aureolin!10}  $91.0$ \\ \midrule\rowcolor{aureolin!10} \rule{0pt}{1.2EM}
    \cellcolor{aureolin!10}     &     \cellcolor{aureolin!10}     &  \cellcolor{aureolin!10}  \tablecheck{\CheckmarkBold}    &    \cellcolor{aureolin!10}     &    \cellcolor{aureolin!10}      &   \cellcolor{aureolin!10}    --   & \cellcolor{aureolin!10}  AlexNet     &  \cellcolor{aureolin!10}  $92.5$   \\ 
   \cellcolor{aureolin!10}  \multirow{-2}{*}{ HiGCIN~\cite{higcin} }    &   \cellcolor{aureolin!10}       &  \cellcolor{aureolin!10}  \tablecheck{\CheckmarkBold}    &   \cellcolor{aureolin!10}      &      \cellcolor{aureolin!10}    &    \cellcolor{aureolin!10}   --   &  \cellcolor{aureolin!10}  ResNet-18      &  \cellcolor{aureolin!10}  $93.4$   \\  \bottomrule \toprule \rule{0pt}{1.2EM}
CRM~\cite{crm}       &        &  \tablecheck{\CheckmarkBold}    &   \tablecheck{\CheckmarkBold}   &        &     --   &   I3D      & $85.8$ \\ \midrule \rule{0pt}{1.2EM}
Ehsanpour~\etal~\cite{ehsanpour2020joint}    &        &  \tablecheck{\CheckmarkBold}    &       &   \tablecheck{\CheckmarkBold}  &      --  &   I3D        &  $89.4$  \\ \midrule \rule{0pt}{1.2EM}
\multirow{3}{*}{AT~\cite{actor-transformer}}   &    \tablecheck{\CheckmarkBold}    &  \tablecheck{\CheckmarkBold}  &    &      & HRNet &   I3D         & $91.0$    \\
   &   \tablecheck{\CheckmarkBold}  &        &  \tablecheck{\CheckmarkBold}  &      & HRNet &   I3D        &  $91.2$    \\ 
   &      &  \tablecheck{\CheckmarkBold}  & \tablecheck{\CheckmarkBold}   &      &   -- &   I3D     &  $92.8$    \\ \midrule \rule{0pt}{1.2EM}
\multirow{2}{*}{SACRF~\cite{pramono2020empowering}}  &         &  \tablecheck{\CheckmarkBold}   &    \tablecheck{\CheckmarkBold}  &    \tablecheck{\CheckmarkBold}   &     --   &  I3D         &  $94.6$  \\
  &   \tablecheck{\CheckmarkBold}        &   \tablecheck{\CheckmarkBold}  &  \tablecheck{\CheckmarkBold}    &    \tablecheck{\CheckmarkBold}   &   AlphaPose   &   I3D     & \magenta{\textit{95.2}}  \\ \midrule \rule{0pt}{1.2EM}
\multirow{3}{*}{GroupFormer~\cite{GroupFormer}} &       &   \tablecheck{\CheckmarkBold}   &       &     \tablecheck{\CheckmarkBold}     &      --  &  Inception-v3    &  $93.6$  \\
  &           &   \tablecheck{\CheckmarkBold}   &     \tablecheck{\CheckmarkBold}  &     \tablecheck{\CheckmarkBold}     &       -- &  I3D &      $94.7$   \\
  &   \tablecheck{\CheckmarkBold}      &   \tablecheck{\CheckmarkBold}   &     \tablecheck{\CheckmarkBold}  &    \tablecheck{\CheckmarkBold}      & AlphaPose    &  I3D    &  \red{$\mathbf{96.3}$}    \\   \bottomrule \toprule \rowcolor{ballblue!10}  \rule{0pt}{1.2EM}   \rule{0pt}{1.2EM} 
\ours (ours)  &    \tablecheck{\CheckmarkBold}        &       &        &         &  HRNet    &      --      &  \blue{\textit{\text{96.2}}$^{\ast}$} \\ 
   \bottomrule
   \end{tabular}
    \end{center} 
   \scriptsize{*Note: ``Flow'' denotes optical flow input, and ``Scene'' denotes additional image context features of the entire frames (\textit{\ul{fewer checks are better}}). Yellow shaded rows highlight that the methods use just the RGB-based input, whereas blue for just keypoint. We are the first to report the result of a keypoint-only method on this dataset. }
   
\end{table*}

\renewcommand{\thesection}{D}
\section{Additional Qualitative Results}
\label{appendix_sec:more_qualitative}

\begin{figure*}[t]
	\centering
	\includegraphics[scale=0.15]{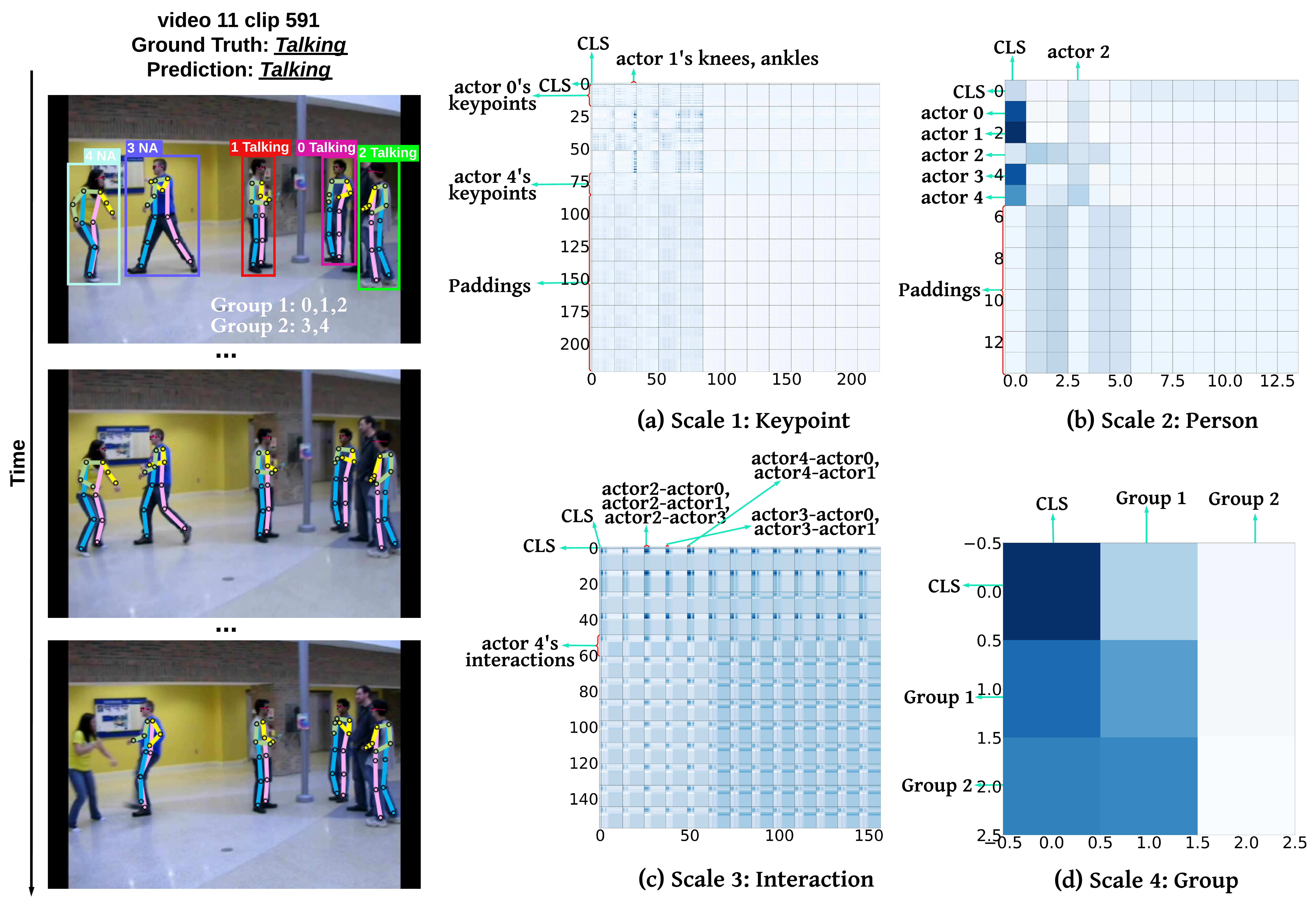}
	\caption{\textbf{Qualitative results of \ours on CAD} -- showcasing attention matrices of a test set instance in the ``\textbf{talking}'' class. The person to person group mapping identified by \ourseos, estimated keypoints from the backbone, and the ground truth person bounding boxes with action labels are also shown on the left side of the figure. 
}
	\label{fig:CAD_good_talking}
\end{figure*}

We visualize the attention matrices produced by the last \mtx block in \ours of CAD test clips in different group activity classes in 
Fig.~\ref{fig:CAD_good_talking},~\ref{fig:CAD_good_moving},~\ref{fig:CAD_good_quequeing} and~\ref{fig:CAD_good_waiting}.
In Fig. 5 of the main paper, we visualize the attention matrices of a VD test clip in group activity class ``pass''. We provide the same visualization for the other $3$ main group activity classes of VD 
in Fig.~\ref{fig:good_r_set},~\ref{fig:good_l_spike} and~\ref{fig:good_r_winpoint}.

In the figures, we highlight the tokens that \ours has mostly attended to at each scale (darker color denotes larger attention weights). Each figure contains rich information. Please zoom in on the images to appreciate the details.
Here, we summarize the main findings:
\begin{enumerate}
    \item Actor-related tokens associated with the key person(s) are often identified as the most important tokens for the \texttt{CLS} and object token by \ours (e.g., Fig.~\ref{fig:good_r_set},~\ref{fig:good_l_spike}, etc). In certain cases, the object token is also identified as important (e.g., as shown in Fig.~\ref{fig:good_l_spike}, the object token `ball' is identified as important at scale $1$, $3$ and $4$). \ours is able to attend to relevant entities across different scales and can produce interpretable results.
    
    \item \ours learns to recognize the group activities based upon the unique characteristics of each group activity class. E.g., in Fig.~\ref{fig:good_r_winpoint}, the pattern of the attention weights  of the \textit{winpoint} class at scale $3$  
    is quite different from the pattern observed in the other 3 main group activity classes in VD.
    Recognition of the other group activities in VD (e.g., \textit{spike}), is often determined by a key player who is performing the key action (e.g., \textit{spiking}). In contrast,  the \textit{winpoint} group activity
    is not heavily dependent on one or two key players; instead, it
    is defined by the overall person-to-person interactions of the team that just scored. As shown in Fig.~\ref{fig:good_r_winpoint}, person-to-person interaction tokens formed by players in the right team tend to closely attend to each other.
     
    \item On CAD, \ours has identified interesting person-to-group mappings as shown in Fig.~\ref{fig:CAD_good_talking},~\ref{fig:CAD_good_moving}, etc. On VD, we find that at the group scale, the \texttt{CLS} and object tokens often mostly attend to each other, and the two teams often mostly attend to each other. We posit that this is because, at the previous three scales, the \texttt{CLS} and object tokens mostly attend to actor-related tokens. The object token, which contains less correlated information, then becomes  the most important signal at scale $4$ for the \texttt{CLS} token in order to correctly recognize the group activity.
    
\end{enumerate}

\renewcommand{\thesection}{E}
\section{Failure Cases and Confusion Matrices}
\label{appendix_sec:failure_cases} 
\subsection{Failure Case Analysis}
\label{appendix_subsec:failure_cases} 
\noindent \textbf{Volleyball}. Through examining failure cases of \ourseos,   we find numerous  mislabeled test clips. We coin these cases as \textit{false failure cases} (i.e., the annotated label is wrong but the prediction from \ours is correct).  
Sometimes the annotation of the team 
is wrong (e.g., a test clip with the \textit{left set} group activity is annotated as \textit{right set} as shown in Fig.~\ref{fig:failure_label_left_right_wrong}), and sometimes the annotated main group activity is wrong (e.g., a test clip with the \textit{right set} group activity is annotated as \textit{right spike} as shown in Fig.~\ref{fig:failure_spike_to_set}). 
As a result, the models that have achieved high accuracy on the Original split of VD might capture misleading noises and cannot generalize, and therefore the pursuit of beating SOTA might be pointless. For future research, proposing methods that make use of RGB signals but overcome scene biases on the Olympic split is a recommended direction.

We visualize two  \textit{true} failure cases of \ours in Fig.~\ref{fig:failure_pass_to_set} and  Fig.~\ref{fig:failure_camera_change}. 
The wrong prediction of \ours made on case shown in Fig.~\ref{fig:failure_pass_to_set}
could be attributed to the fact that the arms of actor $0$ is occluded.
In Fig.~\ref{fig:failure_camera_change}, \ours fails to identify which team is performing the group activity \textit{spike}. However, we notice a major change in the camera location of this test clip, which causes this test clip possibly to be difficult even for humans to make a correct prediction.

\noindent \textbf{Collective Activity}.  We visualize
several failure cases of \ours on the test set of CAD in Fig.~\ref{fig:CAD_bad_occlusion},~\ref{fig:CAD_bad_pred_moving_gt_queueing},~\ref{fig:CAD_bad_pred_moving_gt_waiting}, and~\ref{fig:CAD_bad_pred_short_temporal}. Failures of \ours can be attributed to severe occlusions (Fig.~\ref{fig:CAD_bad_occlusion}), misleading movement of actors (Fig.~\ref{fig:CAD_bad_pred_moving_gt_queueing} and~\ref{fig:CAD_bad_pred_moving_gt_waiting}) and the current use of a relatively short temporal window for each clip (Fig.~\ref{fig:CAD_bad_pred_short_temporal}), which suggest ways to further improve \ourseos, e.g., addressing the issue of severe occlusion, using more frames per clip with attention-based relational reasoning in both spatial and temporal domains, etc.

\begin{figure*}[t]
	\centering
	\includegraphics[scale=0.27]{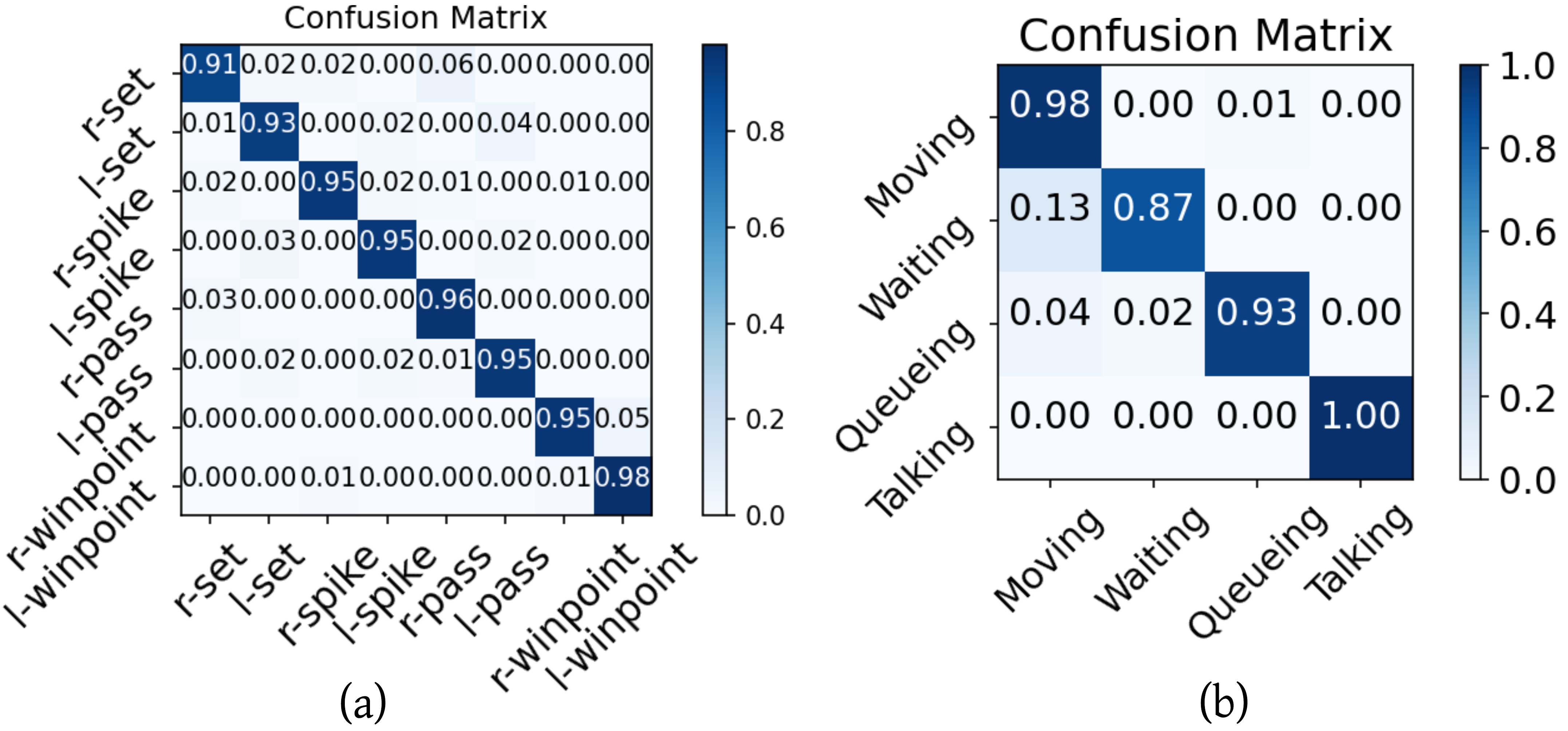} 
	\caption{Confusion matrices  of \ours on (a) the Volleyball dataset (the Original split) and (b) the Collective Activity dataset.
}
	\label{fig:conf_matrices} 
\end{figure*}

\subsection{Confusion Matrix Analysis}
\label{appendix_subsec:conf_matrix}
We show the confusion matrices of \ours in Fig.~\ref{fig:conf_matrices}.
On the Volleyball dataset (the Original split), for each group activity class
\ours achieves an accuracy over $90\%$ with the lowest accuracy for the \textit{right set} class. 
Most failures emerge from discriminating the \textit{set}, \textit{pass}, and \textit{spike} activities which can
be a result of 
highly similar actions or positions of the key player in some clips~\cite{actor-transformer,prl,higcin}. 
Occasionally, the model struggles to distinguish which team (left or right) performs the 
activity. 
We hypothesize that adding more object tokens/keypoints such as the keypoints of the net to \ours may help to address this problem.
Nevertheless, different camera positions of some clips in the dataset (e.g., Fig.~\ref{fig:failure_camera_change}) might cause the difficulty for a model to learn which team performs the activity. 
On the Collective Activity dataset, \ours occasionally mistakes \textit{waiting} to \textit{moving}, which may because the current temporal dynamics of  clips is too short to catch differences between the two classes~\cite{yuan2021learning} and a lot more examples of \textit{moving} than \textit{waiting} in the dataset (more than twice more).

\renewcommand{\thesection}{F}
\section{Method and Implementation Details}
\label{appendix_sec:imple_details}

\subsection{Transformer} 
\label{appendix_subsec:transformer_review} 

We briefly describe the Transformer encoder~\cite{vaswani2017attention} used in \mtx in this subsection. The basic components of the Transformer encoder include 1) Multi-head Self-Attention (MSA), 2) Multi-Layer Perceptron (MLP), and 3) Skip Connection~\cite{he2016deep}, Dropout~\cite{srivastava2014dropout} and Layer Normalization~\cite{ba2016layer} (Add \& Dropout \& LN).

\noindent \textbf{MSA.} 
Central to the Transformer encoder is the self-attention function.
In the self-attention function, the input $X \in \mathbb{R}^{n \times d}$ is first linearly transformed to three parts, i.e., \textit{query} $Q \in \mathbb{R}^{n \times d_{k}}$, \textit{key} $K \in \mathbb{R}^{n \times d_{k}}$ and \textit{value} $V \in \mathbb{R}^{n \times d_{v}}$,  where $n$ denotes the number of tokens in the input sequence, and $d$, $d_{k}$ and $d_{v}$ are the representation dimensions of the input, query (or key), and value, respectively. The Scaled Dot-Product Attention is applied on $Q$, $K$, and $V$:

\begin{equation} 
    \operatorname{Attention}(Q, K, V)=\operatorname{softmax}\left(\frac{Q K^{\top}}{\sqrt{d_{k}}}\right) V
\end{equation}
Then, a linear layer is used to produce the output. 
We use $h$ paralleled heads of the Scaled Dot-Product Attention to increase the representation power. 
Specifically, MSA splits the query, key and value for $h$ times and performs the self-attention function in parallel, and then the output values of each head are concatenated and linearly projected to form the final output~\cite{han2021transformer}.

\noindent \textbf{MLP.} The MLP is for feature transformation and non-linearity:

\begin{equation}
    \operatorname{MLP}(X)=\sigma\left(X W_{1}+b_{1}\right) W_{2}+b_{2}
\end{equation}
where $W_{1} \in \mathbb{R}^{d \times d_{m}}$ and $W_{2} \in \mathbb{R}^{d_{m} \times d}$ are weights of the two fully-connected layers, $b_{1} \in \mathbb{R}^{d_{m}}$ and $b_{2} \in \mathbb{R}^{d}$ are the bias terms, and $\sigma(\dots)$ is the non-linear activation function such as ReLU~\cite{glorot2011deep} or GELU~\cite{hendrycks2016gaussian}.

\noindent \textbf{Add \& Dropout \& LN.} The output from MSA (or MLP) is added with the input of MSA (or MLP) to enforce the skip connection. Then, a dropout layer is used, followed by the Layer Normalization (LN) that enables stable training and faster convergence. Layer normalization is applied over each sample $x \in \mathbb{R}^{d}$ as follows:
\begin{equation}
    LN(x)=\frac{x-\mu}{\delta} \circ \gamma+\beta
\end{equation}
where $\mu \in \mathbb{R}$ and $\delta \in \mathbb{R}$ are the mean and standard deviation of the features, respectively, $\circ$ denotes the element-wise multiplication, $\gamma \in \mathbb{R}^{d}$ and $\beta \in \mathbb{R}^{d}$ are learnable affine transform parameters for scaling and shifting, respectively.

Given the input $X$, the computations in order in the Transformer encoder are: MSA, Add \& Dropout \& LN, MLP, and Add \& Dropout \& LN.

\subsection{Keypoint Initial Representation} 
\label{subsec:entity_repre}

Keypoint is the finest-grained actor-related entity that we consider for GAR. In this subsection, we describe the keypoint initial representation (i.e., representations of the input keypoint tokens of the first \mtx layer). 
As described in the main paper, representations of actor-related tokens in coarser scales are learned and aggregated from that of the finer scales.

\begin{figure}[t]
	\centering
	\includegraphics[scale=0.3]{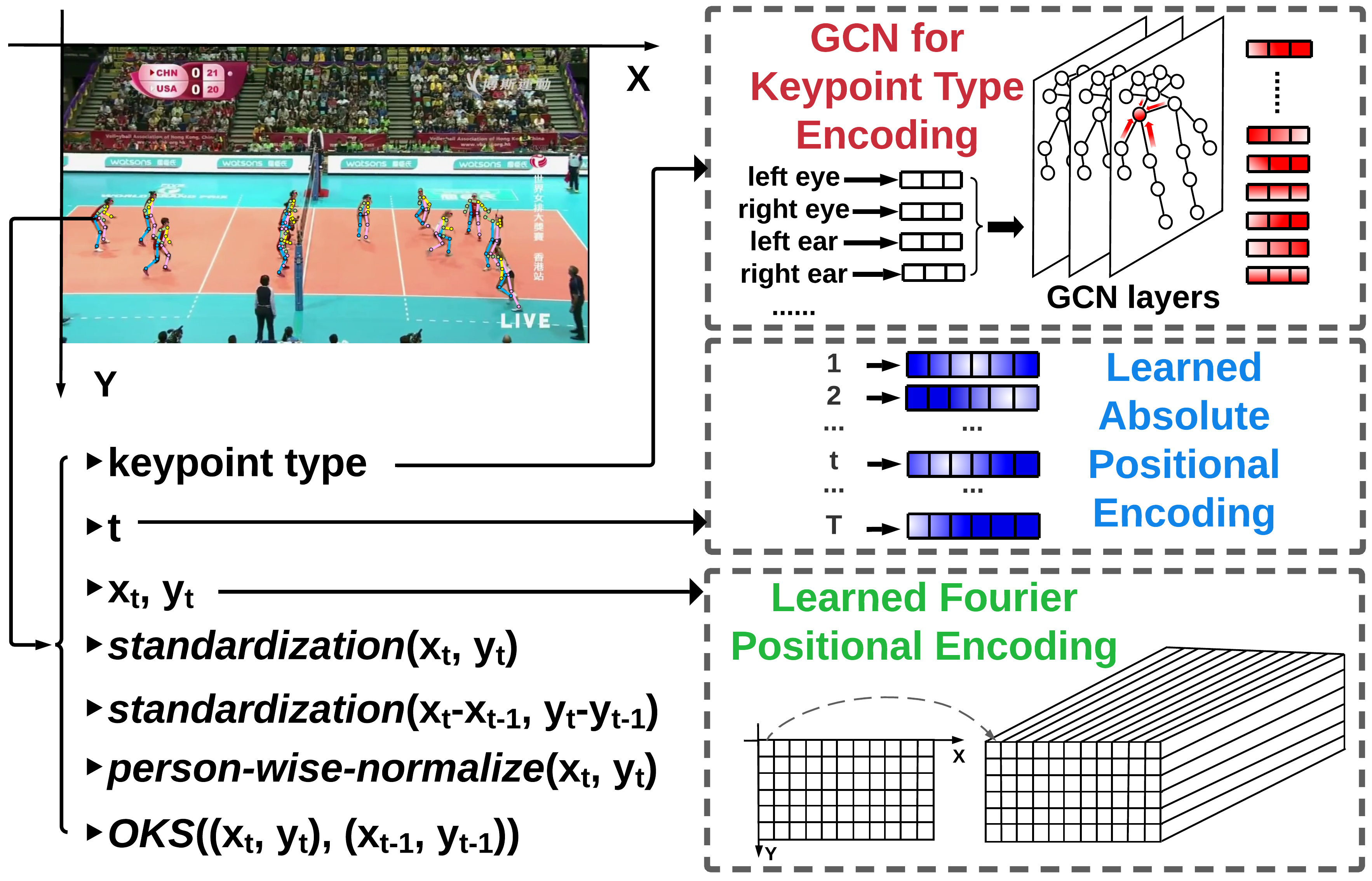}
	\caption{
	Representation of a person keypoint 
	in a particular frame.
}
	\label{fig:keypoint_repr} 
\end{figure}

\begin{figure*}[t]
	\centering
	\includegraphics[scale=0.14]{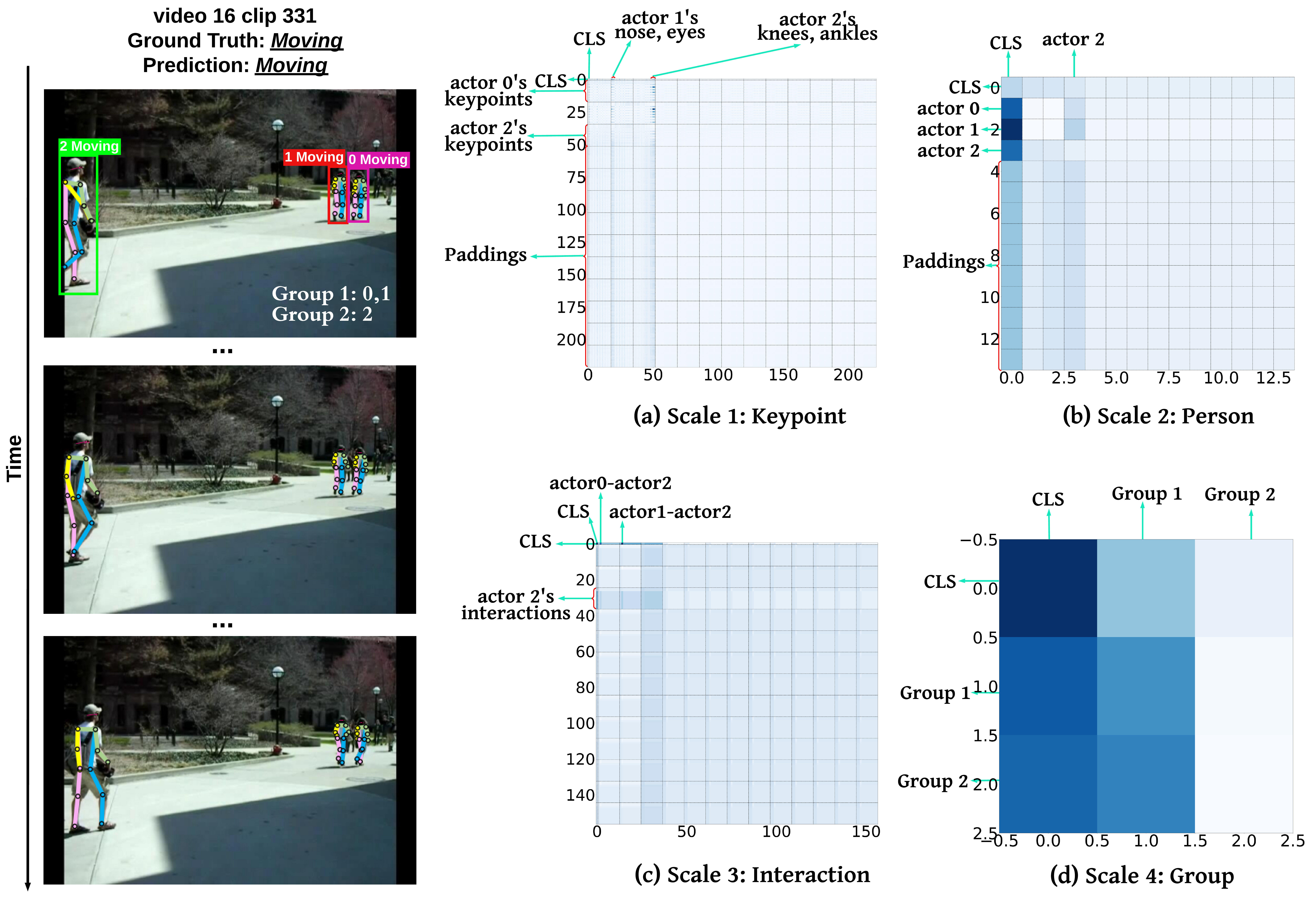} 
	\caption{\textbf{Qualitative results of \ours on CAD} -- showcasing attention matrices of a test set instance in the ``\textbf{moving}'' class.  
}
	\label{fig:CAD_good_moving} 
\end{figure*}

\begin{figure*}[t]
	\centering
	\includegraphics[scale=0.14]{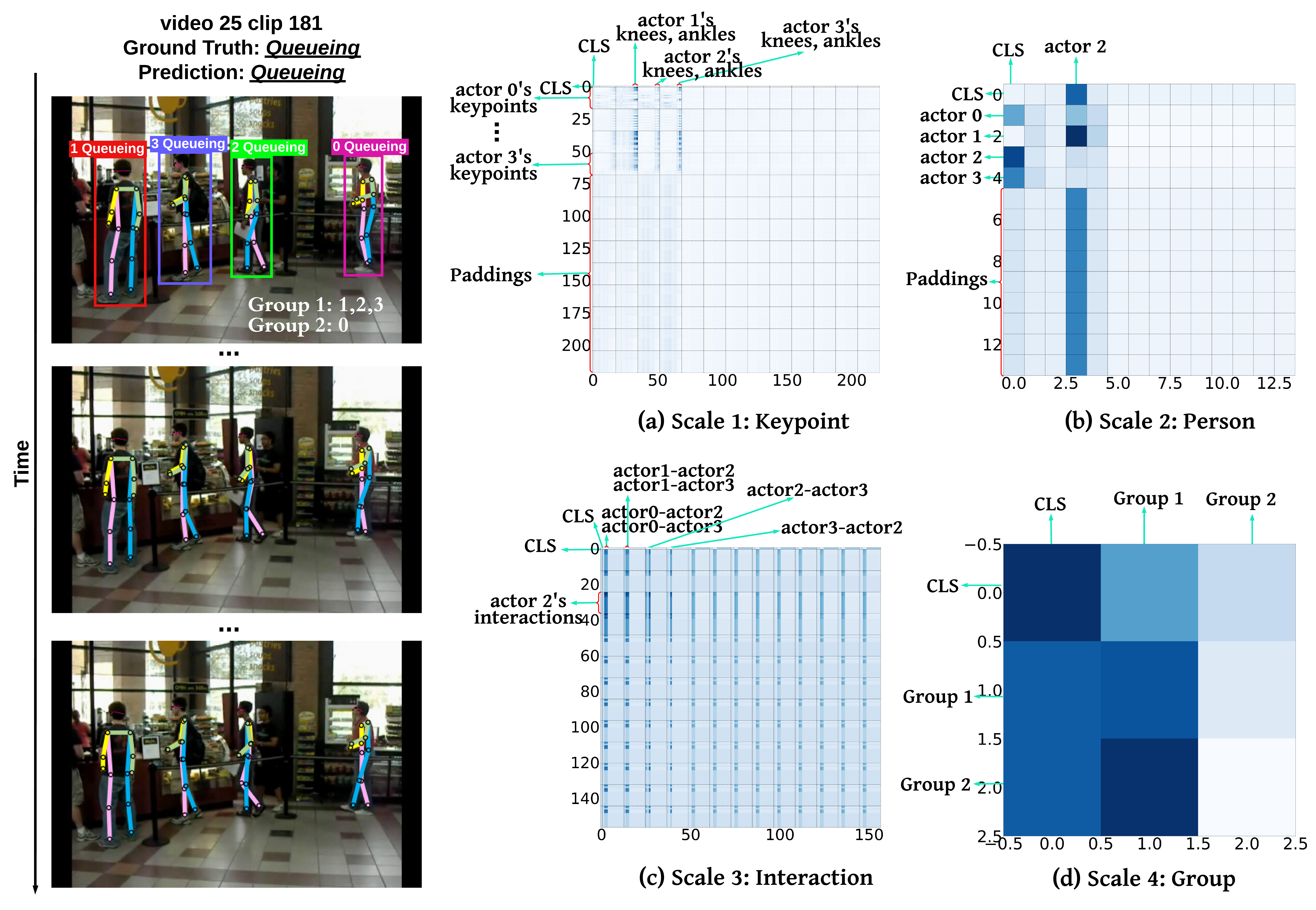} 
	\caption{\textbf{Qualitative results of \ours on CAD} -- showcasing attention matrices of a test set instance in the ``\textbf{quequeing}'' class.
}
	\label{fig:CAD_good_quequeing} 
\end{figure*}

A keypoint in a frame has the information of  keypoint type, as well as keypoint coordinate in both of the image space and the time space.
We use three GCN~\cite{kipf2016semi} layers to map each person keypoint type into a learned vector in order to encode the intrinsic connections of different keypoint types. 
We apply feature standardization 
to the raw $2$D keypoint coordinate and the 
temporal 
difference of coordinates in consecutive two frames.
We also
normalize the keypoint coordinate in a person-wise manner to account for rotation, translation, and scale differences~\cite{zappardino2021learning,sun2020view}. 
In addition, we use the Learned Fourier Positional Encoding~\cite{tancik2020fourier} to map each image coordinate into a learned vector, and use the Learned Absolute Positional Encoding to learn a vector for each time coordinate. 
To mitigate the issue of noisy estimated keypoints, we use the temporal
\textit{Object Keypoint Similarity} (OKS) proposed in~\cite{snower202015}, 
and use the mean  
OKS scores of each person as additional 
features. 
The above procedure 
is summarized in Fig.~\ref{fig:keypoint_repr}.
These features are concatenated to form the initial composite representation of a person keypoint 
in a  
frame, and a keypoint token is represented by concatenation and Feed Forward Network (FFN) based transformation of keypoints' representations in all timestamps.

\subsection{Implementation Details of \textbf{\ours}}
For annotation parsing and video preprocessing (e.g., obtaining clips from a long CAD video), we follow prior works~\cite{yuan2021spatio,arg}.
On the Volleyball dataset, annotations of group activity, 
players' bounding boxes and their actions in the middle frame of each clip are provided. 
Person tracklet data is provided by ~\cite{sendo2019heatmapping}.
For the Collective Activity dataset, annotations include actors’ bounding boxes and their action labels on the center frame of every ten frames and group activity labels for every ten frames. 
In order to make a fair comparison with related works~\cite{yuan2021learning,yuan2021spatio,actor-transformer,arg}, we use $T=10$ frames as the input to our model for both training and testing on both datasets.

We use HRNet~\cite{hrnet}\footnote{\url{https://github.com/leoxiaobin/deep-high-resolution-net.pytorch}} to obtain estimated person keypoints following~\cite{actor-transformer,yuan2021learning} ($j^{\prime}=17$).
There are a total of $17$ different keypoint types: nose, left eye, right eye, left ear, right ear, left shoulder, right shoulder, left elbow, right elbow, left wrist, right wrist, left hip, right hip, left knee, right knee, left ankle and right ankle. 
On the Volleyball dataset,
the maximal number of actors in a video $p^{\prime}=12$, and on Collective Activity, $p^{\prime}=13$.
On both datasets, the number of person groups per video $g^{\prime}=2$.
On Volleyball, actors are grouped into the two person groups by heuristics, i.e., according to the horizontal positions of the actors,
the left most $6$ actors form a sub-group and the rest actors form the other sub-group. 
We find that using clustering algorithms such as K-means (given the coordinates of the actors as features) can generate similar results as the heuristics on the Volleyball dataset. Hence, we choose to use the heuristics for simplicity for Volleyball,
and use K-means\footnote{\url{https://github.com/subhadarship/kmeans_pytorch}} to form person groups on the Collective Activity dataset (given the input of the learned person representations from \ourseos).

On the Volleyball dataset, the object keypoints are from the ball trajectories annotated by~\cite{GIRN} 
($e^{\prime}=1$)\footnote{For real-world data, one can resort to ball trajectory extraction~\cite{balltracktool,patrick2021keeping} for sports videos or object keypoint detection tools~\cite{huangyi,blomqvist2022semi,lu2021few}. }. The initial representation of the object  keypoint is similar to that of the person keypoint, i.e., a concatenation of the time positional encoding, Fourier positional encoding, and standardized keypoint coordinate and temporal difference. The initial object token is formed by concatenation and FFN-based transformation of object keypoints’ representations in all timestamps.
On the other hand, we do not use the object token for the Collective Activity dataset because the Collective Activity dataset does not have any human-object interactions.

On both datasets, the number of \mtx blocks $M$ is set as 2, the number of attention heads of the Transformer encoder at each scale is set to $2$, $8$, $2$ and $2$, respectively, and the dropout rate of the Transformer encoder at each scale is set to $0.5$, $0.2$, $0.2$ and $0$, respectively. We find that a smaller dropout rate in the coarser scales tends to yield a better performance. The dimension of the MLP layer in all Transformer encoders is set to $1024$ (i.e., $d_m=1024$), and the non-linear activation function is ReLU. The hidden dimension $d=256$ ($d_k$, $d_v$ and $d$ are equal) on Volleyball and $d=128$ on Collective Activity.
Because we focus on semantic relational reasoning over temporal relation capturing,
we use MLPs with $1$ hidden layer of dimensionality $d$ to flatten out the time axis
for each entity -- in this way, track-based representations are formed for each entity.
To aggregate the token representations from a finer scale to the coarser scale, 
FFNs
are used when the number of tokens for aggregation 
at that scale 
is fixed (e.g., $2$ persons aggregates to an interaction),
otherwise
summation is used (e.g.,
the number of persons to form a group varies on the Collective Activity dataset due to 
K-means).
The cross entropy loss is used during training for both group activity and person action classification.
We use the Adam optimizer~\cite{kingma2014adam} and train the model for $45$ epochs with an initial learning rate $0.0005$ and decrease the learning rate to $0.0001$ at epoch $40$. The weight decay is $0.001$, $\lambda$ is $3$, and batch size is $256$.
Following~\cite{swav}, the temperature parameter
$\tau=0.1$, $\varepsilon=0.05$, and the number of iterations of the Sinkhorn-Knopp algorithm~\footnote{\url{https://github.com/facebookresearch/swav}}
is set to $3$. The number of prototypes $K$ is $1000$ for all experiments except the ones described in Appendix~\ref{appendix_sec:diff_num_protos}. In data augmentations, the range of random perturbation is set to $1$ pixel location.
We use the PyTorch\footnote{\url{https://pytorch.org}} Python library.

\subsection{Implementation Details of Ablation Studies}
\label{appendix_subsec:ablation_imple_details}

In this subsection, we describe the methodology of the ablations that we present in Appendix~\ref{appendix_sec:ablation_study} in details. For all of the ablations, we use the same set of hyper-parameters as our full model. 

\noindent \textbf{No Clustering}. The only difference from this ablation to our full model \ours is the loss function used in training. Instead of using Equation (\red{6}) (in the main paper) as the loss function, this ablation uses the following loss function:
\begin{equation}
    \mathcal{L}_{\text {total}} = \sum_{m=1}^{M-1}\mathcal{L}_{\text {groupAux}} + \lambda \left(\mathcal{L}_{\text {groupLast}} + \mathcal{L}_{\text {person}}\right)
\end{equation}

\noindent \textbf{Label Consistency~\cite{shi2020online} for Scale Agreement}. Similar to the previous ablation, the only difference from this ablation to our full model \ours is the loss function used in training, which is formulated as follows:
\begin{equation}
    \mathcal{L}_{\text {total}} = \sum_{m=1}^{M-1}\mathcal{L}_{\text {groupAux}} + \lambda \left(\mathcal{L}_{\text {groupLast}} + \mathcal{L}_{\text {person}} + \mathcal{L}_{\text {consistency}}\right)
\end{equation}
where $\mathcal{L}_{\text {consistency}}$ represents the loss term that minimizes the $L2$ distance between $2$ scales of a clip in the logit space. Specifically, for every two pairs of scales, we compute the $L2$ loss given the two sets of GAR logits of the two scales. The $\mathcal{L}_{\text {consistency}}$  term is the mean of such $L2$ losses over all pairs of scales.

\noindent \textbf{$1$-Scale: Keypoint}. 
For this ablation, there is only one Transformer encoder in the \mtx block, and the tokens to the Transformer encoder are:
\begin{equation}
\label{eq:tokens_1_scale}
\begin{aligned}
& \text{\textit{Scale 1:}}  \quad \left\{\left[\text{\texttt{CLS}}\right], \mathbf{e}_1, \cdots, \mathbf{e}_{e^{\prime}}, \mathbf{k}_1^1, \cdots, \mathbf{k}_{p^{\prime}}^{j^{\prime}} \right\}.
\end{aligned}
\end{equation}
Since only the keypoint tokens are refined by the \mtx block in this ablation and there is only $1$ scale, this ablation uses the following loss function:
\begin{equation}
    \mathcal{L}_{\text {total}} = \sum_{m=1}^{M-1}\mathcal{L}_{\text {groupAux}} + \lambda\mathcal{L}_{\text {groupLast}}
\end{equation}  

\noindent \textbf{$2$-Scale: Keypoint $+$ Person}. 
For this ablation, there are two hierarchical scales in the \mtx block, and the tokens to the Transformer encoders are:
\begin{equation}
\label{eq:tokens_2_scales}
\begin{aligned}
& \text{\textit{Scale 1:}}  \quad \left\{\left[\text{\texttt{CLS}}\right], \mathbf{e}_1, \cdots, \mathbf{e}_{e^{\prime}}, \mathbf{k}_1^1, \cdots, \mathbf{k}_{p^{\prime}}^{j^{\prime}} \right\},\\
& \text{\textit{Scale 2:}}  \quad  \left\{\left[\text{\texttt{CLS}}\right], \mathbf{e}_1, \cdots, \mathbf{e}_{e^{\prime}}, \mathbf{p}_1, \cdots, \mathbf{p}_{p^{\prime}} \right\}.
\end{aligned}
\end{equation}
This ablation uses the same loss function as our full model except that the number of pairs of scales for swapped prediction 
is only $1$, i.e., pair scale$1$-scale$2$.

\noindent \textbf{$3$-Scale: Keypoint $+$ Person $+$ Interaction}. 
For this ablation, there are three hierarchical scales in the \mtx block, and the tokens to the Transformer encoders are:
\begin{equation}
\label{eq:tokens_3_scales}
\begin{aligned}
& \text{\textit{Scale 1:}}  \quad \left\{\left[\text{\texttt{CLS}}\right], \mathbf{e}_1, \cdots, \mathbf{e}_{e^{\prime}}, \mathbf{k}_1^1, \cdots, \mathbf{k}_{p^{\prime}}^{j^{\prime}} \right\},\\
& \text{\textit{Scale 2:}}  \quad  \left\{\left[\text{\texttt{CLS}}\right], \mathbf{e}_1, \cdots, \mathbf{e}_{e^{\prime}}, \mathbf{p}_1, \cdots, \mathbf{p}_{p^{\prime}} \right\},\\
& \text{\textit{Scale 3:}}  \quad  \left\{\left[\text{\texttt{CLS}}\right], \mathbf{e}_1, \cdots, \mathbf{e}_{e^{\prime}}, \mathbf{i}_1, \cdots, \mathbf{i}_{{p^{\prime}}\times({p^{\prime}}-1)} \right\}.
\end{aligned}
\end{equation}
This ablation uses the same loss function as our full model except that the number of pairs of scales for swapped prediction 
is $3$, i.e., pair scale$1$-scale$2$, pair scale$1$-scale$3$, and pair scale$2$-scale$3$.

\noindent \textbf{No Auxiliary Prediction}. The only difference from this ablation to our full model \ours is the loss function used in training. 
This ablation uses the following loss function:
\begin{equation} 
    \mathcal{L}_{\text {total}} = \mathcal{L}_{\text {groupLast}} + \mathcal{L}_{\text {person}} + \mathcal{L}_{\text {cluster}}
\end{equation}

\noindent \textbf{No \mtxeos}. In this ablation, the group activity classifier simply takes features of the \textit{initial} object token and person tokens as inputs, and the person tokens are aggregated from the \textit{initial} representations of keypoint tokens through concatenation and FFN. In addition, features of the person tokens are the inputs to the person action classifier.
No Transformers are used in this ablation.  Due to the lack of relational reasoning performed at multiple scales, this ablation uses the following loss function:
\begin{equation}
    \mathcal{L}_{\text {total}} = \mathcal{L}_{\text {groupLast}} + \mathcal{L}_{\text {person}}
\end{equation}

\subsection{Miscellaneous}

The Original split of the Volleyball dataset allows the GAR method to leverage scene biases in order to achieve a high accuracy, on the other hand, the Olympic split can better test the model generalization ability. None of the exsting RGB-baesd GAR methods have performed experiments using the Olympic split. 
To obtain the results of prior RGB-based GAR methods on the Olympic split of the Volleyball dataset (Table 1 in the main paper), our implementations of 
PCTDM~\cite{PCTDM}, SACRF~\cite{pramono2020empowering}, AT~\cite{actor-transformer}, ARG~\cite{arg}, TCE-STBiP~\cite{yuan2021learning}  and  DIN~\cite{yuan2021spatio} are based on the public available codebase\footnote{\url{https://github.com/JacobYuan7/DIN-Group-Activity-Recognition-Benchmark}}\footnote{\url{https://github.com/wjchaoGit/Group-Activity-Recognition}} and we have verified the implementations through obtaining the results of these methods on the Original split of the Volleyball dataset and then comparing with the reported results from authors of each method. For VGG-16~\cite{simonyan2014very}, we use RoIAlign~\cite{he2017mask}\footnote{\url{https://github.com/longcw/RoIAlign.pytorch}} to obtain the person regional features and then use these person features to predict the individual actions and the group activity. 
All of these RGB-based methods (including our \ours RGB-based variant) use the RGB-only modality and share the same VGG-16 backbone.

\renewcommand{\thesection}{G}
\section{Extended Discussion on Related Work}
\label{appendix_sec:related}

\noindent \textbf{Group Activity Recognition.}
Early work on GAR relies on handcrafted features~\cite{choi2009they,lan2011discriminative,hajimirsadeghi2015visual,cheng2010group,choi2011learning,choi2012unified,nabi2013temporal}; yet notable progress 
has been made
by Deep Learning (DL) based approaches~\cite{ibrahim2016hierarchical,deng2016structure}. 
We review DL-based methods and refer readers to the comprehensive review of GAR presented in~\cite{wu2021comprehensive}.

Early DL-based 
methods use Convolutional Neural Networks (CNNs) to extract 
the low-level visual 
features and then apply 
Recurrent Neural Networks
such as LSTM~\cite{hochreiter1997long} 
for temporal modeling~\cite{wang2017recurrent,shu2017cern,li2017sbgar,kim2018discriminative}.
Since learning inter-person interactions is essential for GAR~\cite{wu2021comprehensive}, much of the recent research explores how to capture the contextual information about the actor and their relations~\cite{ibrahim2018hierarchical,azar2019convolutional,arg,prl,pramono2020empowering}.
Several works tackle this problem from a graph-based perspective~\cite{ibrahim2018hierarchical,lu2019gaim,sam,higcin}
such as applying Graph Convolutional Networks (GCNs)~\cite{kipf2016semi} 
for deep relationship modeling~\cite{arg}. 
More recent works utilize attention modeling~\cite{stagnet,xu2020group,lu2019gaim,yuan2021learning} including using 
Transformers~\cite{actor-transformer,GroupFormer}
to perform relational reasoning,
with a focus on determining the most critical persons~\cite{arg,actor-transformer,pramono2020empowering,yuan2021learning}, groups~\cite{ehsanpour2020joint,GroupFormer}, or
interactions~\cite{sam}.
Existing works 
in the field of GAR
have primarily used RGB- and/or optical-flow-based features with RoIAlign~\cite{he2017mask} to represent actors~\cite{higcin,stagnet,arg,ssu}.
A few recent works replace or augment these features with keypoints/poses of the actors~\cite{lu2019spatio,chen2019group,POGARS,GroupFormer}. 
Some 
only 
use
the numerical coordinate-based keypoint  representation~\cite{zappardino2021learning,POGARS,GIRN,pramono2020empowering} while others
use a high-dimensional vector  
from a deep pose backbone~\cite{actor-transformer,yuan2021learning} which is not as efficient.
 In this paper, we use Transformers~\cite{vaswani2017attention}
 for higher-order relationship modeling and use only 
 the light-weight coordinate-based
 keypoint representation. 
Our work differs from prior methods 
in that we propose a \mtx block to hierarchically reason about entities at different semantic scales and we aid learning group activities by improving the multiscale representations.

\noindent \textbf{Action Recognition and Keypoint-based Prediction.}
Action Recognition is one of the primary tasks in video understanding. 
There has been rapid progress
in recent years,
starting from recognition of the 
low-level 
atomic actions performed by an 
individual 
(e.g., hand-waving, dancing, jumping),
to paired-actions 
being 
acted by two persons~\cite{perez2021interaction,shu2019hierarchical,yun2012two,kong2012learning} (e.g., shaking hands, hugging, punching), 
towards group activities that encompass many actors at once~\cite{choi2009they,ibrahim2016hierarchical,ramanathan2016detecting,GIRN} (e.g., attack and defense in a sports game, pedestrians queuing).
Our paper focuses on the most spatially complex scenarios, where multiple interacting individuals form the group activity.
In addition, keypoint-based action recognition has drawn much attention~\cite{li2020dynamic,yan2018spatial,zhu2016co,nguyen2021geomnet,zhao2021learning}.
Keypoint-based representation can be regarded as a high-level representation for dynamic behaviors, and is preferred due to  
benefits such as being 
compact and robust to variations of viewpoints, appearances, and surrounding distractions~\cite{dang2020sensor,lin2020ms2l}. 
We study keypoint-based 
group activity recognition. 
We propose to use
techniques 
including auxiliary prediction and data augmentations 
that can 
aid 
learning group activity from the keypoint modality.

\noindent \textbf{Compositionality
and Multiscale Learning.}
Compositionality is an active field of research 
in computer vision (CV)~\cite{zhan2020multi,kim2020safcar,yan2020interactive,sun2021counterfactual}, 
natural language processing~\cite{socher2014grounded,vendrov2015order,irsoy2014deep,dankers2021paradox}
and machine reasoning~\cite{hudson2018compositional,hudson2019learning,bottou2014machine}.
In terms of understanding videos centered on human actions, compositionality can be studied from different lenses, e.g., through formulating an activity 
as compositions of atomic actions temporally~\cite{rai2021home,grunde2021agqa,chen2020rit,luo2021moma} or semantically~\cite{shao2020finegym,suris2021learning}, or decomposing actions by action-based aspects (verbs) and object components (noun)~\cite{ji2020action,jia2020lemma,materzynska2020something,le2021c,luo2021moma}.
We tackle compositional video understanding by formulating a visual-semantic hierarchy, where each 
semantic
hierarchy is regarded as representation of the video at 
a particular scale. Such an idea of multiscale learning has been a long-standing topic
in CV 
as well~\cite{li2020dynamic,li2016visual,gong2019cnn,haber2018learning}.
Recently, researchers 
have 
started to introduce the concept of multiscale learning to Transformers~\cite{fan2021multiscale,liu2021swin,han2021transformer} by operating self-attention over various scales of resolutions and/or channels, in order to obtain a multiscale pyramid of features often observed in CNNs.
Distinct from prior works, we design \ours
that models semantic scene entities at different hierarchical scales to learn group activities effectively.
\ours is the first Transformer-based method with \textit{explicit} multiscale modeling for GAR that improves the musicale representations
with a contrastive clustering based 
objective.

\renewcommand{\thesection}{H}
\section{Discussion of \ours}
\label{appendix_sec:discussion}

\subsubsection{Motivation}
Intuitively, each scale provides enough information for GAR;
only the information granularity varies, which causes recognition confidence 
to vary scale by scale. 
Therefore, we consider
scales as different but correlated views of the clip, and 
utilize multiscale contrastive clustering learning (MCCL) to allow one scale
to complement another.
This allows learning better compositional structures and
higher-order representations.

\ul{Pull close}: Equation 2 in the main paper (swapped prediction)
trains the model to produce multiscale representations
of 1 clip such that the 
cluster assignment
of the clip representation at one scale can be predicted from the clip representation at another scale, allowing representations of the same clip to be pulled close.

\ul{Pull away}:
If 2 clips are semantically different,
as 
Equation 6 in the main paper
includes 
the supervised GAR loss and the unsupervised MCCL loss, the 2 clips  will be put
to different clusters
and
pushed 
further 
as training goes.

\subsubsection{Key Insight} 
The key novelty of \ours is that the model learns consistent multiscale representations. The idea of considering scales as views and encouraging scale agreement is applicable to numerous Computer Vision tasks, including general-purpose visual model pre-training, because the entities and scales we have considered are common in human-centered videos. By design, \ours is capable of modeling multi-actor multi-object interactions in images or videos. Moreover, \ours offers numerous useful practices, including  auxiliary prediction to aid training stacks of Transformers, and techniques that can aid the model to  learn the high-level knowledge from the low-level coordinate-based keypoint signals 
(e.g. data augmentations with random perturbation and OKS-based keypoint features to mitigate the issue of noisy estimated keypoints).

\subsubsection{Limitation}
As shown in the failure cases of \ourseos, videos with severe occlusions remain challenging. Severe occlusion can be a limitation for all GAR methods as their modalities are derived from the RGB input. 
\ours might handle occlusion better than RGB-based methods. In partial occlusion scenarios (i.e., only a ratio of keypoints are occluded for the person -- examples are cases shown in Fig. 6 of the main paper, Fig.~\ref{fig:CAD_good_waiting} and Fig.~\ref{fig:good_l_spike} in the Appendix), because \ours learns human motion dynamics, better representation of occluded persons can be inferred from the keypoints. In addition, \ours is agnostic to the keypoint backbone, and many SOTA keypoint extractors are robust to occlusion (e.g, BlazePose~\cite{bazarevsky2020blazepose,mediapipeBlazePose}).

 \subsubsection{Societal Impact} 
Human group activity recognition has widespread societal implications in a variety of domains including security, surveillance, kinesiology, sports analysis, and rehabilitation. Privacy and ethical concerns might be raised when deployed in real-world settings
if not done in a careful manner.
In response to these concerns, \ours
utilizes only keypoint input and does not use any personally identifiable information for inferring group activity. 
Even for the backbone which \ours is agnostic to, there are existing works~\cite{hinojosa2021learning} that perform privacy-preserving pose estimation.
Hence, our method can prevent the sensor camera from acquiring detailed visual data that may contain private or biased information of users.

\clearpage

\clearpage
\begin{figure*}[t]
	\centering
	\includegraphics[scale=0.14]{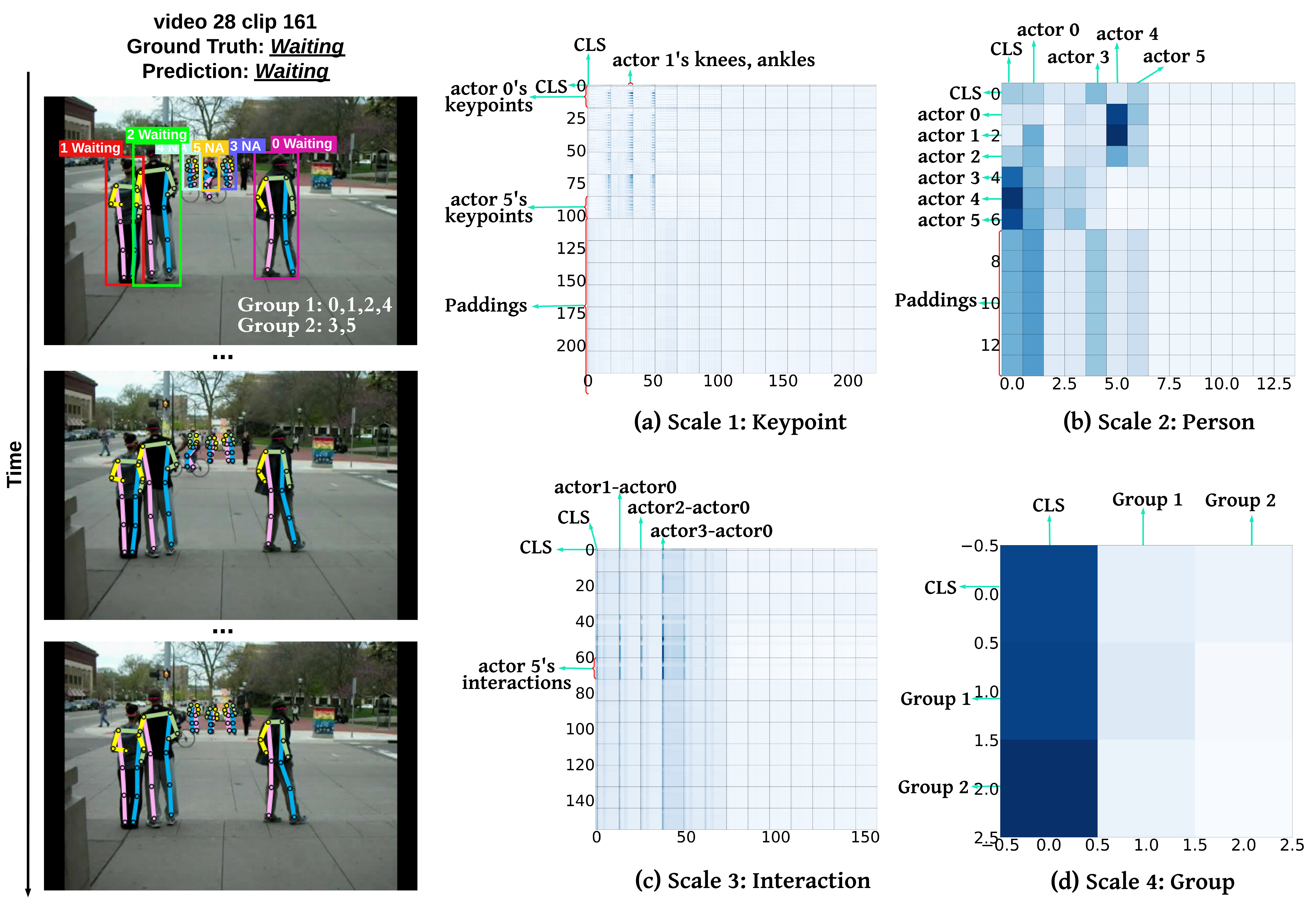} 
	\caption{\textbf{Qualitative results of \ours on CAD} -- showcasing attention matrices of a test set instance in the ``\textbf{waiting}'' class. It is noteworthy that the keypoints of actor $4$ and $5$ are noisy due to occlusion; yet \ours makes a correct prediction.
}
	\label{fig:CAD_good_waiting} 
\end{figure*} 

\begin{figure*}[t]
	\centering
	\includegraphics[scale=0.14]{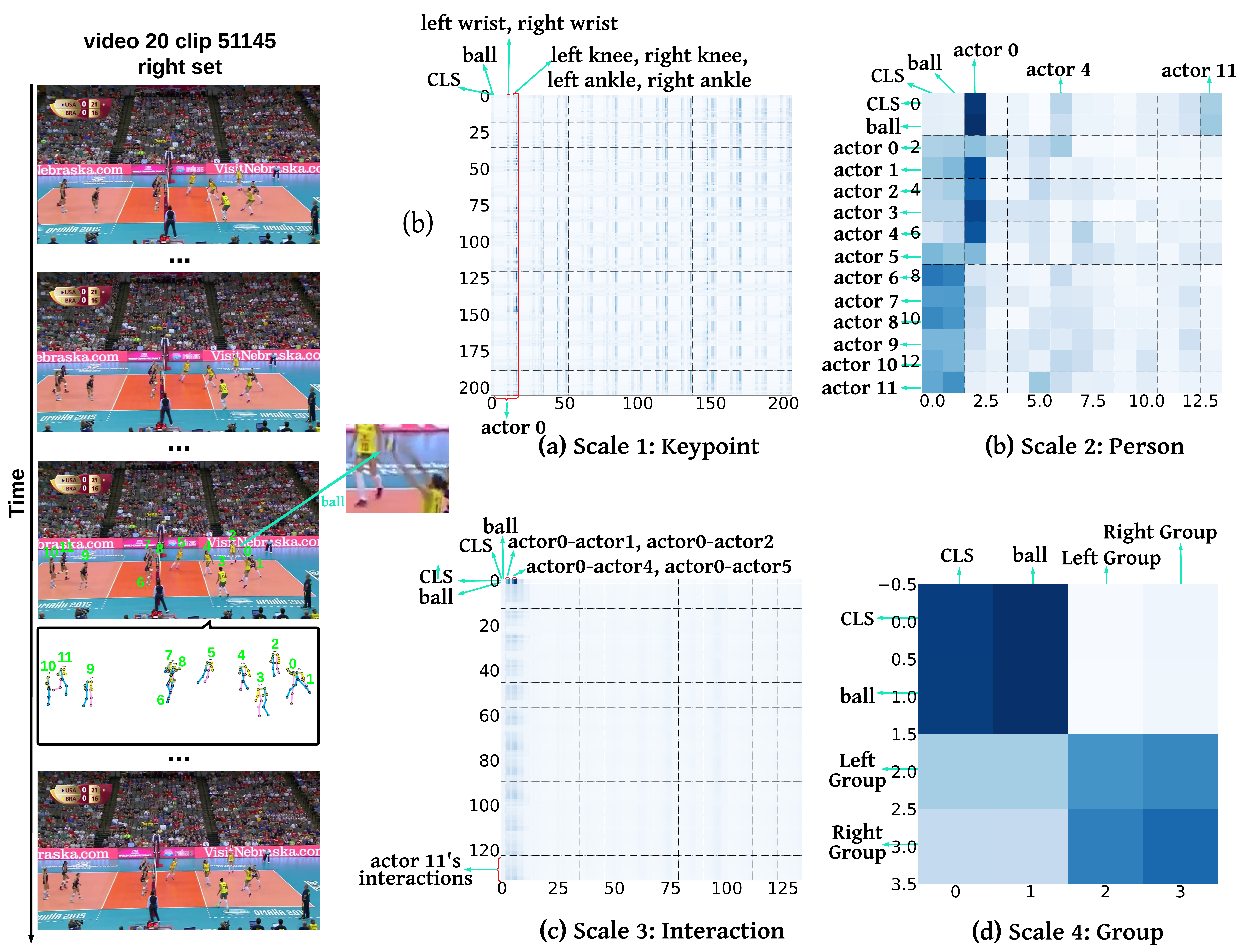} 
	\caption{\textbf{Qualitative results of \ours on VD} -- showcasing attention matrices of a test set instance in the ``\textbf{right set}'' class (key actor is actor 0).  
}
	\label{fig:good_r_set} 
\end{figure*}
  
\begin{figure*}[t]
	\centering
	\includegraphics[scale=0.14]{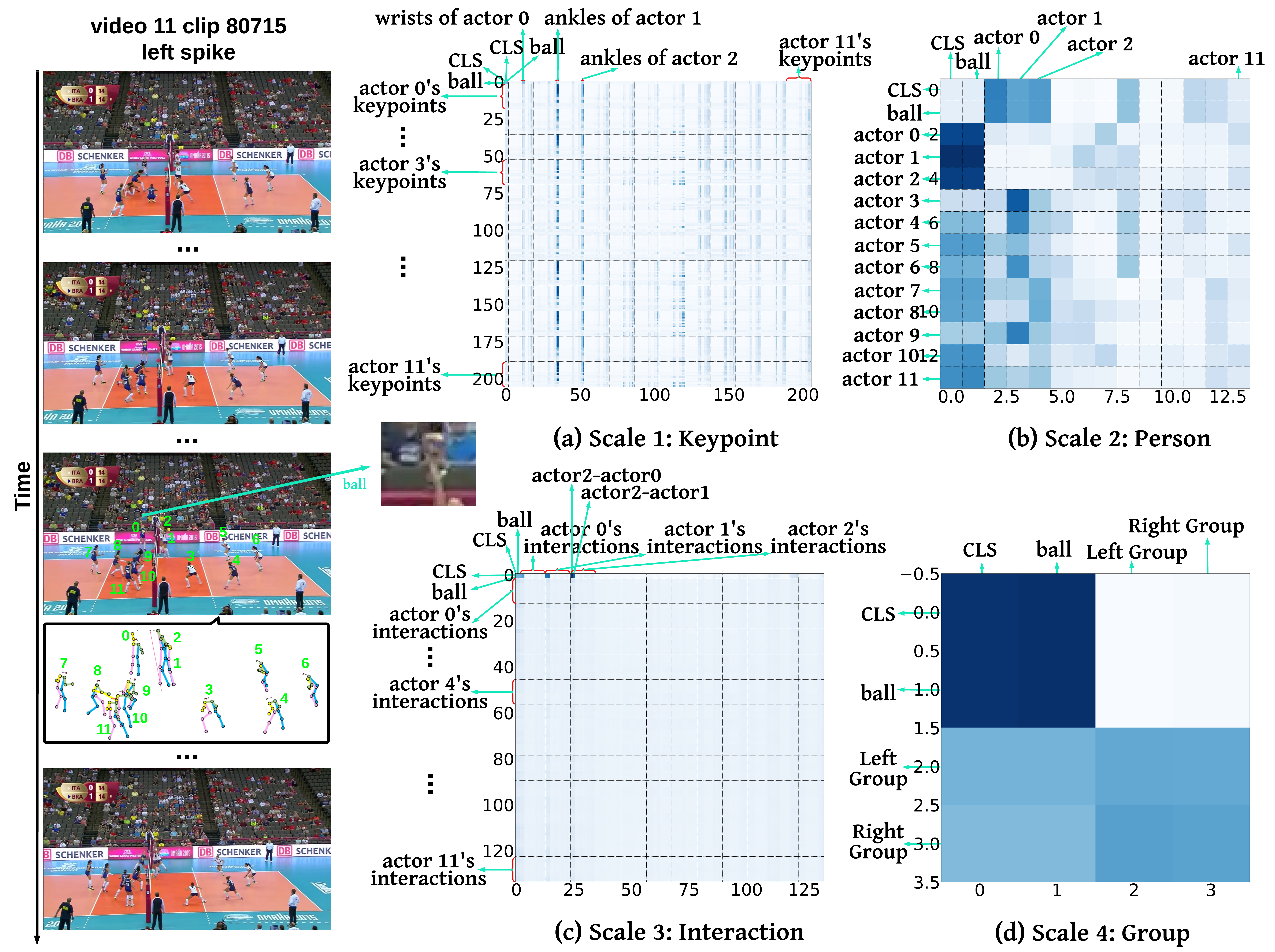} 
	\caption{\textbf{Qualitative results of \ours on VD} -- showcasing attention matrices of a test set instance in the ``\textbf{left spike}'' class (key actor is actor 0). 
}
	\label{fig:good_l_spike} 
\end{figure*}  

\begin{figure*}[t]
	\centering
	\includegraphics[scale=0.14]{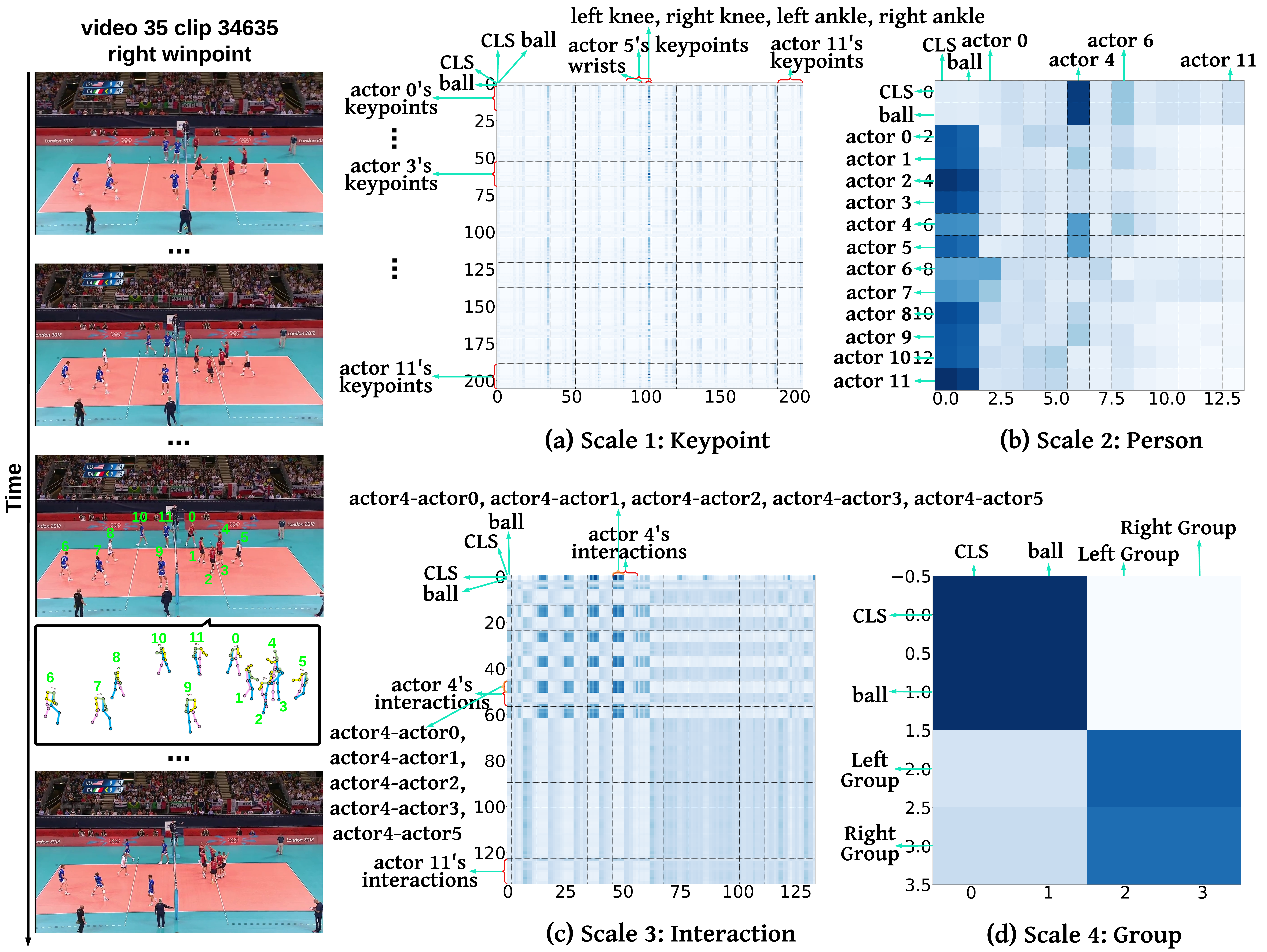} 
	\caption{\textbf{Qualitative results of \ours on VD} -- showcasing attention matrices of a test set instance in the ``\textbf{right winpoint}'' class.  
}
	\label{fig:good_r_winpoint} 
\end{figure*}
\clearpage
\begin{figure*}[t]
	\centering
	\includegraphics[scale=0.14]{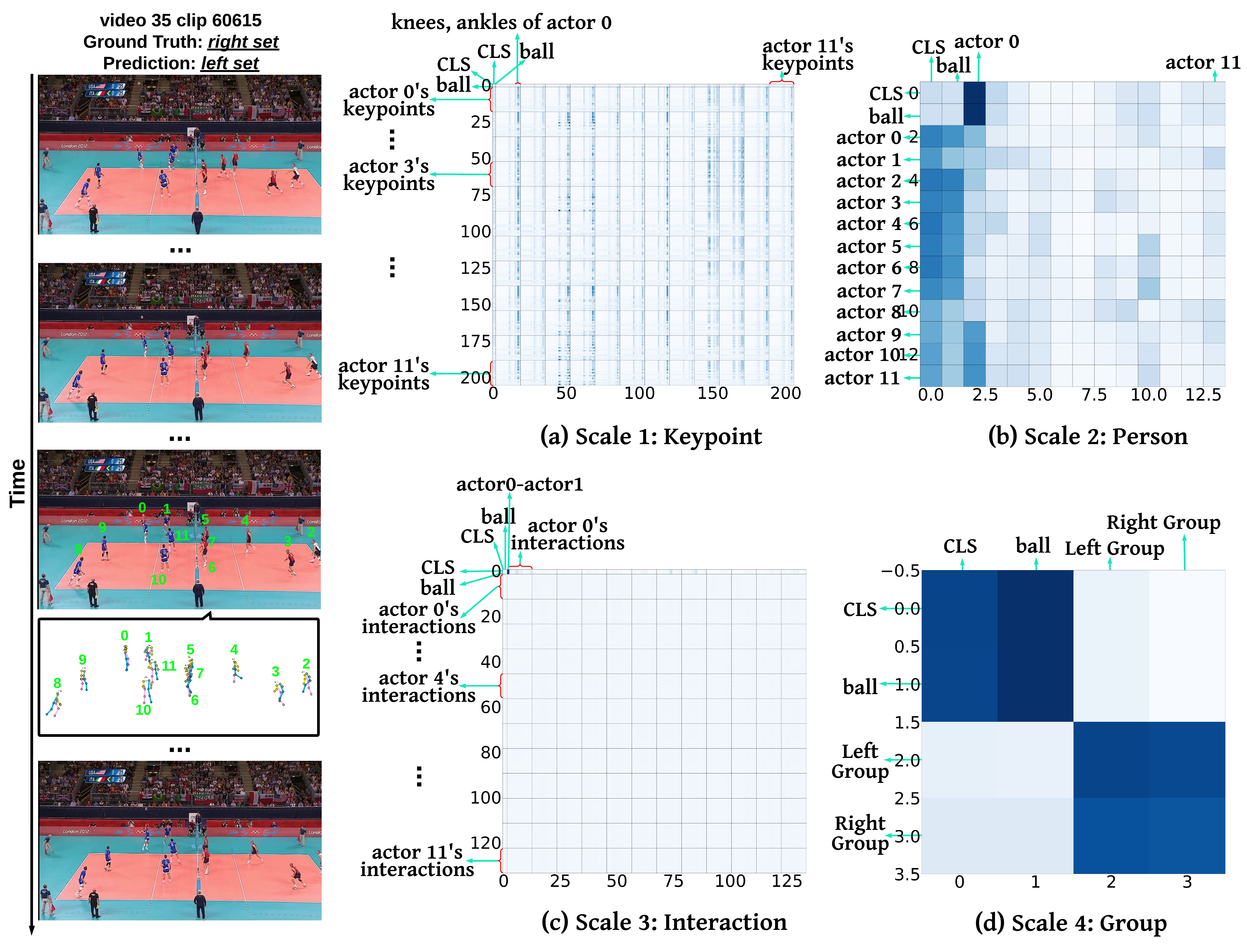} 
	\caption{\textbf{Mislabeled test set clip example of VD}. The annotated label is wrong but the prediction from \ours is correct (key actor is actor $0$). 
}
	\label{fig:failure_label_left_right_wrong} 
\end{figure*}

\begin{figure*}[t]
	\centering
	\includegraphics[scale=0.14]{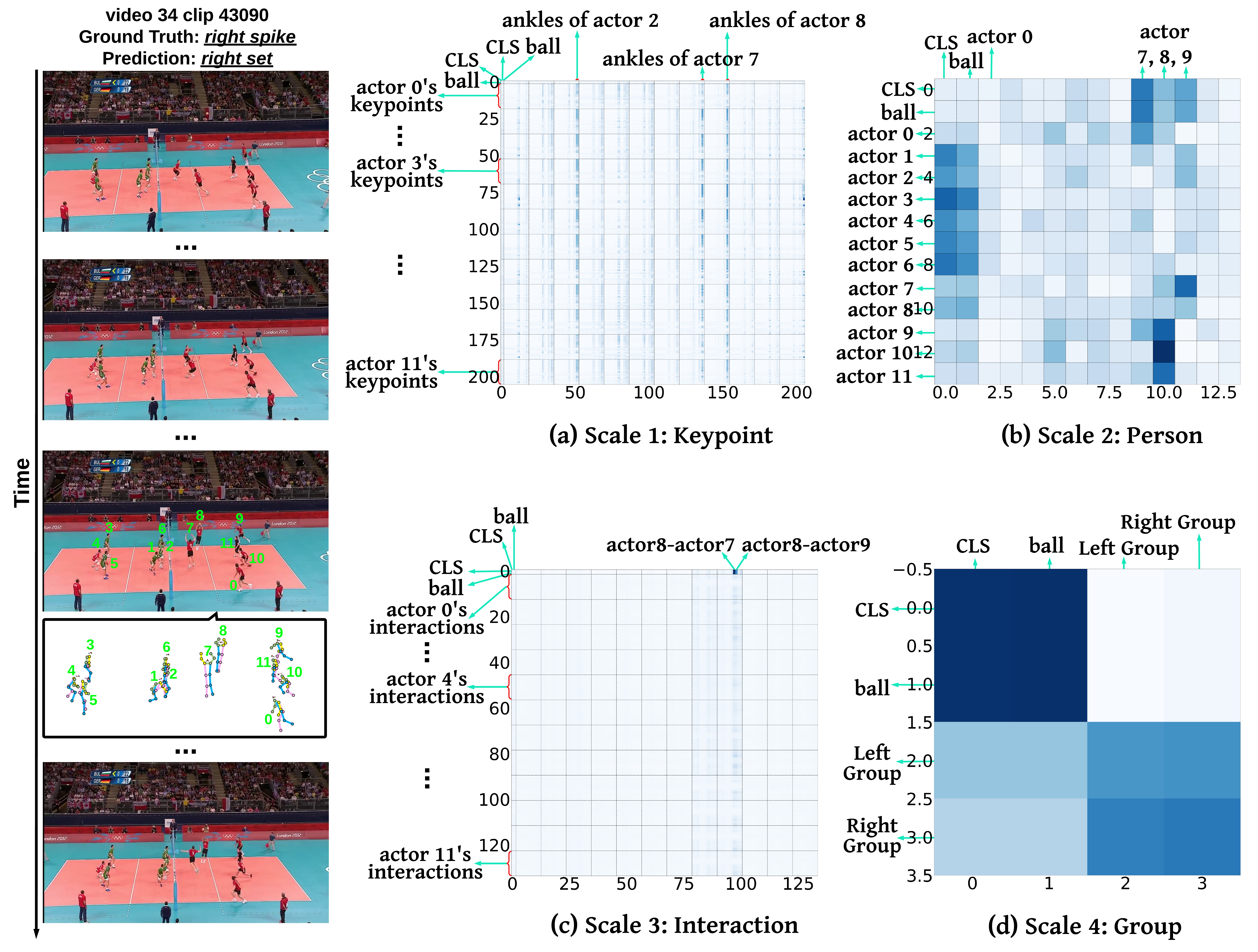}
	\caption{\textbf{Mislabeled test set clip example of VD}. The annotated label is wrong but the prediction from \ours is correct. 
	Actor $8$ is performing the key action \textit{setting} and interacting with the ball. Actor $7$ is performing an action very similar to \textit{spiking} but  actor $7$ is missing interaction with the ball.  
}
	\label{fig:failure_spike_to_set} 
\end{figure*}
\clearpage

\begin{figure*}[t]
	\centering
	\includegraphics[scale=0.14]{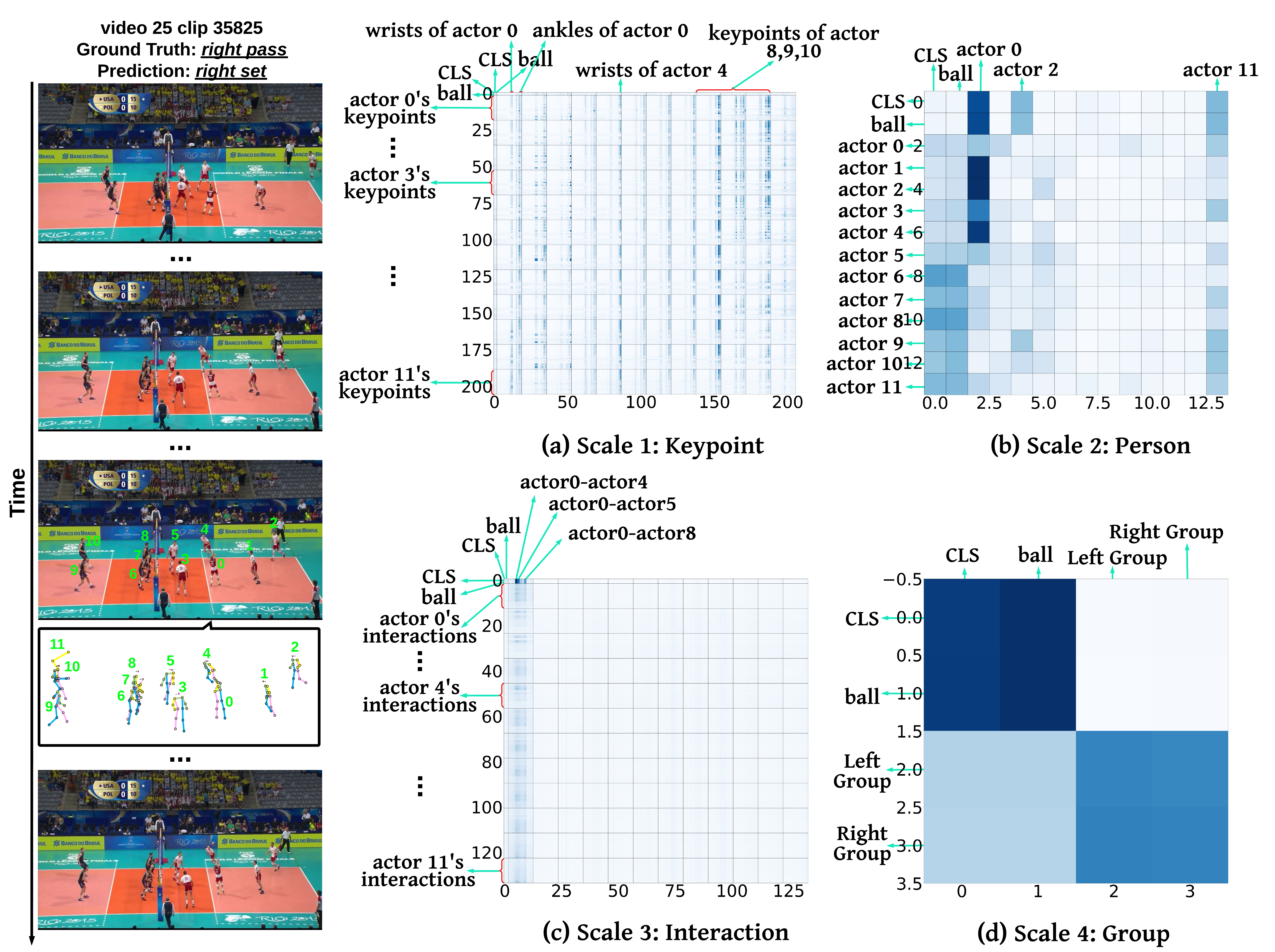} 
	\caption{\textbf{A failure case on VD}.  
	Actor $0$ is the one performing the key action.  
	At scale $2$ and $3$, \ours successfully identifies tokens associated with the key person as the most important tokens. However, at scale $1$, \ours fails to focus on keypoints of actor $0$ and eventually makes a wrong group activity prediction.    
}
	\label{fig:failure_pass_to_set} 
\end{figure*}

\begin{figure*}[t]
	\centering
	\includegraphics[scale=0.14]{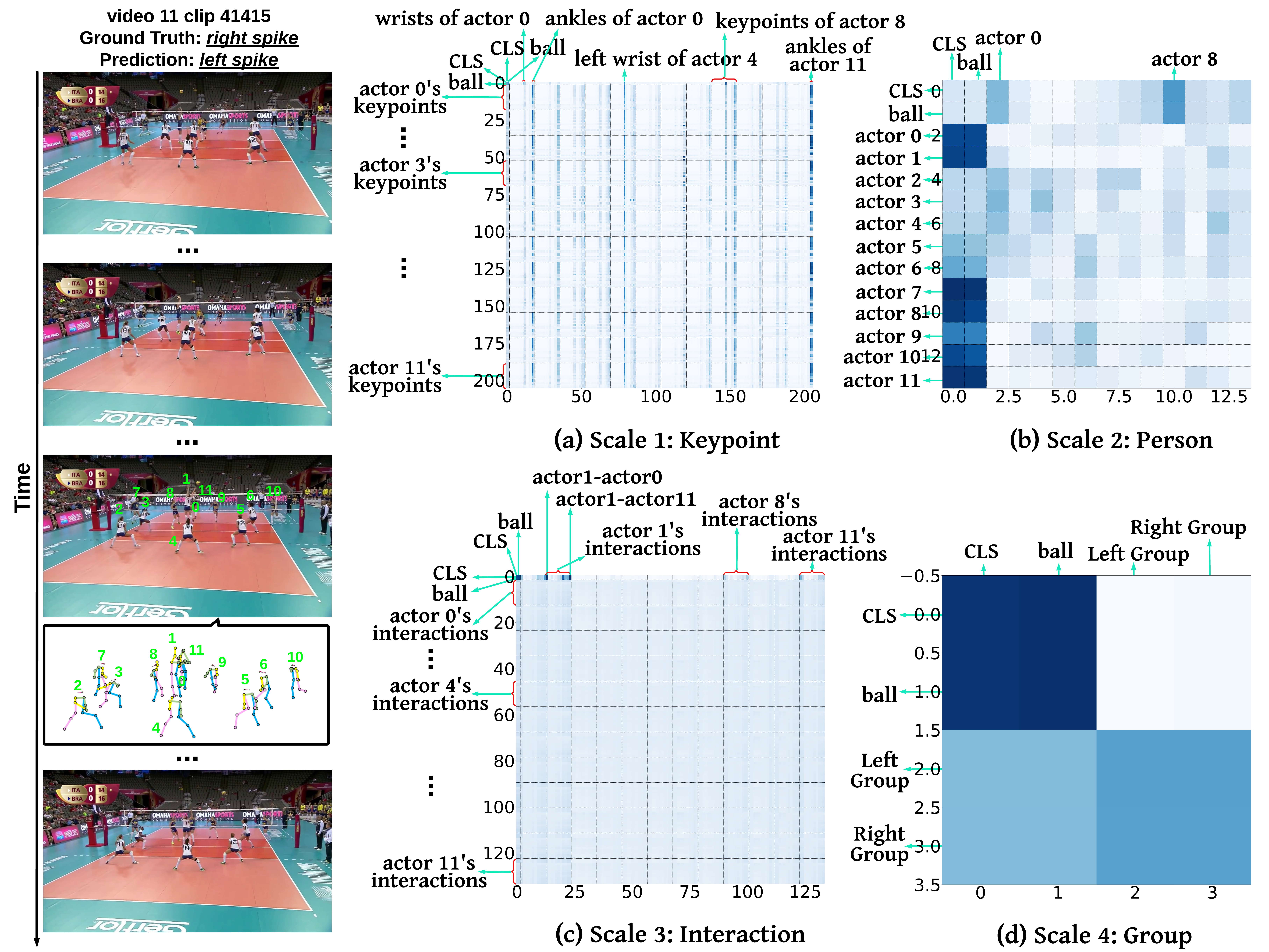} 
	\caption{\textbf{A failure case on VD}. Camera position of this clip is different from other clips in VD, and \ours fails to distinguish which team performs the activity. 
}
	\label{fig:failure_camera_change} 
\end{figure*}

\begin{figure*}[t]
	\centering
	\includegraphics[scale=0.14]{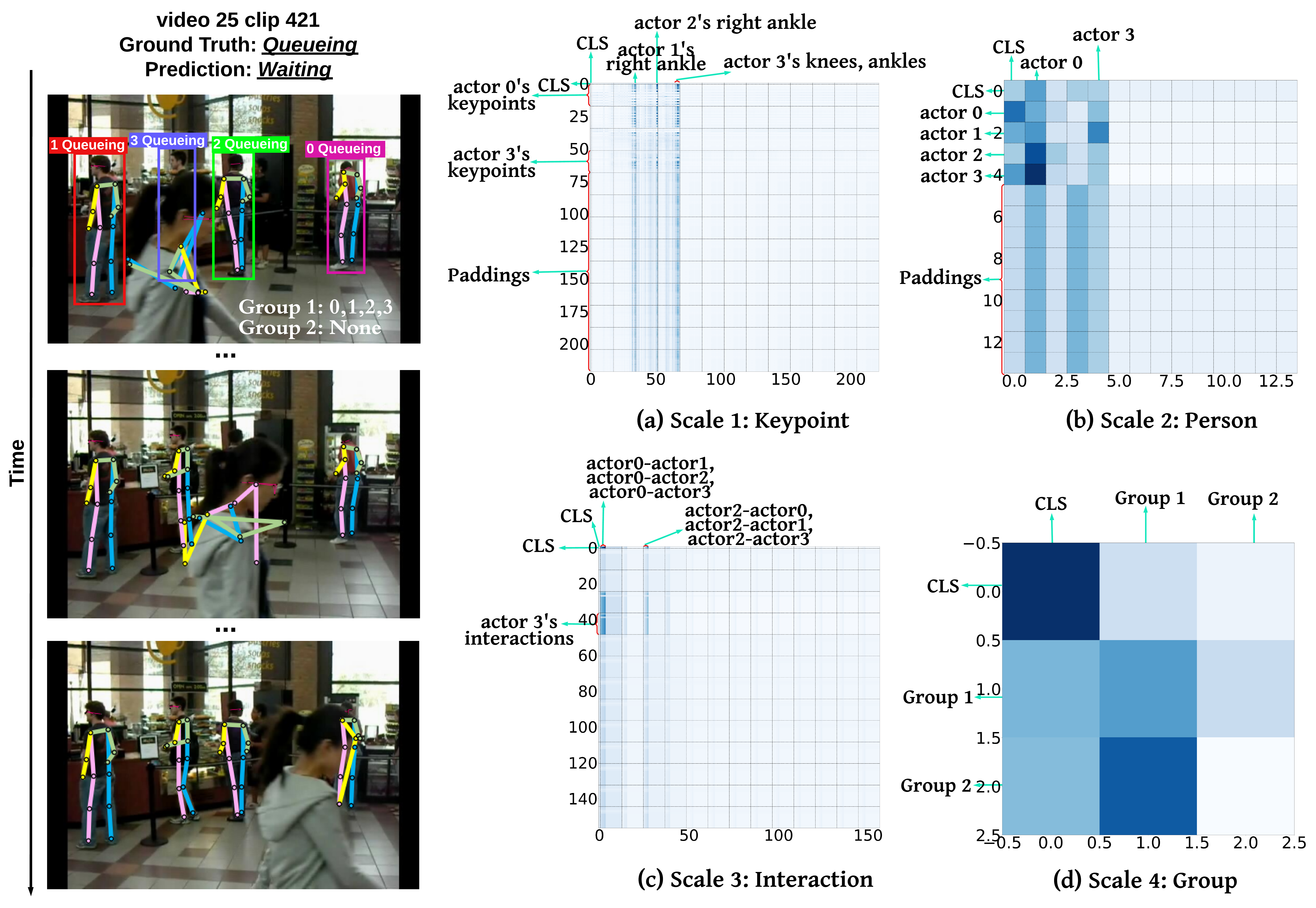} 
	\caption{\textbf{A failure case on CAD.} Because of the severe occlusion, the person queue is interrupted and  \ours predicts \textit{waiting} instead of \textit{queueing}.
}
	\label{fig:CAD_bad_occlusion} 
\end{figure*} 

\begin{figure*}[t]
	\centering
	\includegraphics[scale=0.14]{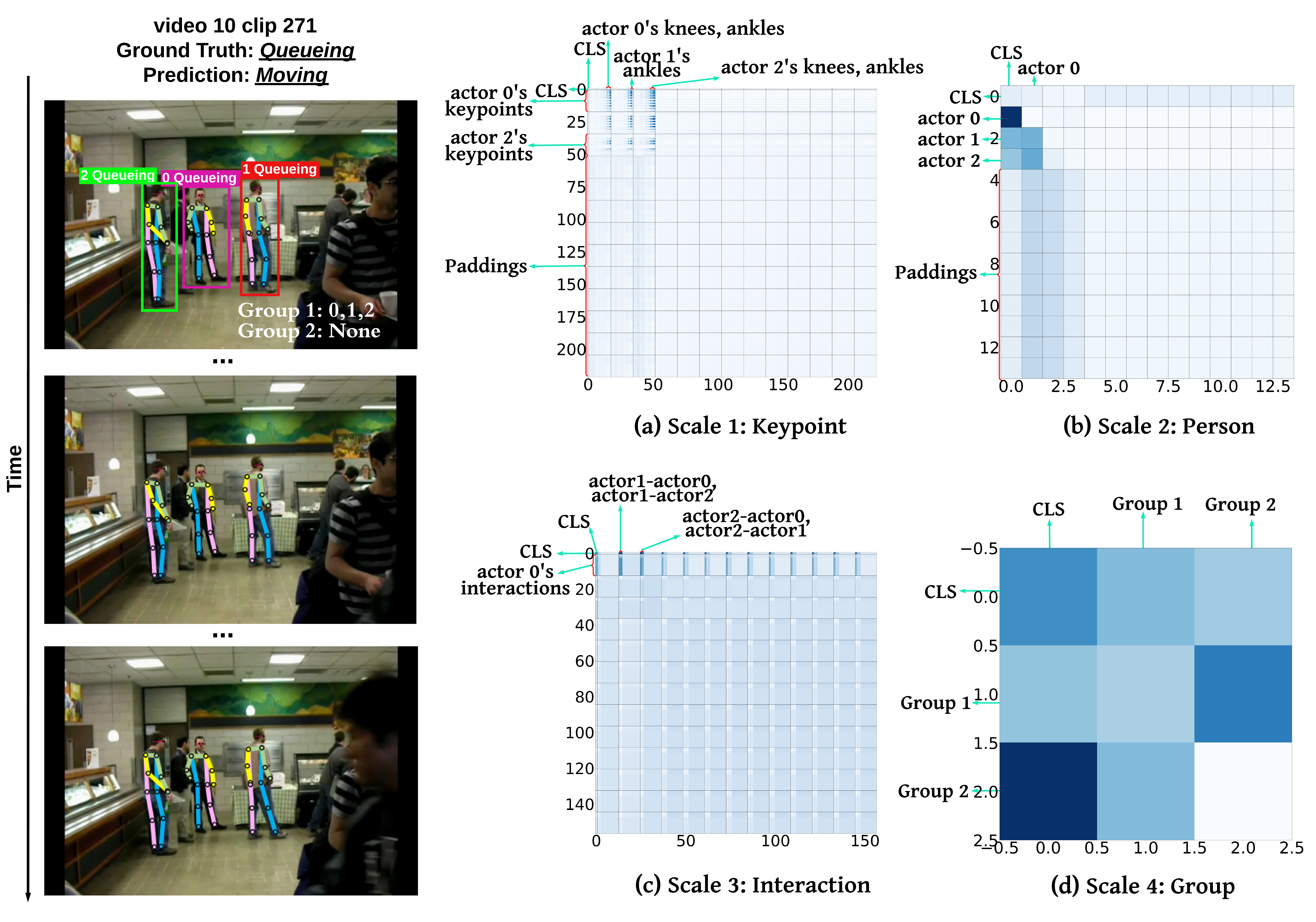} 
	\caption{\textbf{A failure case on CAD.} The movement of actor $1$'s leg might cause the wrong prediction of \ourseos.
}
	\label{fig:CAD_bad_pred_moving_gt_queueing} 
\end{figure*}
\clearpage
\begin{figure*}[t]
	\centering
	\includegraphics[scale=0.14]{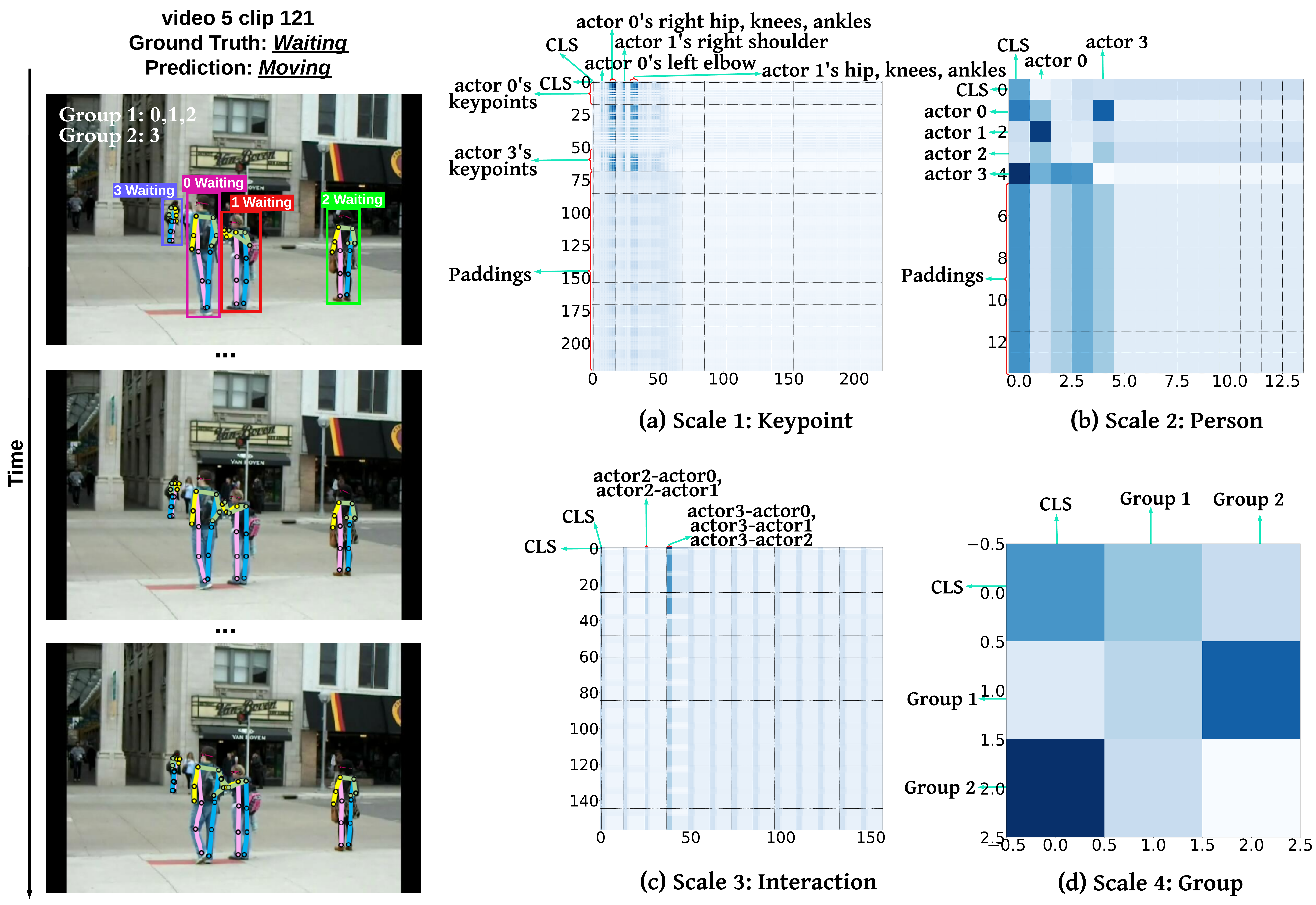} 
	\caption{\textbf{A failure case on CAD.} The movement of actor 0’s leg might cause the
wrong prediction of \ourseos.}
	\label{fig:CAD_bad_pred_moving_gt_waiting} 
\end{figure*} 

\begin{figure*}[t]
	\centering
	\includegraphics[scale=0.14]{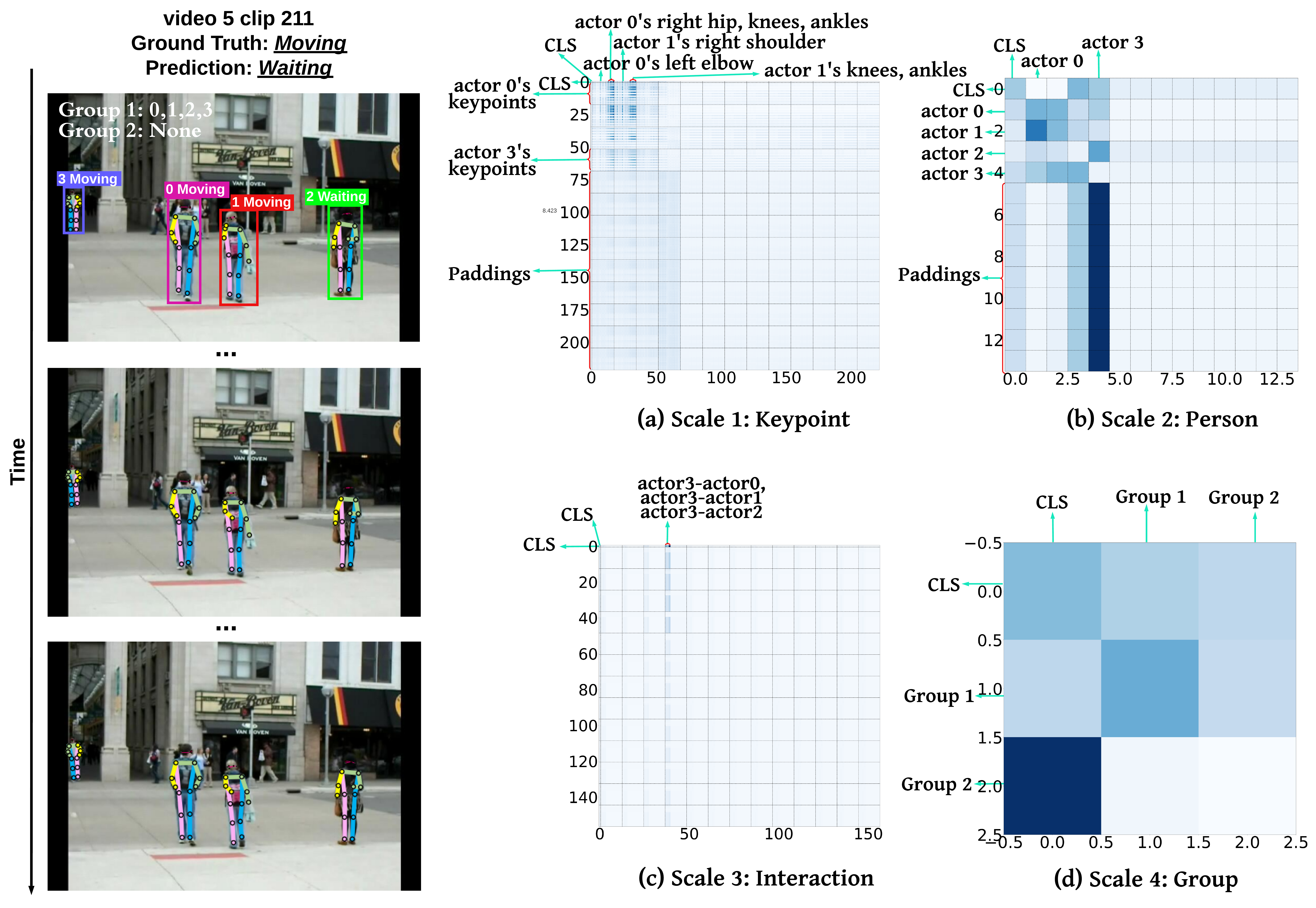} 
	\caption{\textbf{A failure case on CAD.} In the clip, persons were waiting before starting to cross the street. 
	The dynamics are not easily perceptible due to the short temporal window of the clip.
}
	\label{fig:CAD_bad_pred_short_temporal} 
\end{figure*}

\end{document}